\pdfoutput=1
\documentclass{article}

\usepackage[main,final]{neurips_2026}
\usepackage[utf8]{inputenc}
\usepackage[T1]{fontenc}
\usepackage{hyperref}
\usepackage{url}
\usepackage{booktabs}
\usepackage{amsmath}
\usepackage{amssymb}
\usepackage{amsfonts}
\usepackage{amsthm}
\usepackage{mathtools}
\usepackage{nicefrac}
\usepackage{dirtytalk}
\usepackage{enumitem}
\usepackage{microtype}
\usepackage{xcolor}
\usepackage{graphicx}
\usepackage{enumitem}
\usepackage{wrapfig2}
\usepackage{etoc}
\usepackage{soul}

\makeatletter
\AtBeginDocument{%
  \let\NAT@hyper@orig\NAT@hyper@
  \def\NAT@hyper@#1{\mbox{\NAT@hyper@orig{#1}}}%
  \def\hyper@natlinkbreak#1#2{#1}%
}
\makeatother

\newcommand{\Stieltjes}{G}
\newcommand{\LogStieltjes}{F}
\newcommand{\StieltjesB}{B}

\newcommand{\orthogonalgroup}{\mathrm{O}}
\newcommand{\hciz}{\mathcal{J}_{\mathrm{HCIZ}}}
\newcommand{\BigO}{\mathcal{O}}

\newcommand{\Prob}{P}
\newcommand{\Prior}{\Prob}
\newcommand{\Energy}{\mathcal{E}}
\newcommand{\Evidence}{\mathcal{Y}}
\newcommand{\PredPost}{\Prob_{\mathrm{pp}}}
\newcommand{\Wteacher}{W^{\ast}}
\newcommand{\Pteacher}{\Prob^{\ast}}
\theoremstyle{plain}

\DeclareMathOperator{\Trace}{tr}
\DeclareMathOperator{\Rank}{rank}

\title{Spherical Boltzmann machines: a solvable theory of learning and generation in energy-based models}

\author{%
  Thomas Tulinski, Simona Cocco, Rémi Monasson \\
  Laboratoire de Physique de l'École Normale Supérieure, PSL,\\
  CNRS UMR8023, Sorbonne Université, 24 rue Lhomond, 75005 Paris, France
  \And
  Jorge Fernandez-de-Cossio-Diaz \\
  Institut de Physique Théorique, Université Paris-Saclay,\\
  CNRS UMR3681, CEA, Gif-sur-Yvette, France \\
  \texttt{jorge.fdcd@ipht.fr}
}

\begin{document}

\etocdepthtag.toc{mainmatter}

\maketitle

\begin{abstract}
Energy-based models (EBMs) are flexible generative architectures inspired by statistical physics, but their learning and generative properties remain poorly understood. Here, we analyze a solvable EBM in the high-dimensional limit: the spherical Boltzmann machine (SBM). Combining tools from random matrix theory and dynamical mean-field theory, we: derive and solve closed equations describing the training dynamics of the SBM in the high-dimensional limit; compute the asymptotic Bayesian evidence, which acts as a partition function in parameter space and encodes global properties of the trained model; and uncover cascades of phase transitions that occur both during training and as a function of hyperparameters, related to successive alignment and condensation of the top modes of the coupling matrix to the data. We connect these transitions to sampling-time generative phenomena in a teacher-student scenario, including: sampling temperature tuning, double descent as a function of regularization strength, tempered posterior effects, and out-of-equilibrium effects during training that induce biases in the trained model. We provide numerical evidence showing that all these phenomena appear in standard generative architectures, beyond the SBM.\footnote{Code: \url{https://github.com/cossio/spherical-bm-2026}.}
\end{abstract}

\section{Introduction}
\label{sec:intro}

Energy-based models (EBMs) define complex probability distributions over high-dimensional spaces, placing mass on regions where realistic data lie while suppressing it elsewhere. The main ingredients are an energy function $\Energy(\mathbf{x};W)$, defined over data points $\mathbf{x}$ and depending on parameters $W$, which is used to define a probability distribution through a Boltzmann law $\Prob(\mathbf{x}|W) = e^{-\beta\Energy(\mathbf{x};W)}/Z_{\beta}(W)$, where $\beta$ is an inverse sampling temperature and $Z_{\beta}(W) = \int e^{-\beta\Energy(\mathbf{x};W)}\mathrm{d}\mathbf{x}$ a normalization constant known as the partition function in statistical physics. EBMs are competitive generative models in diverse applications, including images \citep{du2019implicit}, biological sequences \citep{fernandez2025designing,russ2020evolution}, and others. But the computation of $Z$ requires a typically intractable integration over the high-dimensional point $\mathbf{x}$. As a consequence, training and sampling of EBMs often relies on approximate Monte Carlo (MC) techniques. Beyond computational cost, the intractable $Z$ also blocks theoretical progress on the analysis of the trained model.

\paragraph{Contributions.}
We study \emph{spherical Boltzmann machines} (SBMs) trained on data in the high-dimensional limit. SBMs are solvable EBMs defined on the $N$-sphere. Combining recent results on finite-rank spherical-integral asymptotics \citep{guionnet_husson_2022} which encode the alignment of $W$'s eigenvectors with the data principal components, with a Coulomb-gas analysis of $W$'s eigenvalues as repelling charges in an external potential set by the data and the prior, we obtain a closed-form large-$N$ theory of this model. We derive the equilibrium phase diagram of the trained SBM, which has qualitatively different properties in different hyperparameter settings; solve the high-dimensional Langevin dynamics of training via dynamical mean-field theory (DMFT) \citep{agoritsas2018out,cugliandolo2023recent} uncovering a non-equilibrium phase transition in the training dynamics; and compute closed-form metrics of generative performance in a teacher-student setting.

Our solvable model predicts four effects, that we fully characterize: \textbf{sampling temperature tuning}, where tuning $\beta\ne1$ post-training can improve sample fidelity; \textbf{double descent} of generative metrics as a function of regularization; \textbf{tempered posterior} effects, where the performance of an ensemble of models weighted by the Bayes posterior improves if the posterior is annealed at a posterior temperature $\eta\ne1$; and \textbf{out-of-equilibrium} training dynamics leading to eventual deterioration of the model during training.

We provide numerical evidence that all these phenomena arise in standard generative architectures, in qualitative agreement with the predictions from our solvable SBM theory. 
These experiments show that the four behaviors recur beyond the SBM and, in several cases, exhibit spectral signatures compatible with the SBM picture.

\textbf{Related works.}
Statistical-mechanics tools have been used to characterize the parameter space of discriminative networks \citep{gardner1988space,zambon2025sampling,zdeborova2016statistical} and of generative models with random weights \citep{amit1985spin,tubiana2017emergence,genovese2020legendre,decelle2021restricted}, neglecting the impact of training on data. For trained EBMs, analytical progress has relied on early-training approximations \citep{bachtis2024cascade,fachechi2025rbm,theriault2024rbm}. The theoretical understanding of the structure of the learned weights is challenged by the complexity of the partition function term. In a related line of works, DMFT has emerged as a standard tool to study the high-dimensional learning trajectories of supervised models \citep{montanari2025dynamical,bonnaire2024high,urbani2026theory}, and has been applied to maximum-likelihood training of RBMs with finitely many hidden units, using an asymptotic reduction of the likelihood to analyze data with order-one covariance spikes and a sample size proportional to the data dimension \citep{xu2025learning}. This reduction makes the negative phase tractable, but leaves open the problem of describing the coupled dynamics of learning and persistent negative-phase sampling, as commonly used in practical EBM training. %

Sampling temperature tuning and out-of-equilibrium training dynamics are familiar to EBM practitioners. For example, in generative models of biological sequences, setting $\beta>1$ at sampling time improves the functionality of designed sequences in experiments \citep{russ2020evolution,fernandez2025designing}, but this often comes at the expense of generation diversity \citep{ilmc2026} 
and the trade-off is not very well understood \citep{fields2025}. Persistent-sampling MC is frequently used to estimate the learning gradient during EBM training, but training dynamics often reach out-of-equilibrium regimes, inducing subtle and hard-to-control biases in the trained model \citep{decelle2023unsupervised,bachtis2024cascade}; in other settings, models are deliberately trained out-of-equilibrium \citep{decelle2021equilibrium,nijkamp2019learning}. %
Analytical results for non-persistent short-run MCMC training have been obtained for Gaussian EBMs and mean-field Boltzmann machines \citep{agoritsas2023explaining}. But overall, these strategies are largely empirical, and the regimes in which they operate remain poorly understood.

Double descent \citep{belkin2019reconciling,mei2022generalization} and tempered posterior effects \citep{wenzel2020cold,pitas2024fine} have been extensively studied in the context of supervised learning, though the origins of these effects are still debated. In generative models, double descent has been reported in GANs \cite{luzi2024double}, LLMs \citep{hernandez2022scaling,morris2025memorize}, diffusion models \citep{zhang2025diffusion,fraij2025double}, autoencoders \citep{rahimi2025unveiling}, RBMs \citep{cheema2020volume}, and others, in a variety of settings, including as a function of hyper-parameters, sample complexity, or training duration. Bayesian ensembles of generative models have also been proposed \citep{murray2012bayesianlearningundirectedgraphical,saatchi2017bayesian}, but to the best of our knowledge, tempered posterior effects have not been reported before in a generative context.

A solvable EBM accounting for all these effects is missing. The SBM closes this gap: equilibrium structure, training dynamics, and generative phenomenology are characterized in a unified asymptotic large-$N$ framework from which the four effects emerge.

\section{Spherical Boltzmann machines: Equilibrium and training dynamics}
\label{sec:model}

The SBM is defined on the $N$-sphere, $\mathcal{S}_N=\{\mathbf{x}\in\mathbb{R}^N:\|\mathbf{x}\|^{2}=N\}$, with the energy function
\begin{equation}\label{eq:SBM}
  \Energy(\mathbf x;W) = -\frac{1}{2}\mathbf{x}^{\top}W\mathbf{x},
  \qquad
  \mathbf{x}\in\mathcal{S}_N,
  \quad
  W\in\mathrm{Sym}_N,
\end{equation}
The SBM is a classical model from the statistical physics of disordered systems \citep{berlin1952spherical,kosterlitz1976spherical}, arising as a continuous relaxation of Ising spin systems, in which the per-site constraint $x_i\in\{\pm1\}$ is replaced by the global constraint $\|\mathbf{x}\|^2=N$. Despite the quadratic energy, $P(\mathbf x|W)$ is not Gaussian: it is the Bingham distribution of directional statistics \citep{bingham1974antipodally,kent1982fisher}. The phenomenology described below depends critically on the spherical constraint. Whereas $W$ is purely random in most prior works, a few exceptions consider learning, such as \citet{decelle2020gaussian}, who study Gaussian-spherical RBMs but do not derive the large-$N$ spectral law of the trained weights. In the present work, the SBM is treated as an EBM trained on data, which imprints statistical structure in $W$ that we characterize both at equilibrium and along the training dynamics.

Without loss of generality, one can assume that $\beta=1$ during training, but $\beta\neq 1$ is often tuned post-training with the hope of improving the quality of generated samples. Given a training dataset $\mathcal{D}=\{\mathbf{x}^k\}_{k=1}^{K}\subseteq\mathcal{S}_N$, the likelihood assigned by the SBM depends on $\mathcal{D}$ only through the rescaled empirical covariance matrix $C=\tfrac{1}{N}\sum_{k}\mathbf{x}^k\mathbf{x}^{k\top}$, with the usual empirical covariance given by $(N/K)C$. The model is then trained by ascent of the log-posterior. In the continuous-time limit, the training dynamics can be described by a Langevin equation,
\begin{equation}\label{eq:W-langevin-1}
\partial_{t}W(t) = \frac12\left(C - \frac{K}{N}\langle\mathbf{x}\mathbf{x}^\top\rangle_{\mathbf{x}\sim \Prob_{W(t)}}\right) 
- \frac{\gamma}{2}W(t) + \frac{\Omega(t)}{\sqrt{\eta N}},
\end{equation}
where $\gamma>0$ controls the strength of weight decay, $\Omega(t)$ is a symmetric matrix-valued white noise satisfying $\left\langle \Omega_{ij}(t)\Omega_{kl}(t') \right\rangle=\left( \delta_{ik}\delta_{jl}+\delta_{il}\delta_{jk} \right)\delta(t-t')$ arising \emph{e.g.} from minibatch sampling in stochastic gradient ascent, and $\eta>0$ an inverse \emph{learning temperature} controlling the amplitude of this noise.
It is important to distinguish the two inverse temperatures from each other: $\eta$ controls the dispersion in the space of model parameters $W$, while $\beta$ acts in the sample-space, $\mathbf x$.

The moment average of $\mathbf{x}\mathbf{x}^\top$, also called the negative phase in the EBM literature, must be computed by drawing equilibrated samples from $\Prob(\mathbf{x}|W(t))$, a daunting task in high-dimensional settings. In practice, this term is approximated by MC sampling. A popular variant of the Boltzmann machine learning algorithm consists of running persistent Markov chains, which are frequently updated before each gradient update of the weights \citep{younes1989parametric,tieleman2008training,rosset2026adabmdca}. In this case, the training dynamics can be modeled by the coupled Langevin equations \citep{dotsenko1994partial,coolen_NIPS1993_c0f168ce}:
\begin{align}
  \partial_{t}W(t) &= 
  \frac{1}{2}\left(C - \frac{K}{N}\mathbf{x}(t)\mathbf{x}(t)^\top\right) - \frac{\gamma}{2}W(t) + \frac{\Omega(t)}{\sqrt{\eta N}},\label{eq:W-langevin} \\
  \partial_{t}\mathbf{x}(t) &= 
  \nu W(t)\mathbf{x}(t) - \kappa(t)\mathbf{x}(t) + \sqrt{2\nu}\,\boldsymbol{\xi}(t),\label{eq:x-langevin}
\end{align}
where $\mathbf x(t)$ is the persistent MC chain, with $\kappa(t)$ a Lagrange multiplier enforcing $\|\mathbf{x}(t)\|^2=N$, the independent white noise $\boldsymbol{\xi}(t)$ satisfying $\left\langle \xi_i(t)\xi_j(t') \right\rangle=\delta_{ij}\delta(t-t')$, and a rate $\nu$ controlling how many sampling steps are performed on the persistent MC chain per gradient update of the weights. In practice, increasing $\nu$ requires spending more compute cycles updating the MC chains during training.

Equations \eqref{eq:W-langevin}-\eqref{eq:x-langevin} define the exact training dynamics of our model. In the following, we study the large-$N$ asymptotic solution of this model in two Data Regimes:

\textbf{DR1.} \emph{Undersampled regime.} The number of training samples $K=\mathcal{O}(1)$ stays finite as $N\to\infty$. The nonzero eigenvalues of $C$ and the hyperparameters are $\mathcal{O}(1)$.

\textbf{DR2.} \emph{Effective low-dimensional data.} $N,K\rightarrow\infty$. A finite number $K'=\mathcal{O}(1)$ of eigenvalues of $C$ grow linearly with $K$, while the remaining nonzero eigenvalues are $o(K)$ (sublinear) as $K$ grows. No restriction is imposed on the relative scaling of $N$ and $K$; in particular, the proportional regime $K/N\rightarrow\alpha\in(0,\infty)$ is included. Intuitively, the $K$ data points are effectively low-dimensional, with $K'$ dominant directions. We consider two sub-cases: \textbf{DR2.1} the hyperparameters stay fixed as $K$ grows, treated in App.~\ref{app:bare-large-k};  and \textbf{DR2.2} the hyperparameters are rescaled with $K$, treated in App.~\ref{app:invar-large-k}.

We focus on DR1 in the remainder of the main text for simplicity. The equilibrium and dynamical phase structure derived below carries over unchanged to DR2.2 in the rescaled variables, while DR2.1 is recovered as a boundary limit of DR2.2 (App.~\ref{app:large-K-recovering}).

\subsection{Equilibrium phase diagram\label{sec:phase-diagram}}

In the limit $\nu\to\infty$, $\mathbf{x}(t)$ approaches an equilibrium sample from $\Prob(\mathbf{x}|W)$ in a quasi-stationary $W$ background. In this case, the trained $W(t\to\infty)$ converges to the stationary distribution:
\begin{equation}\label{eq:tilted-posterior}
\Prob_\eta(W|\mathcal{D}) = \frac{e^{\mathcal{L}(W;C)}}{\Evidence(C)},
\quad\text{with }\quad
\mathcal{L}(W;C) = \frac{N}{2}\eta\Trace CW - \frac{N}{4}\gamma\eta\Trace W^2 - K\eta\ln Z(W),
\end{equation}
a tempered posterior at temperature $1/\eta$, that interpolates between the standard Bayes posterior at $\eta=1$ and the maximum-a-posteriori (MAP) limit $\eta\to\infty$. The tempered evidence, $\Evidence(C) = \int_{W\in\mathrm{Sym}_N} e^{\mathcal{L}(W;C)}\,\mathrm{d}W$, plays the role of a partition function over the weights.

Changing variables to the eigendecomposition $W=\tfrac{1}{N}V\Lambda V^\top$, we can rewrite $\mathcal{L}(W;C)$ as a Coulomb gas over eigenvalues $\Lambda$: $\lambda_1>\dots>\lambda_N$ \citep{livan2018introduction},
\begin{equation}\label{eq:LLambda}
\mathcal{L}_\Lambda(\Lambda;C) = 
-\frac{N}{4}\gamma\eta\Trace\Lambda^2 - K\eta \ln Z(\Lambda) + \ln\hciz(\Lambda;C) + \sum_{i<j}\ln|\lambda_i-\lambda_j|,
\end{equation}
where $\hciz$ is the real analogue of the Harish--Chandra--Itzykson--Zuber (HCIZ) integral \citep{guionnet_husson_2022} over the eigenvectors $V=(\mathbf v_j)$, normalized to $\mathbf v_i\cdot\mathbf v_j=N\delta_{ij}$.

In the large-$N$ limit, the eigenvalue positions are obtained by optimizing $\mathcal{L}_\Lambda$. A dominant $\BigO(N^{2})$ contribution comes from the terms $-\tfrac{N}{4}\gamma\eta\Trace\Lambda^2 + \sum_{i<j}\ln|\lambda_i-\lambda_j|$, which do not involve the training data. These terms describe a standard Coulomb gas, with half-width parameter $\sigma=1/\sqrt{\gamma\eta}$. The bottom $N-K$ eigenvalues then form a Wigner semicircle bulk,
\begin{equation}\label{eq:rho}
  \rho(\lambda) = \frac{1}{N-K}\sum_{i=K+1}^{N} \delta(\lambda_i - \lambda) \underset{N\to\infty}{\to}  \frac{\sqrt{4\sigma^2-\lambda^2}}{2\pi\sigma^2}\mathbf{1}_{|\lambda|\le2\sigma}.
\end{equation}

\begin{figure}[t]
\centering
\includegraphics[width=\linewidth]{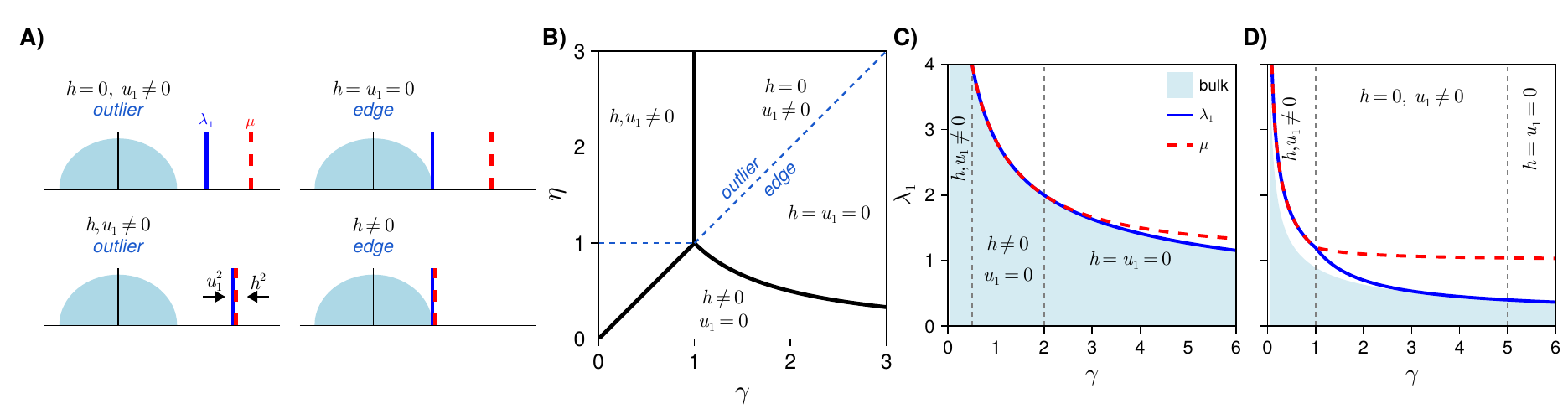}
\caption{\textbf{Equilibrium phase diagram for $K=1$.} \textbf{A)} Schematic eigenvalue configurations in each phase: bulk semicircle (light blue), top eigenvalue $\lambda_1$ and the Lagrange multiplier $\mu$ enforcing $\|\mathbf{x}\|^2=N$. \textbf{B)} Phases in the $(\gamma,\eta)$ plane; the blue dashed line separates outlier from edge regimes. \textbf{C)} $\lambda_1$ and Lagrange multiplier $\mu$ vs.\ $\gamma$ in the small-$\eta$ regime, where $\lambda_1$ sticks to the edge; phase transitions occur at the vertical dashed lines. \textbf{D)} Same in the larger-$\eta$ regime, where $\lambda_1$ detaches before returning to the edge.}
\label{fig:k1-phase}
\end{figure}

The top $K$ eigenvalues interact with the data and may detach from the bulk, so we must treat them separately (Fig.~\ref{fig:k1-phase}A). Let $C=\frac{1}{N}\sum_{k}c_{k}\mathbf{c}_{k}\mathbf{c}_{k}^{\top}$ be the eigendecomposition of $C$, with eigenvectors normalized to $\mathbf{c}_{i}\cdot\mathbf{c}_{j}=N\delta_{ij}$. As a consequence of the $\orthogonalgroup(N)$-rotation invariance of~\eqref{eq:SBM}, the following analysis depends on $C$ only through its eigenvalues, $c_1>\dots>c_K>0$.

We determine the positions of $\lambda_1,\dots,\lambda_K$ by combining finite-rank large-$N$ asymptotics of $\hciz$ derived by \cite{guionnet_husson_2022} with a saddle-point calculation extremizing~\eqref{eq:LLambda}, reported in App.~\ref{app:phase-conditions}. Intuitively, this yields a balance of $\BigO(N)$ forces acting on each of the top $K$ eigenvalues $\lambda_k$. Besides the interaction with the bulk, the likelihood terms contribute two additional forces (arrows in Fig.~\ref{fig:k1-phase}A),
\begin{equation}\label{eq:forces}
\text{HCIZ pull away from bulk: }
\frac{N}{2}\eta c_k u_k^2
\qquad
\text{partition push toward bulk: }
-\frac{NK}{2}\eta h^2
\end{equation}
where $u_{k}^{2}=\langle(\mathbf c_k\cdot\mathbf v_k/N)^{2}\rangle$ is the overlap of the $k$-th eigenvector of $W$ with the $k$-th eigenvector of $C$, $h_{k}^{2}=\langle(\mathbf x\cdot\mathbf v_k/N)^{2}\rangle$ the overlap of typical generated samples with the $k$-th eigenvector of $W$, and $h^{2}=\sum_{k=1}^{K}h_{k}^{2}$. The order parameters $u_k,h$ characterize the trained model parameters and typical data it generates. They concentrate around typical values, defining the following qualitatively distinct regimes:

\begin{itemize}[leftmargin=*,topsep=1pt,itemsep=1pt]
  \item $h=u_{1,\dots,K}=0$ and $\lambda_{1,\dots,K}=2\sigma$. If $\gamma\eta>1$ and $\gamma>\eta c_1^2$, the eigenvalues stick to the bulk edge, the trained weights do not align macroscopically to the data, and generated samples are nearly isotropic on the sphere.
  \item $h=0,u_{1,\dots,a}\ne0$. If $\gamma>c_{1}$ and $\eta c_{a+1}^{2}<\gamma<\eta c_{a}^{2}$; the top $a\in[K]$ modes detach from the bulk at $\lambda_k = (\eta c_k)^{-1}+c_k/\gamma$, aligning to the data. But this is insufficient to trigger condensation: generated samples are still nearly isotropic on the sphere.
  \item $h\ne0,u_{1,\dots,d}\ne0$: the top $d\in[K]$ eigenvalues coalesce at a common position $\lambda_1$ (see Eq.~\eqref{eq:g1-coalesced} in App.~\ref{app:phase-conditions}) aligning to the data. Two sub-phases can be distinguished, according to whether $\lambda_1$ sticks to the semi-circle or detaches (conditions in Eqs.~\eqref{eq:FMe-conditions}, \eqref{eq:FMo-conditions}).
  \item $h\ne0,u_{1,\dots,K}=0$ and $\lambda_{1,\dots,K}=2\sigma$. If $\eta^2 c_1^2<\gamma\eta<1$, generated samples condense along a random direction, unrelated to the data.
\end{itemize}

The phase diagram for $K=1$ is shown in Fig.~\ref{fig:k1-phase}B. Fig.~\ref{fig:k1-phase}C,D show $\lambda_{1}$ as a function of $\gamma$ for two selected values of $\eta$, illustrating the outlier detachment and condensation transitions. For $K>1$ different modes can be in different phases, and top modes can coalesce and condense simultaneously (App. Fig.~\ref{fig:phase-diagram-k=2-appendix} for $K=2$). In phases with $h=0$, the projections $\mathbf x\cdot\mathbf v_{j}$ are all comparable and stay $\mathcal{O}(\sqrt{N})$, covering the sphere in a biased but nearly isotropic fashion. Phases with $h\ne0$ are qualitatively different: samples \emph{condense} along the top mode(s) of $W$, with significantly larger projections $\mathcal{O}(N)$ along $\mathbf v_{1,\dots,d}$, while projections along other directions remain $\mathcal{O}(\sqrt{N})$.
See App.~\ref{app:phase-conditions} for derivations and App.~\ref{app:finite-N-equilibrium} for finite-$N$ equilibrium comparisons. See also \citep{tulinski2026} and App.~\ref{app:replica} for an alternative approach based on the replica method.

\subsection{Training dynamics\label{sec:dynamics}}

We next analyze how the Langevin dynamics~\eqref{eq:W-langevin}--\eqref{eq:x-langevin} approach the equilibrium $\Prob_\eta(W|\mathcal{D})$ described in Sec.~\ref{sec:phase-diagram}. Equilibration is guaranteed only in the fast sampling limit $\nu\to\infty$; at finite $\nu$ training can settle into a non-equilibrium stationary state qualitatively distinct from the equilibrium phase diagram.

We initialize $W(0)$ at random from the prior and $\mathbf{x}(0)$ with a small seed overlap with the top data mode $\mathbf c_1$. Under these assumptions, Eq.~\eqref{eq:W-langevin} can be rewritten in the equivalent integrated form:
\begin{equation}\label{eq:W-integrated-main}
  W(t) = W_{\mathrm{GOE}}(t)+\frac{1-e^{-\gamma t/2}}{\gamma}C-\frac{K}{2}\int_{0}^{t}e^{-\gamma(t-u)/2} \frac{\mathbf{x}(u)\mathbf{x}(u)^{\top}}{N}\mathrm{d}u
\end{equation}
where $W_{\mathrm{GOE}}(t)$ is a standard Gaussian orthogonal ensemble (GOE) matrix process,
\begin{equation}
  \partial_{t}W_{\mathrm{GOE}}(t) = -\frac{\gamma}{2}W_{\mathrm{GOE}}(t) + \frac{\Omega(t)}{\sqrt{\eta N}}, 
  \quad 
  W_{\mathrm{GOE}}(0)=W(0),
\end{equation}

The first term in \eqref{eq:W-integrated-main} is a growing GOE noise, the second a deterministic spike in the data directions, and the third a negative-phase feedback from the generated samples. Their combination produces a cascade of eigenvalue detachments from the bulk. At every time, $W(t)$ consists of a Wigner bulk plus at most $K$ upper outliers (proof in App.~\ref{app:outlier-count}).

\paragraph{Effective DMFT.} We derive a closed set of equations describing the $N\to\infty$ limit of the process \eqref{eq:W-integrated-main} through a DMFT saddle-point analysis. Although non-rigorous, the derivation retains all leading-order terms. As in many standard applications of DMFT, the resulting closure is therefore expected to be asymptotically exact. Briefly, representing the dynamics by a Martin--Siggia--Rose path integral \citep{martin_siggia_rose_1973}, and integrating out the weights and noises yields, at the large-$N$ saddle, the following closed equations for the overlaps $s_{k}(t) = \mathbf{c}_{k}\cdot\mathbf{x}(t)/N$, the correlator $Q(t,t')=\mathbf{x}(t)\cdot\mathbf{x}(t')/N$, and the causal response $R(t,t')$:
\begin{align}
  \partial_{t}s_{k}(t) &= -\kappa(t)s_{k}(t) + \frac{\nu}{\gamma}(1-e^{-\gamma t/2})c_{k}s_{k}(t) + \int_{0}^{t}M(t,u)s_{k}(u)\,\mathrm{d}u, \label{eq:s-eom}\\
  \partial_{t}R(t,t') &= -\kappa(t)R(t,t') + \int_{t'}^{t}M(t,u)R(u,t')\mathrm{d}u,
  \label{eq:R-eom} \\
  \partial_{t}Q(t,t') &= \frac{\nu}{\gamma}(1-e^{-\gamma t/2})\sum_{k=1}^{K}c_{k}s_{k}(t)s_{k}(t') - \kappa(t)Q(t,t') \notag \\
  &\quad + \int_{0}^{t}M(t,u)Q(u,t')\mathrm{d}u + \int_{0}^{t'}D(t,u)R(t',u)\mathrm{d}u, \label{eq:Q-eom}
\end{align}
subject to $Q(t,t')=Q(t',t)$, $R(t,t')=0$ if $t<t'$, $R(t,t)=1$, with $\kappa(t)$ at all times fixed by the spherical constraint $Q(t,t)=1$. See App.~\ref{app:dynamics} for derivations, including the definitions of the memory kernel $M(t,u)$ and the noise correlator $D(t,u)$, and further validation of the closure~\eqref{eq:s-eom}--\eqref{eq:Q-eom} against finite-$N$ simulations.

\begin{figure}[t]
\centering
\includegraphics[width=\linewidth]{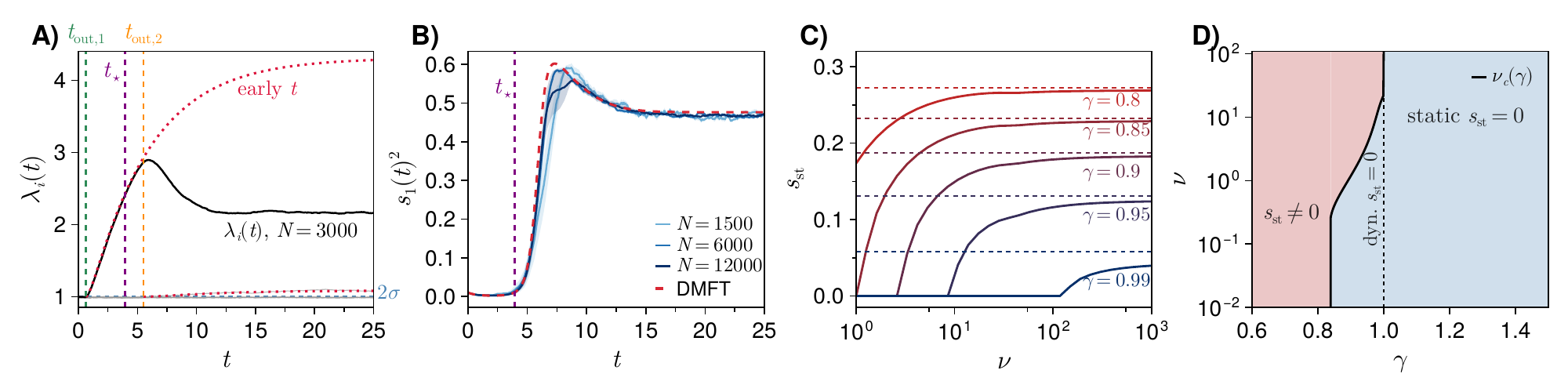}
\caption{\textbf{Training dynamics.} \textbf{A)} Top eigenvalue trajectories $\lambda_{i}(t)$ from finite-$N$ simulations (black) vs. the early-time predictions~\eqref{eq:lambda-para} (red dotted). Vertical dashed lines are the detachment~\eqref{eq:lambda-para} and condensation times~$t_{*}$. Blue horizontal line is the bulk edge ($2\sigma$). \textbf{B)} $s_1(t)$ from DMFT (red) compared to finite-$N$ simulations (blue). \textbf{C)} Stationary overlap $s_{\mathrm{st}}$ as a function of $\nu$ for selected values of $\gamma$ (dashed lines indicate the equilibrium $\nu\to\infty$ asymptotes). \textbf{D)} MAP-limit dynamical phase diagram ($\eta\to\infty$), with $\nu_\mathrm{c}$ separating the $s_\mathrm{st}\ne0$ from the $s_\mathrm{st}=0$ phases. For $\gamma>1$ (dashed), $s_\mathrm{st}=0$ also at equilibrium.}
\label{fig:dynamics}
\end{figure}

\paragraph{Eigenvalue detachment cascade and condensation.}
Early in the dynamics the negative-phase contribution is weak ($s_k^2(t)\ll c_k/\gamma$) and Eq.~\eqref{eq:W-langevin} can be approximated by a deterministic spike growth, which places the $k$-th eigenvalue at the outlier position (App.~\ref{app:dynamics})
\begin{equation}\label{eq:lambda-para}
\lambda_{k}(t) = \begin{cases}
\dfrac{1}{g_k(t)} + \dfrac{g_k(t)}{\gamma\eta} & t \ge t_{\mathrm{out},k},\\
2\sigma & t\le t_{\mathrm{out},k},
\end{cases}\qquad
\begin{aligned}
\text{with } g_{k}(t) &= \frac{\gamma}{c_k(1-e^{-\gamma t/2})}, \text{ and} \\
\text{with } t_{\mathrm{out},k} &= -\frac{2}{\gamma}
\ln\left(1-\frac{1}{c_{k}}\sqrt{\frac{\gamma}{\eta}}\right)
\end{aligned}
\end{equation}
where $g_k(t)=\Stieltjes(\lambda_k(t))$ is the Stieltjes transform \eqref{eq:Stieltjes} evaluated at the $k$-th outlier position, and $t_{\mathrm{out},k}$ is the detachment time, defined for $c_k>\sqrt{\gamma/\eta}$ (otherwise $\lambda_k$ never detaches). Fig.~\ref{fig:dynamics}A shows the top eigenvalue trajectories $\lambda_i(t)$ from finite-$N$ Langevin simulations, in agreement with the early time prediction~\eqref{eq:lambda-para}.

Fig.~\ref{fig:dynamics}B shows $s_{1}(t)$ from DMFT in agreement with averaged finite-$N$ simulations. If $\nu$ is large enough (see next paragraph), $s_{1}(t)$ first decays transiently before condensing along the top mode, after which $\lambda_{1}(t)$ deviates from the early time prediction~\eqref{eq:lambda-para}. The time at which this happens can be computed as the $t_{*}$ for which $s_1(t_{*})$ reaches a prescribed small value, signaling onset of condensation of the training sample, after which $s_1(t)$ grows to a nonzero stationary value. We remark that here we initialize $\mathbf x(0)$ near the data, so $s_1(0)>0$. Otherwise, if $\mathbf x(0)$ were initialized at random, $t_{*}$ can be much larger (growing with $N$) as the training sample searches a distinguished spike direction in high-dimensional space \citep{bonnaire2024high}.

\paragraph{Stationary solution.} For long times, the DMFT admits a stationary solution with $s_k(t)\to s_{\mathrm{st},k}$, $Q(t,t')\to Q_{\mathrm{st}}(t-t')$ and $R(t,t')\to R_{\mathrm{st}}(t-t')$. The trivial branch $s_{\mathrm{st},1}=0$ is the only solution for low $\nu$, but above a critical $\nu_{\mathrm{c}}(\gamma,\eta)$, a condensed branch $s_{\mathrm{st},1}>0$ appears (Fig.~\ref{fig:dynamics}C and App.~\ref{app:dynamics-msr}). Fig.~\ref{fig:dynamics}D shows the dynamical phase diagram in the MAP-limit ($\eta\to\infty$): for $\nu<\nu_c(\gamma)$ with $\gamma<1$, training cannot reach the correct equilibrium. In this case, the negative phase is incorrectly estimated during training and the converged model is biased. EBM training dynamics have been studied in the literature \citep{bachtis2024cascade,decelle2021equilibrium,xu2025learning}, but the DMFT above provides, to our knowledge, the first closed large-$N$ description of the coupled dynamics of EBM weights and a persistent negative-phase sampler at finite sampling rate.

\section{Generative metrics in a teacher--student scenario\label{sec:teacher-student}}

We now evaluate the generative performance of an SBM student trained on $K$ samples drawn from a teacher model $\mathbf x^{k}\sim \Pteacher$ \citep{seung1992statistical,engel2001statistical}. We consider two student models. A \emph{typical} student commits to a single weight matrix $W\sim\Prob_\eta(W|\mathcal{D})$ drawn from the posterior, and then generates data from: $\mathbf x\sim \Prob_{W}(\mathbf x)=\Prob(\mathbf{x}|W)$. The \emph{posterior predictive} student instead marginalizes over the weights before generation, effectively drawing samples from an ensemble of trained models weighted by their posterior probabilities: $\mathbf x\sim\PredPost(\mathbf{x}|\mathcal{D}) = \int \Prob(\mathbf{x}|W)\Prob_\eta(W|\mathcal{D})\,\mathrm{d}W$.

\begin{wrapfigure}{R}{0.4\textwidth}
\vspace{-1.5em}
\centering
\includegraphics[width=0.37\textwidth]{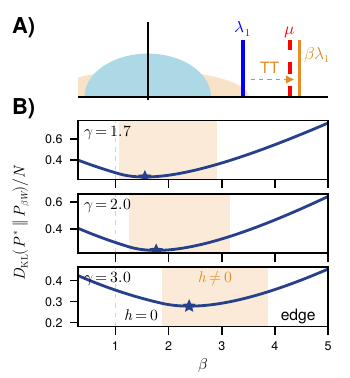}
\vspace{-1em}
\caption{\textbf{Sampling temperature tuning} (TT). \textbf{A)} Mode rescue through temperature tuning, illustrating how the bulk and outlier position change. \textbf{B)} $D_{\mathrm{KL}}(\Pteacher\Vert \Prob_{\beta W})$ vs. $\beta$. %
}
\label{fig:tt-pmo}
\end{wrapfigure}

The quality of a student generated sample $\mathbf x$ is measured by its surprise under the teacher, $-\ln\Pteacher(\mathbf x)$. Averaging over $\mathbf x\sim\Prob_W$ yields the cross-entropy $\mathcal{H}[\Prob_W;\Pteacher]=-\int\Prob_W(\mathbf x)\ln\Pteacher(\mathbf x)\mathrm{d}\mathbf x$. Minimizing this quantity alone, however, selects a singular student placing all its probability mass at the maximizer of $\Pteacher$. Diversity collapse is undesirable for a generative model, so we are naturally led to the reverse Kullback-Leibler (KL) divergence, $D_{\mathrm{KL}}(\Prob_W\|\Pteacher) = \mathcal{H}[\Prob_W;\Pteacher]-\mathcal{H}[\Prob_W]$, where the entropy $\mathcal{H}[\Prob_W]=-\int\Prob_W(\mathbf x)\ln\Prob_W(\mathbf x)\mathrm{d}\mathbf{x}$ penalizes students with poor diversity. Reversing the arguments defines the forward-KL, $D_{\mathrm{KL}}(\Pteacher\|\Prob_W)$. Whereas the reverse-KL is mode-seeking, concentrating student mass where $\Pteacher$ is large, the forward-KL is mode-covering, penalizing regions of $\Pteacher$ the student misses \citep{bishop2006pattern}. Both are widely used to train and evaluate generative models in practice. %
In the large $N$-limit we consider here, they are self-averaging: for any typical realization $W\sim \Prob_{\eta}$, both KLs stay close to their average values, $\left\langle D_{\mathrm{KL}}(\Prob_W\|\Pteacher)\right\rangle_{W\sim \Prob_{\eta}}$ and $\left\langle D_{\mathrm{KL}}(\Pteacher\|\Prob_W)\right\rangle_{W\sim \Prob_{\eta}}$, respectively, so we may omit the average brackets to lighten the notation.

Similarly, the generative performance of the posterior predictive student can be measured by the reverse $D_{\mathrm{KL}}(\PredPost\|\Pteacher)$ or forward $D_{\mathrm{KL}}(\Pteacher\|\PredPost)$. The posterior predictive is popular in Bayesian prediction \citep{aitchison1975goodness,brown2008admissible}, but, with few exceptions \citep{saatchi2017bayesian}, has been less studied in the context of generative modeling, where the focus has been on typical or MAP weights. However, the predictive posterior is expected to outperform typical students: $D_{\mathrm{KL}}(\PredPost\|\Pteacher) \le \left\langle D_{\mathrm{KL}}(\Prob_W\|\Pteacher)\right\rangle_{W\sim \Prob_{\eta}}$ and $D_{\mathrm{KL}}(\Pteacher\|\PredPost) \le \left\langle D_{\mathrm{KL}}(\Pteacher\|\Prob_W)\right\rangle_{W\sim \Prob_{\eta}}$ by convexity (see App.~\ref{app:kl-gaps}). These inequalities justify the interest in the posterior predictive also in a generative setting.

In what follows we consider a simple SBM teacher model with rank-one weights $\Wteacher = \omega^{\ast}\mathbf{w}^{\ast}\mathbf{w}^{\ast\top} - \tfrac{1}{N}\omega^{\ast} I$, where $I$ is the identity matrix, $\mathbf{w}^{\ast}$ is a unit vector and $\omega^{\ast}$ a parameter controlling the signal-to-noise ratio, with $\omega^{\ast}>1$ placing the teacher in a condensed regime. The teacher samples $K=2$ data points which are used to train a student SBM. This is the smallest sample size giving two nonzero covariance modes, one correlated with the teacher direction and the other asymptotically orthogonal to it (App.~\ref{app:teacher-data}), and is sufficient to demonstrate the phenomenology of interest.
The forward and reverse KL divergences, for typical and posterior predictive students, are all computable in closed form in this setting. Each of the phenomena introduced above surfaces most naturally in a different KL, motivating our examination of all four KL variants. We emphasize that both students are constructed from the same tempered posterior~\eqref{eq:tilted-posterior}, based on the regularized log-likelihood; the choice of KL only concerns their evaluation. Detailed calculations are deferred to App.~\ref{app:kl-divergences}.

\textbf{Sampling temperature tuning.}
After training, one can re-introduce a sampling temperature $1/\beta\ne1$. What is the effect of tuning $\beta$ at generation-time? The forward $D_{\mathrm{KL}}(\Pteacher\Vert \Prob_{\beta W})$ is a convex function of $\beta$. Its unique minimizer is $>1$ or $<1$ according to whether the entropy $\mathcal{H}[\Prob_{W}]$ is larger or smaller than the cross-entropy $\mathcal{H}[\Pteacher;\Prob_{W}]$. Intuitively, $\beta>1$ ($<1$) compensates for an excess (lack) of entropy in the trained model.

\begin{wrapfigure}{L}{0.37\textwidth}
\vspace{-2em}
\centering
\includegraphics[width=0.35\textwidth]{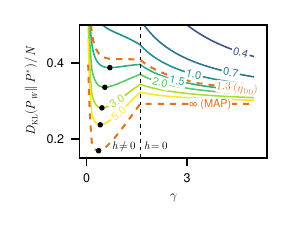}
\vspace{-1.7em}
\caption{\small\textbf{Double descent.} $\eta$ indicated in each curve. Black dots: minima. Orange dashed: thresholds $\eta=\eta_{\mathrm{DD}}$ (upper) and $\eta\to\infty$ (lower). Vertical dashed: boundary between $h\ne0$ and $h=0$ phases.}
\label{fig:double-descent}
\vspace{-0.1em}
\end{wrapfigure}

The model responds in qualitatively different ways to temperature tuning, depending on the starting phase. Furthermore, changing the value of $\beta$ can transport the model from one phase to another. If the trained model starts in the phase $h\ne0,u_{1}=0$ (condensed in a random direction), the optimal $\beta$ is always $<1$, and it is such that the temperature-tuned model lands in an uncondensed phase ($h\to0$). This removes the random condensed direction ($u_1=0$), which results in a lower KL divergence from the teacher $\Pteacher$.

If instead the model is initially in the $h=0,u_{1}\ne0$ phase, the coupling $W$ carries data-aligned modes, but the data signal imprinted in the weights is too weak and samples are swamped by noise. Tuning $\beta>1$ is equivalent to rescaling $W\to\beta W$, which can push the aligned outlier modes beyond the condensation threshold, so that the learned modes become active in generated data, \emph{i.e.}, the temperature-tuned model is transported to a $h,u_{1}\ne0$ phase. This is illustrated in Fig.~\ref{fig:tt-pmo}. See App.~\ref{app:temperature-tuning-theory} for further discussion and derivations.

\begin{wrapfigure}{R}{0.35\textwidth}
  \vspace{-2em}
  \centering
  \includegraphics[width=0.33\textwidth]{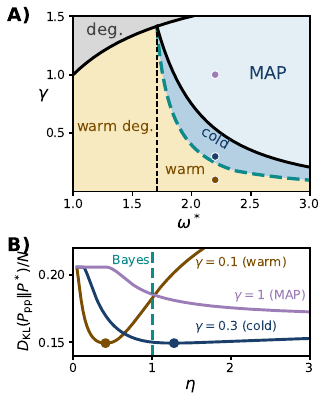}
  \vspace{-1.2em}
  \caption{\small\textbf{Tempered-posterior}.
  \textbf{A)} Phases of the optimal $\eta_{\mathrm{pp}}^{\mathrm{rev}}(\gamma,\omega^{\ast})$. \textbf{B)} Reverse KL vs.\ $\eta$ for selected $\gamma$'s (dots: minima).}
  \label{fig:warm-cold-pp}
  \vspace{-0.5em}
\end{wrapfigure}

\textbf{Double descent.}
Fig.~\ref{fig:double-descent} plots the reverse $D_{\mathrm{KL}}(\Prob_W\|\Pteacher)$ as a function of the regularization strength $\gamma$. If $\eta$ is sufficiently large, it exhibits a double descent. First, for $\gamma\in[0,c_{1}]$, the reverse-KL reaches a minimum (black dots) inside the $h\ne0$ phase, where the student model condenses along a teacher-aligned direction. Beyond the optimal $\gamma$, the student condensation ($h$) weakens and the reverse-KL increases, until it spikes at the condensation threshold $\gamma=c_1$ (vertical dashed line), where the student transitions into the $h=0$ phase. The KL is decreasing throughout this last regime, according to
$\tfrac{1}{N} D_{\mathrm{KL}}(\Prob_W\|\Pteacher) = \tfrac{\omega^{\ast}-1-\ln\omega^{\ast}}{2} + \tfrac{1}{4\gamma\eta}$
(see Eq.~\eqref{eq:reverse-kl-result} for the general expression). The first contribution is an irreducible gap between the teacher and a student uniformly distributed on the sphere, $\tfrac{1}{N}D_{\mathrm{KL}}(\Prob_{W=0}\|\Pteacher)$, while the second term $1/(4\gamma\eta)$ accounts for the random anisotropy of the modes in the bulk of the student weight spectrum.

The double descent behavior is present only in a window of posterior temperatures, $\eta_{\mathrm{DD}}(\omega^{\ast})<\eta<\infty$. The value of $\eta_{\mathrm{DD}}(\omega^{\ast})$ is determined in App.~\ref{app:double-descent-kl} (upper orange dashed curve in Fig.~\ref{fig:double-descent}). Below $\eta_{\mathrm{DD}}(\omega^{\ast})$, the student bulk is over-spread and the reverse-KL is a monotonically decreasing function of $\gamma$, with no minima in the $h\ne0$ phase.  In the opposite MAP limit $\eta\to\infty$ (lower orange dashed in Fig.~\ref{fig:double-descent}), the anisotropy contribution in the $h=0$ phase is sent to zero, collapsing the second descent to a flat line, while preserving the first minimum.

\textbf{Tempered-posterior effects.}
The MAP limit $(\eta\to\infty)$ in our framework corresponds to the standard practice of training a single generative model by minimizing a loss. Few works have attempted to target a Bayesian $(\eta=1)$ posterior predictive distribution \citep{saatchi2017bayesian}. In our solvable setting, the optimal $\eta$ may differ from both endpoints. 
The posterior-predictive KL divergences are not convex functions of $\eta$, and their behavior is consequently more complex than the sampling temperature-tuning scenario discussed above. We focus in this section on the reverse KL, and define an optimal value of $\eta$ minimizing this quantity: $\eta_{\mathrm{pp}}^{\mathrm{rev}}(\gamma,\omega^{\ast}) = \operatorname*{argmin}_{\eta>0} D_{\mathrm{KL}}\left(\PredPost\bigl\Vert\Pteacher\right)$.
The optimal $\eta_{\mathrm{pp}}^{\mathrm{rev}}$ depends on the values of the parameters $\gamma,\omega^{\ast}$. In our teacher-student scenario, small $\gamma$ and/or small $\omega^{*}$ favor $\eta<1$ (warm posterior effect), while larger $\gamma$ and/or larger $\omega^{*}$ favors $\eta>1$ (cold posterior effect).

\begin{wrapfigure}{L}{0.38\textwidth}
  \vspace{-2em}
  \centering
  \includegraphics[width=0.35\textwidth]{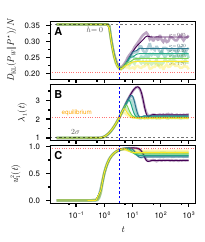}
  \vspace{-1em}
  \caption{\small\textbf{Out-of-equilibrium training.} \textbf{A)} Reverse $D_{\mathrm{KL}}(\Prob_{W(t)}\Vert\Pteacher)$ versus training time $t$, \textbf{B)} $\lambda_{1}(t)$, \textbf{C)} $u_{1}(t)$. Color: $\nu$. Blue dashed: optimal early-stopping time. Orange dashed: equilibrium values (at $\nu\to\infty$).}
  \vspace{-1em}
  \label{fig:ooe-wrap}
\end{wrapfigure}

An optimal posterior temperature different from 1 (the Bayesian prescription) has been referred to as a \emph{cold/warm-posterior effect} \citep{wenzel2020cold,noci2021disentangling,nabarro2022data,pitas2024fine} in supervised learning. Here it arises in an exactly solvable EBM in the high-dimensional limit.

Sweeping $(\omega^{\ast},\gamma)$ partitions the plane into the five regions shown in Fig.~\ref{fig:warm-cold-pp}A, with closed-form thresholds derived in App.~\ref{app:rev-pp-temperature}. In the \emph{warm} phase the optimum lies below the Bayesian value $\eta=1$, because predictions benefit more from averaging across the posterior than from concentrating on its mode. The \emph{cold} phase is the opposite: the posterior is already informative and concentrating on its mode at finite $\eta>1$ outperforms averaging. The \emph{MAP} phase is the extreme limit in which collapsing onto the posterior mode is optimal ($\eta_{\mathrm{pp}}^{\mathrm{rev}}\to\infty$). In the two \emph{deg.} regions, either $h$ or $u_1$ vanishes, and the loss becomes degenerate in $\eta$ over a range of values, making the minimum $\eta_{\mathrm{pp}}^{\mathrm{rev}}$ non-unique. Within \emph{warm deg.}, there is an interval of minimizers which are all $<1$, whereas in \emph{deg.} the reverse-KL becomes independent of $\eta$ so the minimum is completely degenerate. 
Fig.~\ref{fig:warm-cold-pp}B plots the reverse KL as a function of $\eta$ for selected values of $\gamma$, having warm, cold, and MAP minima.
The posterior predictive forward-KL has a similar behavior, with large $\omega^{*},\gamma$ selecting cold/MAP optima $(\eta_{\mathrm{pp}}^{\mathrm{fwd}}>1)$, while small $\omega^{*},\gamma$ select warm optima $(\eta_{\mathrm{pp}}^{\mathrm{fwd}}<1)$. See Fig.~\ref{fig:warm-cold-pp-fwd} in Appendix.

\textbf{Out-of-equilibrium training dynamics.}
The dynamical regimes identified in Sec.~\ref{sec:dynamics} shape the generative quality of the trained model. Fig.~\ref{fig:ooe-wrap} shows example training trajectories at varying sampling rates $\nu$, where we also track the reverse $D_{\mathrm{KL}}(\Prob_{W(t)}\Vert\Pteacher)$ during training (panel A). For $\nu\gg\nu_\mathrm{c}$, the negative phase persistent MC chain equilibrates quickly. The reverse-KL, the top eigenvalue $\lambda_{1}(t)$ and the overlap $u_{1}(t)$ all converge to close to their equilibrium ($\nu\to\infty$, dashed orange) values.

As $\nu$ decreases, $\lambda_{1}(t)$ detaches from the bulk and initially follows the same high-$\nu$ trajectory, but then keeps growing past its equilibrium value, overshooting before slowly relaxing back. The overlap $u_{1}(t)$ stays close to 1 longer than at high $\nu$, then drops to an asymptote below its equilibrium plateau. The KL loss dips almost reaching the equilibrium value, but then increases again and settles at a higher value. The early-time spike dynamics \eqref{eq:lambda-para} can be used to estimate an optimal early-stopping time (dashed blue, Fig.~\ref{fig:ooe-wrap}; see Eq.~\eqref{eq:early-stopping-time}) that is independent of $\nu$. In this setting, early stopping offers no benefit in the equilibrated limit $\nu\to\infty$.

\section{Validation of phenomenology in standard generative architectures}

We next test whether the four effects found above in the SBM also occur in standard architectures (Table~\ref{tab:experiment-summary}). Although some have been reported previously, they remain poorly understood in generative models. In particular, posterior tempering is little explored in generative models to the best of our knowledge.

We first consider small BMs and RBMs with exactly computable partition functions and numerically estimated posterior averages, trained on spiked teachers as in Sec.~\ref{sec:teacher-student}. In the RBM, we verify beneficial warming and cooling of both sampling and posterior temperatures, together with nonmonotonic typical reverse KL as regularization varies. Slower persistent sampling amplifies weight modes and causes alignment to decline after an initial rise, favoring early stopping over continued training; matched controls separate infrequent-refresh effects from an additional Gibbs contribution. These finite-size experiments do not determine SBM phase boundaries. Full results are reported in Apps.~\ref{app:experiments-small-bm} and~\ref{app:experiments-small-rbm}.

We next consider larger models. The relevant posterior averages and observables are difficult to evaluate jointly in high-dimensional models, and we could not probe all four effects in a single large model. We instead use complementary architectures, each suited to probing one or more aspects of the phenomenology. Although we do not claim quantitative agreement, and other factors may play a role in larger models, these experiments provide qualitative evidence that the four effects recur beyond the SBM and, where spectral observables are accessible, reveal signatures compatible with the SBM mechanism. Details are in App.~\ref{app:comp-exp}. Briefly,

\textbf{Sampling temperature tuning.} We trained Potts Boltzmann machines on protein sequences from Pfam \citep{weigt2009identification,marks2011protein,hopf2017mutation}, at various hyperparameter settings. In sufficiently regularized models, the learned coupling matrix contains modes aligned with the top covariance modes of the data that stay approximately silent in the generated data, reminiscent of the $h=0,u_1\ne0$ phase in the SBM. Tuning $\beta>1$ rescues the silent modes in generated data, in a manner compatible with the mechanism depicted in Fig.~\ref{fig:tt-pmo}. See App.~\ref{app:temperature-tuning}.

\textbf{Double descent.} 
We confirmed double descent as a function of regularization in two generative architectures: a Gaussian--Bernoulli restricted Boltzmann machine (RBM) trained on a financial dataset~\citep{fama1997french,laloux1999noise}, and autoregressive belief networks trained on samples from the 2D Ising and Curie--Weiss models. See App.~\ref{app:double-descent-EBM}.

\textbf{Tempered-posterior effects.} To the best of our knowledge, warm/cold posterior effects have not been studied in generative models. We trained Bayesian GANs \cite{saatchi2017bayesian} on synthetic datasets, and demonstrate that the optimal posterior temperature can be $<1$ or $>1$, depending on the regularization, as in the SBM. See App.~\ref{app:tempered-posterior-gen}.

\textbf{Out-of-equilibrium training dynamics.} We trained Potts Boltzmann machines on protein sequence alignments at varying sampling rates $\nu$ during training, observing overshooting of the top weight modes, with data overlaps that converge to sub-optimal values, reproducing the qualitative behavior of the SBM. See App.~\ref{app:ooe-potts}.

\section{Conclusion, limitations, and broader impact}
\label{sec:discussion}

Solvable models provide guidance to navigate complex phenomena in machine learning \citep{simon2026scientifictheorydeeplearning}. We introduced the SBM as a solvable EBM, derived its equilibrium phase diagram, solved its Langevin training dynamics via DMFT, and obtained teacher--student generative metrics in the high-dimensional limit.
Standard generative architectures reproduce qualitatively the predicted temperature tuning, double-descent as a function of regularization strength, tempered-posterior effects, and out-of-equilibrium training behavior. These effects depend critically on the spherical constraint of the SBM, which drives the condensation transition, and would not occur in a multivariate Gaussian model (see App.~\ref{app:comparison-gaussian} for further discussion).

\textbf{Limitations.}
The SBM is a minimal solvable EBM rather than a realistic generative architecture.
Our analysis assumes effectively low-dimensional data (App.~\ref{app:ooe-low-dim-data}).
Our derivations are heuristic in the statistical-physics sense (saddle-point evaluation of the Coulomb gas, MSR/DMFT closure of the Langevin dynamics) rather than rigorous proofs; results are expected to be asymptotically exact in the $N\to\infty$ limit, and we verify them against finite-$N$ simulations throughout.

\textbf{Broader impact.}
This is a theoretical paper on the statistics of generative models; it does not propose new models or release datasets, and we do not see a direct avenue for societal harm. %

\newpage
\begin{ack}
We acknowledge support from ANR grants MEMNET, ANR-24-CPJ1-0172-01 and ANR-25-CE45-5343 ``MANUFACTURE''. We used HPC resources from GENCI–IDRIS, Grant 2025-AD011015768R1.
\end{ack}

\small
\bibliographystyle{plainnat-lastfirst}
\bibliography{refs}
\normalsize

\newpage
\appendix

\etocdepthtag.toc{appendix}

{\etocsettagdepth{mainmatter}{none}%
\etocsettagdepth{appendix}{subsection}%
\etocsetnexttocdepth{subsection}%
\etocsettocstyle{\section*{Appendix contents}}{}%
\tableofcontents}

\medskip

\section{Equilibrium phase diagram\label{app:equilibrium}}

\subsection{Large $N$ asymptotic expressions}

We consider an asymptotic expansion of the evidence at large $N$,
\begin{equation}
\!\!\!\!\mathcal{Y}(C) = \int \exp\left\{
-\tfrac{N}{4}\gamma\eta\Trace\Lambda^2 - \eta\Trace(C) \ln Z(\Lambda) + \ln\hciz(\Lambda;C) + \sum_{i<j}\ln|\lambda_i-\lambda_j|
\right\} \mathrm{d}\Lambda.
\end{equation}
Note that if $\Trace(C)=K$ we recover the definition used in the main text. Here it will be convenient to consider a generalization where $C$ is not required to satisfy this constraint.

The dominant terms are $\mathcal O(N^{2})$ and can be written:
\begin{equation}\label{eq:lnY-N2}
\frac{\ln\mathcal{Y}(C)}{N^{2}} \sim 
-\frac{\gamma\eta}{4}\int\lambda^{2}\rho(\lambda)\mathrm{d}\lambda + \frac{1}{2}\int\ln|\lambda-\lambda'|\rho(\lambda)\rho(\lambda')\mathrm{d}\lambda\mathrm{d}\lambda' \sim 
-\frac{3}{8} + \frac{1}{2}\ln\sigma
\end{equation}
up to terms of lower-order, where $\rho(\lambda)$ is given by Eq.~\eqref{eq:rho}, with $\sigma=1/\sqrt{\gamma\eta}$. Thus the leading $\BigO(N^{2})$ contribution comes from standard Coulomb-gas terms. Extremization of ~\eqref{eq:lnY-N2} with respect to $\rho$ results in the Wigner semi-circle law ~\eqref{eq:rho}, that describes the bulk of eigenvalues (substituting Eq. \eqref{eq:rho} gives the evaluated result on the right-hand side of Eq. \eqref{eq:lnY-N2}). It is convenient to define:
\begin{equation}\label{eq:Stieltjes}\begin{aligned}
\Stieltjes(z) &= \int\frac{\rho(\lambda)\mathrm{d}\lambda}{z-\lambda}
= \frac{1}{2\sigma} \left( \frac{z}{\sigma} - \sqrt{ \frac{z^{2}}{\sigma^{2}} - 4 } \right),
\\
\LogStieltjes(z) &= \int\rho(\lambda)\ln(z-\lambda)\mathrm{d}\lambda
= \frac{\sigma^{2}}{2}G^{2}(z) - \ln G(z),
\\
\StieltjesB(z) &= \int\frac{\lambda}{z-\lambda}\rho(\lambda)\mathrm{d}\lambda = \frac{z}{2\sigma}\left( \frac{z}{\sigma} - \sqrt{ \frac{z^{2}}{\sigma^{2}}-4 } \right)-1 = z\Stieltjes(z)-1.
\end{aligned}\end{equation}
We will use the notation $g_{k}=\Stieltjes(\lambda_{k})$.

To capture the phenomenology of the outliers, it is important to consider the $\mathcal{O}(N)$ contribution. We write $\ln Y=N^{2}(\dots) + N\Phi(C) + \BigO(1)$. Once the $K$ outliers are separated, the bulk integral runs over $M=N-K$ eigenvalues, and expanding its Coulomb-gas free energy in $K/N$ produces an $\BigO(N)$ boundary correction $\tfrac{K}{2}\ln(\gamma\eta)$ (an accompanying additive $K/2$ is absorbed into the HCIZ normalization convention used below). We retain the Gaussian base-measure constant in $Z(\Lambda)$. Collecting all order-$N$ contributions,
\begin{equation}\label{eq:Phi1-def}
\Phi(C) = \frac{K}{2}\ln(\gamma\eta) + \sum_{i=1}^{K}
\left(- \frac{\gamma\eta}{4}\lambda_{i}^{2} + \LogStieltjes(\lambda_{i})\right) - \frac{\eta}{N}\Trace(C)\ln Z(\Lambda) + \frac{1}{N}\ln\hciz(\Lambda;C)
\end{equation}
In the following sections we evaluate $\ln Z(\Lambda)$ and $\ln\hciz(\Lambda;C)$ to leading order in $N$.

\subsubsection{Evaluation of $Z(W)$ and condensation phase transition}

Following standard steps (\emph{e.g.} Ch. 5 in \citep{Baxter1990}):
\begin{equation}\label{eq:Gaussian-Fourier}
Z \propto
\int_{-\infty}^{\infty} \mathrm{d}\mathbf{x} \int_{c-i\infty}^{c+i\infty}\mathrm{d}\mu
\exp\left( \frac{1}{2}\sum_{i=1}^{N}(\lambda_{i}-\mu)x_{i}^{2} + \frac{N}{2}\mu \right)
\end{equation}
where $\mu$ traverses a line in the complex plane parallel to the imaginary axis with constant real part, $\Re\mu=c$ larger than all eigenvalues. In anticipation of the condensation along the top modes \citep{kosterlitz1976spherical,steinberg2024replica}, we introduce order parameters $\mathbf{h}=(h_1,\dots,h_K)$ with $h_i=x_i/\sqrt{N}$, along the top $K$ modes. From \eqref{eq:Gaussian-Fourier},
\begin{align}\label{eq:Z-Fourier}
Z &\propto
\int \mathrm{d}h_{1}\dots\mathrm{d}h_{K}\,
\mathrm{d}x_{K+1}\dots\mathrm{d}x_{N} \, \mathrm{d}\mu \\
&\qquad\exp\left(
\frac{N}{2}\sum_{i=1}^{K}(\lambda_{i}-\mu)h_{i}^{2} + \frac{1}{2}\sum_{i=K+1}^{N}(\lambda_{i}-\mu)x_{i}^{2} + \frac{N\mu}{2} \right) \\
&\propto \int \mathrm{d}h_{1}\dots\mathrm{d}h_{K}\,\mathrm{d}\mu \,
\exp\left\{ \frac{N}{2}\left(
\sum_{i=1}^{K}(\lambda_{i}-\mu)h_{i}^{2} -
\LogStieltjes(\mu) + \mu \right) \right\}.
\end{align}
For large $N$, the integrations over $h_1,\dots,h_K$ and $\mu$ concentrate at a saddle-point, satisfying the stationarity conditions:
\begin{equation}
1 - h^{2} = \Stieltjes(\mu),\quad
(\lambda_{i} - \mu) h_{i} = 0
\end{equation}
where $h^{2}=\sum_{i=1}^{K}h_{i}^{2}$. The solution of these equations can be written
\begin{equation}\label{eq:SBM-m2-mu}
h^{2} = \begin{cases}
0, \\
1-g_{1},
\end{cases}\quad
\mu=\begin{cases}
 \Stieltjes^{-1}(1) &\quad\text{if } g_{1} \ge 1, \\
 \lambda_{1} &\quad\text{if } g_{1} \le 1,
\end{cases}
\end{equation}
where $z=\Stieltjes^{-1}(a)$ for $a>0$ denotes the unique root of the equation $\Stieltjes(z)=a$ satisfying $z\geq\lambda_{K+1}$.

Evaluating \eqref{eq:Z-Fourier} at the saddle point yields
\begin{equation}\label{eq:SBM-log-Z}
\frac{1}{N}\ln Z \sim \frac{\ln(2\pi)+\mu-\LogStieltjes(\mu)}{2}
=\frac{1}{2}\ln(2\pi)+\begin{cases}
\frac{1}{2}+\frac{1}{4\gamma\eta} & g_1\ge1, \\
\frac{1}{2}\left(\frac{g_{1}}{\gamma\eta}+\frac{1}{g_{1}}-\frac{g_{1}^{2}}{2\gamma\eta}+\ln g_{1}\right) & g_1\le1.
\end{cases}
\end{equation}
to leading order in $N$, where the additive $\tfrac{1}{2}\ln(2\pi)$ comes from the Gaussian base measure of \eqref{eq:Gaussian-Fourier} and is the constant carried explicitly by the teacher partition function \eqref{eq:lnZ-and-entropy-teacher-sbm-rank-1}.

To compute the \textbf{average energy} we use:
\begin{equation}\label{eq:SBM-avg-energy} 
\left\langle \langle \Energy(\mathbf{x};W) \rangle_{\mathbf x\sim\Prob_{W}} \right\rangle_{W\sim \Prob_\eta}
=
-\left\langle \left.\frac{\partial\ln Z(\beta W)}{\partial\beta}\right|_{\beta=1} \right\rangle_{W\sim \Prob_\eta}
= \frac{N}{2}(1-\mu)
\end{equation}
to leading order in $N$. The \textbf{entropy} of the SBM is given by,
\begin{equation}\label{eq:SBM-entropy}
\left\langle \mathcal{H}[\Prob_{W}] \right\rangle_{W\sim \Prob_\eta}
=
\left\langle \langle \Energy(\mathbf{x};W) \rangle_{\mathbf x\sim\Prob_{W}} + \ln Z(W) \right\rangle_{W\sim \Prob_\eta}
= \frac{N}{2}[\ln(2\pi e) - \LogStieltjes(\mu)]
\end{equation}

\subsubsection{The real spherical HCIZ integral and its large-$N$ asymptotics\label{app:hciz-asymptotics}}

The real (orthogonal-group) spherical HCIZ integral is defined as
\begin{equation}\label{eq:hciz}
\hciz(\Lambda;C) = \int_{V\in\orthogonalgroup(N)} \mathcal{D}V \exp\!\left(\tfrac{N\eta}{2}\Trace CV\Lambda V^{\top}\right).
\end{equation}
where $\orthogonalgroup(N)$ is the group of $N\times N$ orthogonal matrices and $\mathcal{D}V$ is the flat Haar measure on $\orthogonalgroup(N)$. For $K$ fixed while $N\to\infty$, its leading-$\BigO(N)$ asymptotics \citep{guionnet_husson_2022} take the form
\begin{equation}\label{eq:hciz-N}
\ln\hciz(\Lambda;C) \underset{N\to\infty}{\sim}
\frac{N}{2} \sum_{k=1}^{K}
\left\{ \eta c_{k}(\lambda_{k}u_{k}^{2} + \chi_{k} - \chi_{k}u_{k}^{2}) - \LogStieltjes(\chi_{k}) - \ln \eta c_{k} \right\},
\end{equation}
with saddle-point variables $(u_k^2,\chi_k)$ obtained by extremization of this expression. The stationarity conditions result in two branches:
\begin{equation}\label{app:eq:hciz-branches}
\chi_{k}=\begin{cases}
    \lambda_{k}, \\
    \Stieltjes^{-1}(\eta c_{k}),
\end{cases}\quad
u_{k}^{2}=\begin{cases}
    1-g_{k}/(\eta c_{k}) & \quad\text{if } g_{k}\le\eta c_{k},\\
    0 & \quad\text{if } g_{k}\ge\eta c_{k}.
\end{cases}
\end{equation}
Substituting the stationary values, \eqref{eq:hciz-N} simplifies to
\begin{equation}\label{eq:hciz-N-extremized}
\ln\hciz(\Lambda;C) \underset{N\to\infty}{\sim} \frac{N}{2} \sum_{k=1}^{K}
\left\{ \eta c_{k}\chi_{k} - \LogStieltjes(\chi_{k}) - \ln \eta c_{k} \right\}.
\end{equation}
These asymptotics also cover coincident limiting outlier positions, by the continuity and uniform-approximation results of \citep{guionnet_husson_2022}.

\subsubsection{Evaluated form of $\Phi$}\label{app:evidence}

Substituting the saddle values of $\ln Z(\Lambda)$~\eqref{eq:SBM-log-Z} and $\ln\hciz(\Lambda;C)$~\eqref{eq:hciz-N} into~\eqref{eq:Phi1-def} and extremizing over $\chi_{k}$ gives $\Phi(C)$ as a closed-form function of the data spectrum. For a source $C$ with eigenvalues $c_{1},\dots,c_{K}$ ($K=\Rank C$, $\Trace C=\sum_{k}c_{k}$),
\begin{equation}\label{eq:Phi1-eval}
\Phi(C) = \frac{K}{2}\ln(\gamma\eta)
- \frac{\eta}{2}[\ln(2\pi)+\mu-\LogStieltjes(\mu)]\Trace C
+ \sum_{k=1}^{K}\psi_{k}
,
\end{equation}
where the per-mode contribution $\psi_{k}$ is
\begin{align}\label{eq:hk-def}
\psi_{k} &= -\frac{\gamma\eta}{4}\lambda_{k}^{2} + \LogStieltjes(\lambda_{k}) + \frac{\eta c_{k}\chi_{k} - \LogStieltjes(\chi_{k}) - \ln(\eta c_{k})}{2} \\
&= \begin{cases}
\frac{\eta c_{k}^{2}}{4\gamma}-\frac{1}{2}\ln(\gamma\eta), & \text{uncondensed }(k>d), \\
-\frac{1}{2}-\frac{\gamma\eta}{4g_{1}^{2}}+\frac{c_{k}g_{1}}{2\gamma}+\frac{\eta c_{k}}{2g_{1}}-\frac{\ln(g_{1}\,\eta c_{k})}{2}, & \text{condensed }(k\le d), \\
\end{cases}
\end{align}
with $\lambda_{k}$ and $\chi_{k}$ determined by the phase diagram. The first row applies to a mode in uncondensed position ($\lambda_{k}$ either at the bulk edge or detached at $1/(\eta c_{k})+c_{k}/\gamma$). The second row applies to aligned condensed modes with $\lambda_{k}=\chi_{k}=\lambda_{1}$ and $g_{1}=\Stieltjes(\lambda_{1})$; it spans the edge sub-phase ($g_{1}=\sqrt{\gamma\eta}$, $\lambda_{1}$ pinned at the bulk edge) and outlier sub-phase ($g_{1}<\sqrt{\gamma\eta}$, $\lambda_{1}$ detached at $1/g_{1}+g_{1}/(\gamma\eta)$) of the $h,u_1\ne 0$ phase continuously.

\subsection{Optimal outlier positions and general structure of the phase diagram}\label{app:phase-conditions}%

We now optimize \eqref{eq:Phi1-def} with respect to the outlier eigenvalue positions. The optimal positions of $\lambda_{1},\dots,\lambda_{K}$ must be stable under small perturbations $\lambda_{1}+\epsilon_{1},\dots,\lambda_{K}+\epsilon_{K}$. 

Suppose a stable configuration has $\lambda_{1}=\dots=\lambda_{d}$ of degeneracy $d$, while $\lambda_{d}>\lambda_{d+1}$. To first-order in the perturbations $\epsilon_{k}$, the variation of the objective is:
\begin{equation}
-\frac{\eta}{2}h^{2}\epsilon_{*}\Trace C + \sum_{k}\left\{ -\frac{\gamma\eta}{2}\lambda_{k} + \frac{\eta}{2}c_{k}u_{k}^{2} + g_{k} \right\}\epsilon_{k},
\end{equation}
where $\epsilon_{*}=\max\{\lambda_{1}+\epsilon_{1},\dots,\lambda_{d}+\epsilon_{d}\}-\lambda_{1}\le\max\{\epsilon_{1},\dots,\epsilon_{d}\}$ (the inequality follows from $\lambda_{1}\ge\lambda_{k}$ for all $k$ and it is proved below). After some algebra,
\begin{equation}
-\frac{\eta\Trace C}{2}h^{2}\epsilon_{*}
+ \sum_{k\leq d}\left\{ -\frac{\gamma\eta}{2g_{1}} + \frac{1}{2}\eta c_{k}u_{k}^{2} + \frac{g_{1}}{2} \right\}\epsilon_{k}
+ \sum_{k>d}\left\{ -\frac{\gamma\eta}{2g_{k}} + \frac{1}{2}\eta c_{k}u_{k}^{2} + \frac{g_{k}}{2} \right\}\epsilon_{k}\,.
\end{equation}
The $\epsilon_{k}$ for $k>d$ can be varied independently. They are free if $\lambda_{k}>\frac{2}{\sqrt{ \gamma\eta }}$, in which case we must have $-\frac{\gamma\eta}{2g_{k}} + \frac{1}{2}\eta c_{k}u_{k}^{2} + \frac{g_{k}}{2}=0$. On the other hand if $\lambda_{k}=\frac{2}{\sqrt{ \gamma\eta }}$, we can only perturb $\epsilon_{k}\geq0$, so we only require $-\frac{\gamma\eta}{2g_{k}} + \frac{1}{2}\eta c_{k}u_{k}^{2} + \frac{g_{k}}{2}\leq0$. Solving these conditions gives the positions of uncondensed eigenvalues (\emph{i.e.}\ $\lambda_{k}<\mu$ for $k>d$):
\begin{equation}
\begin{aligned}
\lambda_{k} &= \frac{2}{\sqrt{\gamma\eta}}, &
g_{k} &= \sqrt{\gamma\eta}, &
u_{k}^{2} &= 0, &
\text{ if }\eta c_{k}\leq\sqrt{\gamma\eta},
\\
\lambda_{k} &= \frac{1}{\eta c_{k}}+\frac{c_{k}}{\gamma}, &
g_{k}&=\frac{\gamma}{c_{k}}, &
u_{k}^{2}&=1-\frac{\gamma}{\eta c_{k}^{2}}, &
\text{ if }\eta c_{k} \geq \sqrt{\gamma\eta}.
\end{aligned}
\end{equation}
Now we focus on the condensed part, $k\leq d$. If $h^{2}=0$ we have the same conditions as before and these eigenvalues are also in their uncondensed positions. So let's assume we are in the $h\ne 0$ phase with $h^{2}>0$. We then have $h^{2}=1-g_{1}$ and $u_{k}^{2}=1-\frac{g_{1}}{\eta c_{k}}$, and in this case $\lambda_{1}$ is an outlier so $\epsilon_{1},\dots,\epsilon_{d}$ are free. Then
\begin{equation}
-\frac{\eta\Trace C}{2}h^{2}\epsilon_{*} + \sum_{k\leq d}\left\{ -\frac{\gamma\eta}{2g_{1}} + \frac{\eta}{2}c_{k}u_{k}^{2} + \frac{g_{1}}{2} \right\}\epsilon_{k}
\end{equation}
simplifies to
\begin{equation}
-\eta\epsilon_{*}(1-g_{1})\Trace C - \frac{\gamma\eta}{g_{1}}\sum_{k\leq d}\epsilon_{k} + \eta\sum_{k\leq d}c_{k}\epsilon_{k}.
\end{equation}
Note that $\eta c_{k}\geq \frac{\gamma\eta}{g_{1}}$ for $k\leq d$, because these eigenvalues are outliers that have been pushed away from their uncondensed positions by $\mu$ towards the bulk. Therefore
$$- \frac{\gamma\eta}{g_{1}}\sum_{k\leq d}\epsilon_{k} + \eta\sum_{k\leq d}c_{k}\epsilon_{k} \geq 0,$$
so the best perturbation we can do (in the sense of trying to maximize the objective) is to set all $\epsilon_{k}=\epsilon_{*}$ equal to the maximum. Then, setting $\epsilon_{1}=\dots=\epsilon_{d}$ all equal, gives the condition
\begin{equation}
-(1-g_{1})\Trace C - \frac{d\gamma}{g_{1}} + \sum_{k\leq d}c_{k} = 0
\end{equation}
that solves for the optimal position of the coalesced eigenvalues (given in terms of their Stieltjes transform $g_{1}$):
\begin{equation}\label{eq:g1-coalesced}
g_{1} = \frac{1}{2\Trace C}\left\{ \sum_{k=d+1}^{K}c_{k} + \sqrt{ \left( \sum_{k=d+1}^{K}c_{k} \right)^{2} + 4 d\gamma \Trace C } \right\}.
\end{equation}

\paragraph{An auxiliary inequality.}
The first-order perturbation analysis above invokes the bound $\epsilon_{*}\le\max_{k}\epsilon_{k}$ on the quantity
\begin{equation}
\epsilon_{*}=\max\left\{ \lambda_{1}+\epsilon_{1},\dots,\lambda_{K}+\epsilon_{K} \right\}-\lambda_{1},
\end{equation}
where $\lambda_{1}=\max\{\lambda_{1},\dots,\lambda_{K}\}$ because the eigenvalues are sorted in decreasing order. Setting $j=\operatorname{argmax}_{k}\{\lambda_{k}+\epsilon_{k}\}$,
\begin{equation}
\max_{k} \{ \lambda_{k} + \epsilon_{k} \} = \lambda_{j} + \epsilon_{j}
\le \lambda_{1} + \epsilon_{j}
\le \lambda_{1} + \max_{k}\epsilon_{k},
\end{equation}
and subtracting $\lambda_{1}$ from both sides yields $\epsilon_{*} \le \max_{k}\epsilon_{k}$ as claimed.

\subsubsection{Concavity and uniqueness of the equilibrium saddle}\label{app:saddle-unique}

The previous argument establishes a saddle-point of the action that determines the positions of the outlier eigenvalues. However, this does not rule out the existence of competing saddle-points. In this section, we show by a concavity argument that the saddle-point we described above is in fact the unique global maximum.

After fixing the semicircular bulk, the remaining order-$N$ variational problem can be written, up to terms independent of the saddle variables, as
\begin{equation}
\Phi(C) =
\max_{\mu \geq \lambda_1 \geq \cdots \geq \lambda_K \geq 2\sigma} \mathcal{J}(\mu,\lambda_1,\ldots,\lambda_K) + \mathrm{const.},
\label{eq:reduced-evidence-variational}
\end{equation}
where
\begin{equation}
\mathcal{J}(\mu,\lambda_1,\ldots,\lambda_K)
=
-\frac{\eta}{2}\operatorname{tr}C
\bigl[\mu-F(\mu)\bigr]
+\sum_{k=1}^{K}\psi_k(\lambda_k).
\label{eq:reduced-evidence-objective}
\end{equation}
Here $\psi_k(\lambda_k)$ is defined in Eq.~\eqref{eq:hk-def}, with $\chi_k$ set to its saddle value in Eq.~\eqref{app:eq:hciz-branches}, before optimizing over $\lambda_k$.

We now show that $\mathcal{J}$ is strictly concave. From Eq.~\eqref{eq:Stieltjes},
\begin{equation}
F'(z)=G(z),
\qquad
G'(z)= -\int \frac{\rho(x)}{(z-x)^2}\,dx<0,
\qquad
z>2\sigma .
\label{eq:stieltjes-monotonicity}
\end{equation}
The dependence on $\mu$ therefore satisfies
\begin{equation}
    \frac{\partial \mathcal{J}}{\partial\mu}
    =
    -\frac{\eta}{2}\operatorname{tr}C\,
    \bigl[1-G(\mu)\bigr],
    \qquad
    \frac{\partial^2 \mathcal{J}}{\partial\mu^2}
    =
    \frac{\eta}{2}\operatorname{tr}C\,G'(\mu)
    <0.
    \label{eq:mu-concavity}
\end{equation}
The constrained maximization over $\mu\geq\lambda_1$ consequently gives $G(\mu)=1$ when the unconstrained maximum is feasible, and $\mu=\lambda_1$ otherwise, reproducing Eq.~\eqref{eq:SBM-m2-mu}.

For the outlier contribution, using $g_k=G(\lambda_k)$ and the two HCIZ branches in Eq.~\eqref{app:eq:hciz-branches}, one obtains
\begin{equation}
\psi_k'(\lambda_k)
=
\begin{cases}
\displaystyle
-\frac{\gamma\eta}{2}\lambda_k+g_k,
    & g_k\geq \eta c_k,
    \\[2mm]
\displaystyle
-\frac{\gamma\eta}{2}\lambda_k
+\frac{g_k+\eta c_k}{2},
    & g_k\leq \eta c_k .
\end{cases}
\label{eq:psi-first-derivative}
\end{equation}
Equivalently, the derivative of the optimized HCIZ contribution is
zero on the $u_k=0$ branch and $(\eta c_k-g_k)/2$ on the $u_k\neq0$
branch. Away from the switching point,
\begin{equation}
    \psi_k''(\lambda_k)
    =
    \begin{cases}
    \displaystyle
    -\frac{\gamma\eta}{2}+G'(\lambda_k),
        & g_k>\eta c_k,
        \\[2mm]
    \displaystyle
    -\frac{\gamma\eta}{2}
    +\frac{1}{2}G'(\lambda_k),
        & g_k<\eta c_k ,
    \end{cases}
    \qquad <0.
    \label{eq:psi-concavity}
\end{equation}
At $g_k=\eta c_k$, the two expressions in
Eq.~\eqref{eq:psi-first-derivative} coincide. Thus
$\psi_k'(\lambda_k)$ is continuous and strictly decreasing throughout
$\lambda_k\geq2\sigma$, and $\psi_k$ is globally strictly concave,
including across the alignment transition.

It follows that $\mathcal{J}$ is strictly concave on the convex domain
\begin{equation}
\mu\geq\lambda_1\geq\cdots\geq\lambda_K\geq2\sigma.
\end{equation}
Moreover, the objective is coercive and its maximum is attained. Strict concavity then implies that this maximizer is unique. In particular, every feasible solution of the corresponding stationarity and complementarity conditions is necessarily the global maximizer.

\subsubsection{Phase diagram}\label{app:phase-conditions-general}

The different phases are given by:

\begin{itemize}
\item $h=u_{1,\dots,K}=0$. If $\gamma\eta>1$ and $\gamma>\eta c_{1}^{2}$. The eigenvalues stick to the edge, $\lambda_{1}=\dots=\lambda_{K}=2\sigma$.
    
\item $h=0$, $u_{1,\dots,a}\ne0$. If $\gamma>c_{1}$, $\gamma\eta>1$, and $\eta c_{a+1}^{2}<\gamma<\eta c_{a}^{2}$ for some $a\in[K]$. The top $a$ eigenvalues detach from the bulk at 
\begin{equation}\label{eq:lambda-PMo}
\lambda_{k}=\frac{1}{\eta c_{k}}+\frac{c_{k}}{\gamma},\qquad u_{k}^{2}=1-\frac{\gamma}{\eta c_{k}^{2}}\qquad(1\le k \le a),
\end{equation}
while $\lambda_{a+1}=\dots=\lambda_{K}=2\sigma$ stick to the edge with $u_{a+1}=\dots=u_{K}=0$.

\item $h\ne0$, $u_{1,\dots,K}=0$. If $\eta^{2}c_{1}^{2}<\gamma\eta<1$, then $h^{2}=1-g_{1}$. All eigenvalues stick to the edge, $\lambda_{1}=\dots=\lambda_{K}=2\sigma$.

\item $h\ne0$, $u_{1,\dots,d}\ne0$ (edge). If
\begin{equation}\label{eq:FMe-conditions}
\eta c_{d+1}<\sqrt{\gamma\eta}<\min\{1,\eta c_{d}\}
\quad \text{and} \quad
\sqrt{\gamma\eta}\,(\eta\Trace C-d)+\eta\sum_{k=1}^{d}c_{k}<\eta\Trace C,
\end{equation}
for some $d\in[K]$ (taking $c_{d+1}=0$ if $d=K$), then all eigenvalues stick to the edge, $\lambda_{1}=\dots=\lambda_{K}=2\sigma$, the top $d$ of them align with the data,
\begin{equation}
u_{k}^{2}=1-\frac{1}{c_{k}}\sqrt{\frac{\gamma}{\eta}}\quad(k\le d),
\end{equation}
and $h\ne0$.

\item $h\ne0$, $u_{1,\dots,d}\ne0$ (outlier). If
\begin{equation}\label{eq:FMo-conditions}
\frac{\gamma}{c_{d}}<g_{1}<\min\!\left\{1,\,\eta c_{d},\,\frac{\gamma}{c_{d+1}},\,\sqrt{\gamma\eta}\right\},
\end{equation}
for $d\in[K]$
(with $c_{d+1}=0$ when $d=K$), with $g_{1}$ given by Eq.~\eqref{eq:g1-coalesced}, the top $d$ eigenvalues coalesce at a common outlier position obtained by inverting $g_{1}=\Stieltjes(\lambda_{1})$. Then $h^{2}=1-g_{1}$ and $u_{k}^{2}=1-g_{1}/(\eta c_{k})$ for $k\le d$. For the remaining modes $k>d$, if $\eta c_{k}^{2}\ge\gamma$, then $\lambda_{k}=1/(\eta c_{k})+c_{k}/\gamma$ and $u_{k}^{2}=1-\gamma/(\eta c_{k}^{2})$; otherwise, $\lambda_{k}=2/\sqrt{\gamma\eta}$ and $u_{k}^{2}=0$.
\end{itemize}

\begin{figure}
\centering
\includegraphics[width=\textwidth]{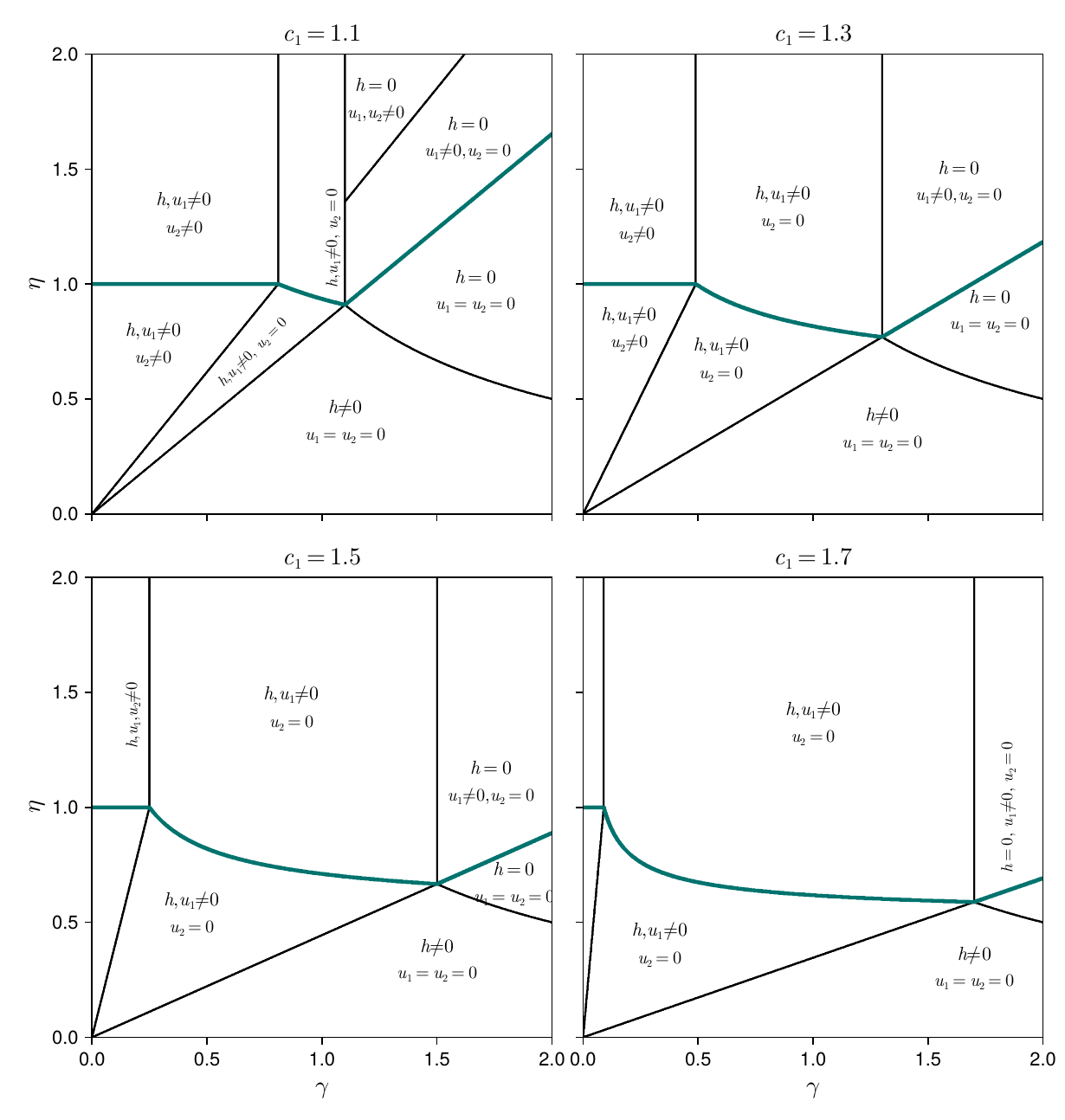}
\caption{\textbf{$K=2$ phase diagrams.} Phase diagram in the $(\gamma,\eta)$ plane for $c_1\in\{1.1,1.3,1.5,1.7\}$ (with $c_2=2-c_1$). Black lines are phase boundaries; the teal line separates outlier from edge phases.}
\label{fig:phase-diagram-k=2-appendix}
\end{figure}

Figure~\ref{fig:phase-diagram-k=2-appendix} shows how the $K=2$ phase diagrams deform as $c_{1}$ varies across the allowed range $1<c_{1}<2$. Near the symmetric limit $c_{1}=c_{2}=1$ the $h\ne0$ regions with $d=2$ (both edge and outlier) are largest; as $c_{1}$ increases toward $2$ these regions shrink and the outlier--edge boundary (teal) flattens.

\subsubsection{Finite-size equilibrium validation}\label{app:finite-N-equilibrium}

\begin{figure}
\centering
\includegraphics[width=5.5in]{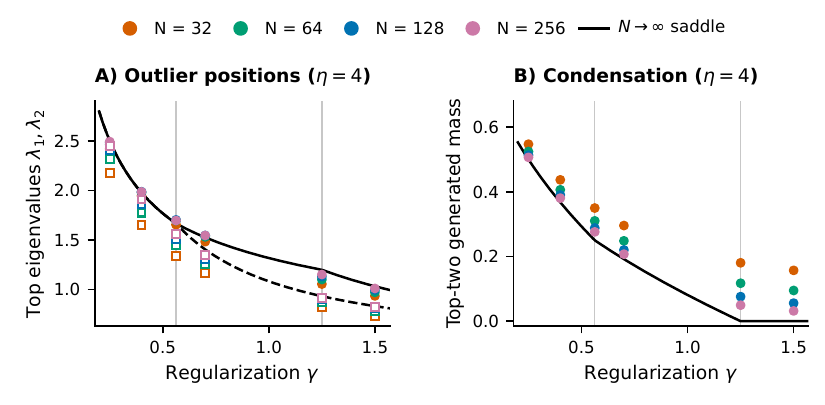}
\caption{\textbf{Finite-size equilibrium validation.} We sample the finite-$N$ posterior with $K=2$ and $(c_1,c_2)=(1.25,0.75)$. Colors denote $N=32,64,128,256$ and bars show $95\%$ Monte Carlo intervals. \textbf{A)} Top eigenvalues at $\eta=4$ (filled circles: $\lambda_1$; open squares: $\lambda_2$); black solid and dashed curves give the saddle values of $\lambda_1$ and $\lambda_2$. \textbf{B)} Generated mass $\sum_{j=1}^2\langle(\mathbf x\cdot\mathbf v_j/N)^2\rangle$ compared with the limiting $h^2$. The uncondensed modes contribute a nonzero finite-size baseline. Gray vertical lines mark the coalescence and condensation boundaries.}
\label{fig:finite-N-equilibrium}
\end{figure}

We sample the finite-$N$ posterior~\eqref{eq:tilted-posterior} with the exchange algorithm~\citep{murray2006mcmc}. We write $W=A+tI$ with $\Trace A=0$. The spherical constraint $\|\mathbf x\|^2=N$ gives $Z(A+tI)=e^{Nt/2}Z(A)$, so the normalized spherical law is independent of $t$. The posterior therefore factorizes into the law of the traceless matrix $A$ and an independent $t\sim\mathcal N(0,2/(N^2\gamma\eta))$. We update $A$ using proposals reversible with respect to the Gaussian factor of its posterior. Each proposal $A'$ draws $b=K\eta$ independent samples $\mathbf y^r\sim\Prob(\mathbf x\mid A')$ by exact Bingham rejection sampling~\citep{kent2018unified} and is accepted with probability
\begin{equation}
\min\left\{1,\exp\left[\frac12\sum_{r=1}^{b}(\mathbf y^r)^\top(A-A')\mathbf y^r\right]\right\}.
\label{eq:finite-N-exchange}
\end{equation}
Here $K\eta$ is integer. The auxiliary densities cancel the partition functions, retaining the finite-$N$ matrix posterior without a saddle approximation. We restore the independent trace when measuring weight observables. Controls against exact $N=2$ quadrature and the GOE spectrum at $K=\eta=1$ agree within 1.6 Monte Carlo standard errors.

We compare $K=2$, $(c_1,c_2)=(1.25,0.75)$ and $N\in\{32,64,128,256\}$ at $\eta=4$, with $\gamma\in\{0.25,0.4,0.5625,0.7,1.25,1.5\}$ spanning the coalescence and condensation boundaries. Four chains start from different spectra and alignments. For $N=32,64,128$, we use $20{,}000$ warmup transitions followed by $150{,}000$, $300{,}000$ and $600{,}000$ production transitions, respectively. For $N=256$, we use $100{,}000$ warmup transitions followed by $1.2\times10^6$ production transitions ($2.4\times10^6$ at $\gamma=0.25$). We freeze proposal adaptation after warmup, retain every twentieth state, and use sixteen fresh spherical samples per retained state to measure the generated mass. Across the plotted settings, the maximum rank-normalized split or folded $\widehat R$ is 1.0094 and the minimum bulk or tail effective sample size is 1112, meeting thresholds $1.01$ and $400$. A separate four-chain replication at $(N,\eta,\gamma)=(128,4,0.25)$ uses $50{,}000$ warmup transitions and the same production length; it meets the same gates, with a largest difference of 0.98 combined Monte Carlo standard errors across eleven prespecified observables (Bonferroni threshold $2.84$).

Figure~\ref{fig:finite-N-equilibrium} separates Monte Carlo uncertainty from finite-size discrepancies. At $N=256$, the maximum discrepancies from the saddle are 6.5\% (relative) for the eigenvalues and 0.049 (absolute) for the generated mass. These measurements test selected equilibrium observables, with finite-size rounding near phase boundaries; they neither estimate the absolute evidence nor replace the uniqueness argument in App.~\ref{app:saddle-unique}.

\subsection{Computation of sample and weight statistics from the evidence integral}\label{app:evidence-generating}

In this Appendix we establish the following two formulas:
\begin{equation}\label{eq:W-mean}
  \left\langle W \right\rangle_{W\sim\Prob_{\eta}}
  = \frac{1}{N\eta}\sum_{k=1}^{K} \left( \eta\chi_{k} - \frac{1}{c_{k}} \right)
  \left( \mathbf{c}_{k}\mathbf{c}_{k}^{\top}-I \right).
\end{equation}
and
\begin{equation}\label{app:eq:x-statistics-avg-x}
\frac{K}{N} \left\langle \left\langle \mathbf{x}\mathbf{x}^{\top} \right\rangle_{\mathbf x\sim\Prob_{W}} \right\rangle_{W\sim\Prob_{\eta}}
= C - \frac{\gamma}{N\eta}\sum_{k=1}^{K} 
\left( \eta\chi_{k} - \frac{1}{c_{k}} \right)
\left( \mathbf{c}_{k}\mathbf{c}_{k}^{\top} - I \right)
\end{equation}

\subsubsection{Perturbation theory of the evidence}

Let $U$ be a symmetric matrix of finite rank, with $\Trace U=0$. Calling $c_{k}'$ the eigenvalues of $C'=C+zU$, first-order perturbation theory gives
\begin{equation}
c_{k}' = c_{k} + \frac{z}{N} \mathbf{c}_{k}^{\top}U\mathbf{c}_{k} + \BigO(z^{2})
\end{equation}
for $k\leq K$ (the non-zero eigenvalues of $C$ are assumed non-degenerate, with eigenvectors $\mathbf{c}_{k}$ normalized to $\mathbf{c}_{k}^{\top}\mathbf{c}_{k}=N$). Let $U'=\mathbb{P}_{\operatorname*{ker}(C)}U\mathbb{P}_{\operatorname*{ker}(C)}$, where $\mathbb{P}_{\operatorname*{ker}(C)}=I-\frac{1}{N}\sum_{k=1}^{K}\mathbf{c}_{k}\mathbf{c}_{k}^{\top}$. Then $C'$ gains $R=\Rank U'$ additional non-zero eigenvalues,
$$c_{j}'=z\alpha_{j}+ \BigO(z^{2}),\quad(j>K)$$
where $\alpha_{j}$ are the eigenvalues of $U'$.

The only explicit dependence on the eigenvalues of $C$ comes from the HCIZ integral:
\begin{equation}\label{eq:evidence-perturb}
\begin{aligned}
\frac{1}{N}\left.\frac{ \partial }{ \partial z }\ln\Evidence(C+zU)\right|_{z=0}
&= \frac{1}{N}\left.\frac{ \partial \ln\hciz(\Lambda;C+zU) }{ \partial z }\right|_{z=0} \\
&= \frac{1}{2N}\sum_{k=1}^{K} \left(\eta\chi_{k} - \frac{1}{c_{k}} \right) \mathbf{c}_{k}^{\top}U\mathbf{c}_{k} \\
\end{aligned}
\end{equation}
where we used $\frac{1}{\eta c_{k}'}\Stieltjes(\chi_{k}) = 1-u_{k}^{2}$ and $(\lambda_{k}-\chi_{k})u_{k}^{2}=0$. For $j>K$, $c_{j}'\to0$ gives $\chi_{j}=\Stieltjes^{-1}(\eta c_{j}')\sim(\eta c_{j}')^{-1}$, so $(\eta\chi_{j}-1/c_{j}')(\partial c_{j}'/\partial z)=\BigO(c_{j}')\to 0$; the newborn modes $j>K$ drop out.

\subsubsection{Weight statistics}\label{app:w-stats}

Directly differentiating $\ln\Evidence(C+zU)$ in its integral representation gives
\begin{equation}
\left. \frac{ \partial }{ \partial z }\ln\Evidence(C+zU) \right|_{z=0}
= \frac{N\eta}{2}\left\langle \Trace UW \right\rangle_{W\sim\Prob_{\eta}}.
\end{equation}
Combined with~\eqref{eq:evidence-perturb},
\begin{equation}\label{app:eq:weight-statistics}
\left\langle \Trace UW \right\rangle_{W\sim\Prob_{\eta}}
= \frac{1}{N\eta}\sum_{k=1}^{K} \left( \eta\chi_{k} - \frac{1}{c_{k}} \right) \mathbf{c}_{k}^{\top}U\mathbf{c}_{k}
\end{equation}
for any $U$ with $\Trace U=0$. Since $\Trace\langle W\rangle_{\Prob_{\eta}}=0$, this pins down the mean to Eq. \eqref{eq:W-mean}.

\subsubsection{Sample statistics}\label{app:x-stats}

The same generating trick on the other side of the $\ln Z(W)$ factor yields sample moments under the posterior predictive. Shifting $W\to W-zU$ inside the posterior-weighted integral gives
\begin{multline}
\left.\frac{ \partial }{ \partial z } \ln\int \exp \left\{ -K\eta\ln Z(W - zU) - \frac{N\gamma\eta}{4}\Trace W^{2} + \frac{N\eta}{2}\Trace CW \right\} \mathrm{d}W \right|_{z=0} \\
= \frac{K\eta}{2} \left\langle \left\langle \mathbf{x}^{\top}U\mathbf{x} \right\rangle_{\Prob_{W}} \right\rangle_{W\sim\Prob_{\eta}}.
\end{multline}
Completing the square, the log of the integral equals $\ln\Evidence(C-z\gamma U)-\tfrac{N}{4}z^{2}\gamma\eta\Trace U^{2}+\tfrac{N\eta}{2}z\Trace CU$. Differentiating and using~\eqref{eq:evidence-perturb},
\begin{equation}\label{app:eq:x-stats-U}
\begin{aligned}
\frac{K\eta}{2}\left\langle \left\langle \mathbf{x}^{\top}U\mathbf{x} \right\rangle_{\Prob_{W}} \right\rangle_{W\sim\Prob_{\eta}} &= \left. \frac{ \partial }{ \partial z } \ln\Evidence(C-z\gamma U) \right|_{z=0} + \frac{N\eta}{2}\Trace CU \\
&= -\frac{\gamma}{2}\sum_{k=1}^{K} \left( \eta\chi_{k} - \frac{1}{c_{k}} \right) \mathbf{c}_{k}^{\top}U\mathbf{c}_{k} + \frac{N\eta}{2}\Trace CU.
\end{aligned}
\end{equation}
Since $\Trace\langle\langle\mathbf{x}\mathbf{x}^{\top}\rangle_{\Prob_W}\rangle_{\Prob_\eta}=N$, the arbitrariness of $U$ then pins down the full covariance to Eq. \eqref{app:eq:x-statistics-avg-x}.

\subsection{Replica approach to equilibrium properties of trained SBM ensembles}\label{app:replica}

The equilibrium properties of SBM ensembles learned from data were recently derived in an independent manner using the replica method from the statistical physics of disordered systems \citep{tulinski2026}. We summarize here how their approach relates to ours. %

\paragraph{Effective theory.} The derivation is based on the rewriting \begin{align}
    \mathcal{Y}(C)=\lim_{n\to -K\eta}\langle Z(W)^n\rangle_{W\sim\exp(-\frac{N\gamma\eta}{4}\operatorname{tr}W^{2} + \frac{N\eta}{2}\text{tr}(CW))},
\end{align}
where $n$ is the replica number. Pretending that $n$ is a positive integer, $Z(W)^n$ is the partition function of a system made of $n$ \say{replicas} of the SBM. The samples $\Vert\mathbf{x}^a\Vert^2=N$ generated by replicas $a=1,\dots,n$ are mutually independent conditionally on $W$. Integrating over $W$ creates effective interactions of replicas with each other, $Q_{ab}=\mathbf{x}_a\cdot\mathbf{x}_b/N$, and with the data, $M_{ak}=\mathbf{x}_a\cdot\mathbf{c}_k/N$, both of which are introduced as auxiliary integration variables, leading to the expression \begin{align}
    \langle Z^n\rangle \propto \int \mathrm{d}M\,\mathrm{d}Q\,\exp\left(N\Phi_n^{\text{rep}}(M,Q)\right)
\end{align}
where
\begin{align}\label{eq:Phi_rep}
    \Phi_\text{rep}(M,Q)
    &=\frac{\eta\mathrm{tr}(C^2)}{4\gamma}+\frac{\mathrm{tr}(Q^2)}{4\gamma\eta}+\frac{1}{2\gamma}\mathrm{tr}(M\text{diag}(c_k)M^\top)\nonumber\\
    &\quad+\frac{1}{2}\mathrm{tr}\ln(Q-MM^\top)+\frac{n}{2}\ln(2\pi e).
\end{align}
The spherical constraint $\Trace Q=n$ is enforced by a Lagrange multiplier $\Phi_{\mathrm{rep}}\mapsto\Phi_{\mathrm{rep}}-\frac{\mu}{2}(\mathrm{tr}(Q)-n)$. 

\paragraph{Saddle-point analysis.} It was shown in \citep{tulinski2026} that, to leading order in $N$, the integral $\langle Z^n\rangle$ is invariant not only under permutations of replicas $a=1,\dots,n$, but also under orthogonal transformations in $n$-dimensional replica space. As a result, the effective action $\Phi_{\text{rep}}$ depends on $M$ and $Q$ only through the squared norm of the $d\le K$ nonzero columns $\mathbf{m}_1,\dots,\mathbf{m}_d$ of the former and eigenvalues of the latter, with $d$ determined \textit{a posteriori} from the optimizing of $\Phi_{\mathrm{rep}}$.

Integrating over $Q$ and $M$ by saddle point reveals that 
\begin{enumerate}[label=(\roman*)]
    \item\label{item:(i)} columns of the optimal $M$ are eigenvectors of the optimal $Q$, with eigenvalues $q_1,\dots,q_d$;
    \item\label{item:(ii)} the optimal $Q$ is isotropic in subspace orthogonal to their span, with eigenvalue $\widetilde{q}$;
    \item\label{item:(iii)} these $\text{O}(n)$-invariants are related via \begin{align}\label{eq:saddle-point-eqs}
    \mu&=\frac{q_k}{\eta\gamma}+\frac{c_k}{\gamma}=\frac{q_k}{\eta\gamma}+\frac{1}{q_k-m_k^2}=\frac{\widetilde{q}}{\eta\gamma}+\frac{1}{\widetilde{q}},
    \end{align}
    where $m_k^2=\Vert \mathbf{m}_k\Vert^2/n$; 
    \item\label{item:(iv)} it is sufficient for the spherical constraints $\Vert\mathbf{x}^a\Vert^2=N$ to hold on average, \begin{align}\label{eq:rep-spherical-constraint}
        \Trace Q=n,
    \end{align} by $\text{O}(n)$-invariance to leading order in $N$.
\end{enumerate}  

The replica order parameter $m_k$ tracks the overlap $\mathbf{x}\cdot\mathbf{c}_k/N$ of generated samples with the data eigenvector $\mathbf{c}_k$, whereas the condensation parameter $h_k$ of \S\ref{sec:phase-diagram} tracks the overlap $\mathbf{x}\cdot\mathbf{v}_k/N$ with the $k$-th eigenvector $\mathbf{v}_k$ of $W$. At saddle the two are related by $m_k=h_k u_k$ with $u_k=\mathbf{v}_k\cdot\mathbf{c}_k/N$ the $W$-to-data alignment, so $m_k=0$ whenever $h_k=0$ or $u_k=0$. The phase labels below are written in the main-text $(h,u)$ convention of \S\ref{sec:phase-diagram} for direct cross-referencing.

Eqs.~\eqref{eq:Phi_rep}-\eqref{eq:rep-spherical-constraint} are valid provided $n>n_c$ where
\begin{align}
    n_c=-K\eta \, \frac{\sum_{k\le d}\frac{c_k}{K}(1-\frac{1}{c_k}\sqrt{\frac{\gamma}{\eta}})}{1-\sqrt{\eta\gamma}}.
\end{align}The boundary of the $h,u_1\ne 0$ edge phase solves the equation $n(\eta)=n_c(\gamma,\eta,c_1,\dots,c_d)$, which is equivalent to Eq.~\eqref{eq:FMe-conditions}. For $n<n_c$, the solution $\{m_k(n),\widetilde{q}(n)\}$ obtained for $n\ge n_c$ is frozen, and the analytical continuation of $\Phi_{\text{rep}}(n)$ viewed as a function of $n$ is $\Phi_{\text{rep}}(n)=\Phi_{\text{rep}}(n_c)+(n-n_c)\Phi_{\text{rep}}'(n_c)$, see \citep{pastore2019, tulinski2026} for an interpretation of the freezing.

Eq.~\eqref{eq:rep-spherical-constraint} is quadratic in $\widetilde{q}$. The solution, $\widetilde{q}(d)$, depends on $d$ and equals the coalesced $g_1$, Eq.~\eqref{eq:g1-coalesced}. The conditions determining $d$ can be derived from the local stability of the $d$-coalesced saddle points \citep{tulinski2026}, and are $\gamma/c_d<\widetilde{q}(d)<\min\{1,\sqrt{\eta\gamma},\gamma/c_{d+1}\}$ for $n=-K\eta>n_c$, consistent with Eq.~\eqref{eq:FMo-conditions}, and $d=\#\{c_k:c_k>\sqrt{\gamma/\eta}\}$ for $n=-K\eta<n_c$, see Eqs.~\eqref{eq:FMe-conditions}.

From Eqs.~\eqref{eq:saddle-point-eqs}, we have that $(Q-MM^\top)^{-1}$ has $d$ eigenvalues $c_k/\gamma$ along with an $(n-d)$-fold degenerate eigenvalue $g_1^{-1}$, and $q_k=g_1+\eta(\gamma/g_1-c_k)$, $m_k^2=q_k-\gamma/c_k$. It follows that\begin{align}
    \frac{\eta c_k^2}{4\gamma}+\frac{q_k^2}{4\gamma\eta}+\frac{c_km_k^2}{2\gamma}=\begin{cases}
     \frac{g_1^2}{4\gamma\eta}+\frac{\gamma\eta}{4g_{1}^2}, & k\le d,\\
      \frac{\eta c_k^2}{4\gamma}, & k>d.
    \end{cases} 
\end{align}Substituting into $\Phi_{\text{rep}}$ evaluated at $n=-K\eta>n_c$ yields 
\begin{align}
    \Phi_{\text{rep}}(-K\eta)
    &=\frac{\eta}{4\gamma}\sum_{k>d}c_k^2+\frac{d\gamma\eta}{4g_1^2}-\frac{Kg_1^2}{4\gamma}\nonumber\\
    &\quad+\frac{1}{2}\sum_{k=1}^d\ln\frac{\gamma}{c_k}-\frac{K\eta+d}{2}\ln g_1-\frac{K\eta}{2}\ln(2\pi e).
\end{align}
We turn to the total $\Phi$ predicted by Eq.~\eqref{eq:Phi1-eval}. Collecting logarithms from the uncondensed and condensed modes and the base measure, \begin{align*}
    &\frac{K}{2}\ln(\gamma\eta)-\frac{K-d}{2}\ln(\gamma\eta)-\frac{K\eta}{2}\ln g_1-\frac{1}{2}\sum_{k\le d}\ln(g_1\eta\, c_k)-\frac{K\eta}{2}\ln(2\pi)\\
    &\qquad=\frac{1}{2}\sum_{k\le d}\ln \frac{\gamma}{c_k}-\frac{K\eta+d}{2}\ln g_1-\frac{K\eta}{2}\ln(2\pi),
\end{align*}which is the logarithmic part of $\Phi_{\text{rep}}$. The non-logarithmic terms of $\Phi$ give \begin{align}
    \frac{\eta}{4\gamma}\sum_{k>d}c_k^2+\Big(\frac{\eta}{2g_1}+\frac{g_1}{2\gamma}\Big)\Big(\sum_{k\le d}c_k-K\Big)+\frac{Kg_1^2}{4\gamma}-\frac{d}{2}-\frac{d\gamma\eta}{4g_1^2}=\frac{\eta}{4\gamma}\sum_{k>d}c_k^2+\frac{d\gamma\eta}{4g_1^2}-\frac{Kg_1^2}{4\gamma}-\frac{K\eta}{2},\nonumber
\end{align}where we used $\sum_{k\le d}c_k-K=-Kg_1+d\gamma/g_1$. Summing both logarithmic and non-logarithmic parts shows that $\Phi=\Phi_{\text{rep}}(n)$ for $n=-K\eta>n_c$, which describes the $h=u_1=0$, $h=0,u_1\ne 0$ and $h,u_1\ne 0$ outlier phases.

We now turn to $n=-K\eta<n_c$. In that case, the replica solution is $\widetilde{q}=\sqrt{\gamma\eta}$, $\mu=2/\sqrt{\gamma\eta}$, for which the transverse modes of the Hessian $\partial^2\Phi_{\text{rep}}/\partial Q_{ab}\partial M_{ak}$ are flat. The slope of $\Phi_{\text{rep}}$ is given by \begin{align}
    \Phi_{\text{rep}}'(n_c)=\frac{1}{\sqrt{\gamma\eta}}+\frac{1}{4}\ln(\gamma\eta)-\frac{1}{4}+\frac{1}{2}\ln(2\pi).
\end{align} 
Using $(1-1/\sqrt{\gamma\eta})n_c=\sqrt{\eta/\gamma}\sum_{k\le d}c_k-d$, we obtain the expression \begin{align}
    \Phi_{\text{rep}}
    &=\frac{\eta}{4\gamma}\sum_{k>d}c_k^2+\sqrt{\frac{\eta}{\gamma}}\Big(\sum_{k\le d}c_k-K\Big)+\frac{K\eta}{4}-\frac{3d}{4}\nonumber\\
    &\quad+\frac{1}{2}\sum_{k\le d}\ln\frac{\gamma}{c_k}-\frac{K\eta+d}{4}\ln(\gamma\eta)-\frac{K\eta}{2}\ln(2\pi),
\end{align}
valid for $n=-K\eta<n_c$. 

We again turn to the $\Phi$ predicted by Eq.~\eqref{eq:Phi1-eval}. The contributions from the uncondensed and condensed modes are, respectively, $\frac{\eta}{4\gamma}\sum_{k>d}c_k^2-\frac{K-d}{2}\ln(\gamma\eta)$ and $-\frac{3d}{4}+\sqrt{\frac{\eta}{\gamma}}\sum_{k\le d}c_k-\frac{1}{2}\sum_{k\le d}\ln(\sqrt{\eta\gamma}\eta c_k)$, while $-K\eta(\ln Z)/N=-K\sqrt{\eta/\gamma}-\frac{K\eta}{4}\ln(\gamma\eta)+\frac{K\eta}{4}-\frac{K\eta}{2}\ln(2\pi).$ Summing these contributions together with $\frac{K}{2}\ln(\gamma\eta)$, we see that $\Phi=\Phi_{\text{rep}}(n)$ for $n=-K\eta <n_c$ too, thus capturing the $h\ne 0,u_1=0$ and $h,u_1\ne 0$ edge phases.

\clearpage
\section{Training dynamics: Martin--Siggia--Rose derivation}\label{app:dynamics-msr}\label{app:dynamics}

Equations~\eqref{eq:W-langevin}--\eqref{eq:x-langevin} model persistent contrastive divergence in continuous time. Here $t$ denotes parameter-training time, and the chain accumulates $\nu$ units of Langevin sampling time per unit $t$. The weight update uses the data covariance $C$ and the current chain state $\mathbf{x}(t)$ for the positive and negative phases, respectively, while the same chain continues under the evolving weights without requiring equilibration at each update.

\textbf{DMFT notation recap.} For reference, we restate here the main-text DMFT quantities. The two-time order parameters are
\begin{equation}\label{eq:order-params-MSR-Q-R}
  Q(t,u) = \frac{\mathbf{x}(t)\cdot\mathbf{x}(u)}{N},\qquad
  R(t,u) = \left.\frac{\delta\langle x_i(t)\rangle}{\delta j_i(u)}\right|_{j=0},
\end{equation}
with $j_i(t)$ a fictitious source added to the right-hand side of~\eqref{eq:x-langevin}; the memory kernel and noise correlator are
\begin{equation}\label{eq:MD-kernels}
\begin{aligned}
M(t,u) &= e^{-\frac{\gamma}{2}(t-u)}\left[-\frac{K\nu}{2}Q(t,u) + \frac{\nu^2}{\eta\gamma}R(t,u)\right], \\
D(t,u) &= 2\nu\delta(t-u) + \frac{\nu^2}{\eta\gamma}e^{-\frac{\gamma}{2}|t-u|}Q(t,u).
\end{aligned}
\end{equation}

\begin{figure}[!t]
\includegraphics[width=\textwidth]{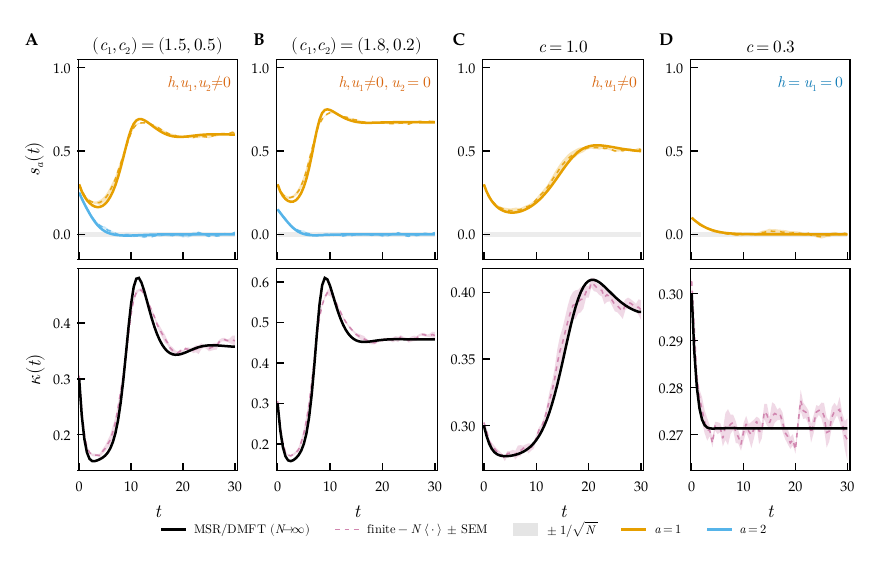}
\caption{\label{fig:msr-langevin-validation}%
\textbf{Validation of the large-$N$ DMFT against finite-$N$ SBM training.} Each panel compares DMFT (solid) with finite-$N$ SGD runs (full positive phase, one persistent negative-phase particle) simulated by Euler integration of the coupled Langevin system~\eqref{eq:x-langevin}--\eqref{eq:W-langevin} at $N=4000$, averaged over five seeds (dashed, with $\pm\mathrm{SEM}$ bands). Top row: signal overlaps $s_{k}(t)$; bottom row: Lagrange multiplier $\kappa(t)$. The gray strip marks the $\pm 1/\sqrt{N}$ finite-$N$ noise floor ($\approx 0.016$). \textbf{A)} $K=2$, $\mathbf{c}=(1.5,\,0.5)$: one channel condenses ($s_{1}\to 0.60$), the other stays at $s=0$. \textbf{B)} $K=2$, $\mathbf{c}=(1.8,\,0.2)$: stronger signal imbalance, $s_{1}\to 0.67$. \textbf{C)} $K=1$, $c=1$: pure $h\ne 0$ phase. \textbf{D)} $K=1$, $c=0.3$: pure $h=0$ phase; both $s$ and $\kappa$ fluctuate near zero. Common parameters: $\gamma=0.5$, $\eta=3$, $\nu=0.3$, $T_{\max}=30$. Finite-$N$ seeds are sign-aligned to the DMFT solution to absorb the $\mathbf{x}\to-\mathbf{x}$ symmetry of~\eqref{eq:x-langevin}--\eqref{eq:W-langevin}.}
\end{figure}

We sketch the derivation of the dynamical mean-field theory summarized in Sec.~\ref{sec:dynamics}. The intermediate expressions~\eqref{eq:W-integrated}--\eqref{eq:Ax-action} are displayed at $K=1$ with a single data direction $\mathbf{c}$ and scalar eigenvalue $c$. Throughout, data eigenvectors are normalized to $\|\mathbf{c}_k\|^{2}=\sum_i c_{k,i}^{2}=N$, so that overlaps $\mathbf{c}_k\!\cdot\!\mathbf{v}_k/N$ are $\BigO(1)$ and the rank-one matrix $\mathbf{c}\mathbf{c}^{\top}\!/N$ in~\eqref{eq:W-integrated} carries the K=1 data covariance at unit scalar eigenvalue. The general-$K$ reduction to Eqs.~\eqref{eq:s-eom}--\eqref{eq:Q-eom} follows by linearity, summing the positive-phase source over the $K$ data modes (restoring the eigenvalue factors $c_k$) and restoring the factor $K$ in the negative-phase self-interaction. The finite-$N$ SGD comparison in Fig.~\ref{fig:msr-langevin-validation} (including the $K=2$ panels) runs with this multi-mode extension. Starting from the coupled Langevin equations~\eqref{eq:x-langevin}--\eqref{eq:W-langevin}, the weight equation can be integrated explicitly against the Ornstein--Uhlenbeck process $W_{\mathrm{GOE}}(t)$ that shares the same noise $\Omega$ but drops the source terms. With the stationary GOE initial condition $W(0)\sim e^{-\frac{N\gamma\eta}{4}\Trace W^{2}}$, the noise correlator satisfies $\langle W_{\mathrm{GOE},ij}(t)W_{\mathrm{GOE},kl}(t')\rangle = (\delta_{ik}\delta_{jl}+\delta_{il}\delta_{jk})e^{-\gamma|t-u|/2}/(N\gamma\eta)$, and
\begin{equation}\label{eq:W-integrated}
  W(t) = W_{\mathrm{GOE}}(t) + \frac{1-e^{-\gamma t/2}}{\gamma} \frac{\mathbf{c}\mathbf{c}^{\top}}{N} - \frac{1}{2}\int_{0}^{t}e^{-\gamma(t-u)/2} \frac{\mathbf{x}(u)\mathbf{x}(u)^{\top}}{N}\,\mathrm{d}u.
\end{equation}
Representing the Langevin equations as delta functionals with response fields $\hat{\mathbf{x}}(t)$ and $\hat{W}(t)$, averaging over the noises, and integrating out $W,\hat{W}$ using~\eqref{eq:W-integrated} yields a closed generating functional over $(\mathbf{x},\hat{\mathbf{x}})$ alone,
\begin{equation}
  \mathcal{Z} \propto \int\mathcal{D}[\mathbf{x},\hat{\mathbf{x}}]\,e^{-N\mathcal{A}_{\mathrm{x}}[\mathbf{x},\hat{\mathbf{x}}]},
\end{equation}
with action
\begin{multline}\label{eq:Ax-action}
  N\mathcal{A}_{\mathrm{x}} = \int\mathrm{d}t\,\left\{\hat{\mathbf{x}}\cdot\dot{\mathbf{x}}+\kappa(t)\hat{\mathbf{x}}\cdot\mathbf{x} - \nu\|\hat{\mathbf{x}}\|^{2} - \frac{\nu}{\gamma N}(1-e^{-\gamma t/2})\,\hat{\mathbf{x}}^{\top}\mathbf{c}\mathbf{c}^{\top}\mathbf{x}\right\} \\
  + \frac{1}{N}\int\mathrm{d}t\,\mathrm{d}u\,\Theta(t-u)\,e^{-\gamma(t-u)/2}\,\biggl\{\tfrac{\nu}{2}(\mathbf{x}(t)\!\cdot\!\mathbf{x}(u))(\hat{\mathbf{x}}(t)\!\cdot\!\mathbf{x}(u)) \\
  - \tfrac{\nu^{2}}{\eta\gamma}(\mathbf{x}(t)\!\cdot\!\mathbf{x}(u))(\hat{\mathbf{x}}(t)\!\cdot\!\hat{\mathbf{x}}(u)) - \tfrac{\nu^{2}}{\eta\gamma}(\hat{\mathbf{x}}(t)\!\cdot\!\mathbf{x}(u))(\mathbf{x}(t)\!\cdot\!\hat{\mathbf{x}}(u))\biggr\},
\end{multline}
in which the Heaviside $\Theta(t-u)$ is inherited from the causal structure of~\eqref{eq:W-integrated}. The bilocal terms depend on $\mathbf{x},\hat{\mathbf{x}}$ only through the rotation-invariant combinations $\mathbf{x}(t)\cdot\mathbf{x}(u)$, $\mathbf{x}(t)\cdot\hat{\mathbf{x}}(u)$, and $\hat{\mathbf{x}}(t)\cdot\hat{\mathbf{x}}(u)$. Introducing the order parameters $Q,R$ of~\eqref{eq:order-params-MSR-Q-R} and the signal $s$ through Hubbard--Stratonovich identities, and by $\hat{\mathbf{x}}(t)\cdot\hat{\mathbf{x}}(u)=\BigO(1)$ (response fields generate no equal-time variance), the saddle reduces to a sum of decoupled one-site contributions at fixed $c$,
\begin{equation}
  \mathcal{A}_{\mathrm{x}}[\mathbf{x},\hat{\mathbf{x}}] = \frac{1}{N}\sum_{i}\mathfrak{a}[x_{i},\hat{x}_{i}|c_{i}],
\end{equation}
from which the effective single-site Gaussian process reads
\begin{equation}\label{eq:x-effective}
  \dot{x}(t) = -\kappa(t)x(t) + \frac{\nu}{\gamma}(1-e^{-\gamma t/2})s(t)\,c + \int_{0}^{t}M(t,u)\,x(u)\,\mathrm{d}u + \zeta(t),
\end{equation}
with the kernels $M(t,u)$ and $D(t,u)$ of~\eqref{eq:MD-kernels} and $\langle\zeta(t)\zeta(u)\rangle=D(t,u)$. Because~\eqref{eq:x-effective} is linear, $\langle x(t)|c\rangle = c\,h(t)$ for a scalar $h(t)$, and the self-consistency $s(t)=\tfrac{1}{N}\sum_{i}c_{i}\langle x(t)|c_{i}\rangle=h(t)$ (using $\sum_{i}c_{i}^{2}=N$) closes the signal equation. The two-time order parameters then satisfy the ODEs~\eqref{eq:s-eom}--\eqref{eq:Q-eom} of Sec.~\ref{sec:dynamics}, supplemented by $R(t,u)=0$ for $t<u$, $\lim_{t\to u^{+}}R(t,u)=1$, $Q(t,u)=Q(u,t)$, and the sphericity constraint $Q(t,t)=1$ which also fixes $\kappa(t)$.

\textbf{Stationary regime.} For long times, the system reaches a stationary, time-translation invariant, regime, for which: $Q(t,u) \to Q_{\mathrm{st}}(t-u)$, $R(t,u)\to R_{\mathrm{st}}(t-u)$, $s(t)\to s_{\mathrm{st}}$, $\kappa(t)\to\kappa_{\mathrm{st}}$, with $Q_{\mathrm{st}}(-\tau)=Q_{\mathrm{st}}(\tau)$. The stationary kernels collapse to 
\begin{equation}
  M_{\mathrm{st}}(\tau)=e^{-\gamma\tau/2}\left\{ -\frac{K\nu}{2}Q_{\mathrm{st}}(\tau)+\frac{\nu^{2}}{\eta\gamma}R_{\mathrm{st}}(\tau) \right\},\quad
D_{\mathrm{st}}(\tau)=2\nu\delta(\tau)+\frac{\nu^{2}}{\eta\gamma}e^{-\gamma|\tau|/2}Q_{\mathrm{st}}(\tau).
\end{equation}
The stationary equations become,
\begin{align}
0 &= \left( -\kappa_{\mathrm{st}}+\frac{\nu c_{a}}{\gamma}+\int_{0}^{\infty}M_{\mathrm{st}}(\tau)\mathrm{d}\tau \right)s_{a,\mathrm{st}},
\qquad a=1,\dots,K, \\
\partial_{\tau}R_{\mathrm{st}}(\tau) &= -\kappa_{\mathrm{st}}R_{\mathrm{st}}(\tau)+\int_{0}^{\tau}M_{\mathrm{st}}(\tau-\sigma)R_{\mathrm{st}}(\sigma)\mathrm{d}\sigma, \\
\partial_{\tau}Q_{\mathrm{st}}(\tau) &= \frac{\nu}{\gamma}\sum_{a=1}^{K}c_{a}s_{a,\mathrm{st}}^{2}-\kappa_{\mathrm{st}}Q_{\mathrm{st}}(\tau)
+\int_{0}^{\tau}M_{\mathrm{st}}(\tau-\sigma)Q_{\mathrm{st}}(\sigma)\mathrm{d}\sigma \notag \\
&\quad+ \int_{0}^{\infty}M_{\mathrm{st}}(\tau+\sigma)Q_{\mathrm{st}}(\sigma)\mathrm{d}\sigma 
+ \int_{0}^{\infty}D_{\mathrm{st}}(\tau+\sigma)R_{\mathrm{st}}(\sigma)\mathrm{d}\sigma 
\end{align}
These equations always admit a solution with $s_{\mathrm{st},k}=0$. The corresponding solution of $R_{\mathrm{st}}(\tau),Q_{\mathrm{st}}(\tau),\kappa_{\mathrm{st}}$ with $s_{\mathrm{st}}=0$ will be denoted by $R_{P}(\tau),Q_{P}(\tau),\kappa_{P}$.

A condensed solution may also exist, satisfying $\kappa_{\mathrm{st}}=\nu c_a/\gamma+\int_{0}^{\infty}M_{\mathrm{st}}(\tau)\mathrm{d}\tau$. To locate its onset, we linearize the stationary signal equation around the uncondensed bath $(R_{P},Q_{P},\kappa_{P})$ and set $s(t)=s_{0}e^{zt}$. The memory integral evaluates as $\int_{0}^{t}M_{P}(t-u)e^{zu}\mathrm{d}u\to e^{zt}\check{M}_{P}(z)$ at large $t$, with $\check{M}_{P}(z)\equiv\int_{0}^{\infty}e^{-z\tau}M_{P}(\tau)\,\mathrm{d}\tau$, so the signal equation reduces to
\begin{equation}\label{eq:signal-spectral}
  z+\kappa_{P}-\check{M}_{P}(z)=\frac{\nu c_a}{\gamma}.
\end{equation}
The Laplace transform of the uncondensed-bath response equation $\partial_{\tau}R_{P}(\tau)=-\kappa_{P}R_{P}(\tau)+\int_{0}^{\tau}M_{P}(\tau-\sigma)R_{P}(\sigma)\,\mathrm{d}\sigma$ with $R_{P}(0)=1$ gives
\begin{equation}\label{eq:RP-laplace}
  \check{R}_{P}(z)=\frac{1}{z+\kappa_{P}-\check{M}_{P}(z)}.
\end{equation}
Substituting~\eqref{eq:RP-laplace} into~\eqref{eq:signal-spectral} yields the spectral equation
\begin{equation}\label{eq:growth-rate}
  1 = \frac{\nu c_a}{\gamma}\int_{0}^{\infty}e^{-z\tau}R_{P}(\tau)\,\mathrm{d}\tau, 
\end{equation}
whose marginal solution $z=0$ isolates the static condensation threshold of mode $a$, $\nu_{c,a}=\gamma/(c_a\chi_{P})$, with $\chi_{P}=\int_{0}^{\infty}R_{P}(\tau)\,\mathrm{d}\tau$. Since $c_1>\dots>c_K$, mode $1$ destabilizes first, fixing the dynamical phase boundary at $\nu_{c}=\gamma/(c_1\chi_{P})$.
This boundary marks the linear instability of the zero-overlap solution. 

Note that if $N\to\infty$ is taken before the long-time limit with random initialization, Eq.~\eqref{eq:s-eom} preserves $s_1(t)=0$ even for $\nu>\nu_c$; reaching a condensed branch requires a nonzero seed or finite-$N$ fluctuations. For this reason, we initialize $s_1(0)$ at a small nonzero value.

Numerically, the two-time system~\eqref{eq:s-eom}--\eqref{eq:Q-eom} is solved by causal row-by-row time marching with an implicit predictor--corrector at each new time, enforcing $Q(t_{n},t_{n})=1$ through $\kappa(t_{n})$.

\paragraph*{Numerical validation against finite-$N$ SGD training.}
Figure~\ref{fig:msr-langevin-validation} compares the DMFT solution of Eqs.~\eqref{eq:s-eom}--\eqref{eq:Q-eom} with finite-$N$ SBM training simulated by Euler--Maruyama integration of the coupled Langevin system~\eqref{eq:x-langevin}--\eqref{eq:W-langevin}, using the full positive phase and one persistent negative-phase particle, with the GOE noise $\Omega/\sqrt{\eta N}$ playing the role of the stochastic-gradient noise. This validates the large-$N$ DMFT~\eqref{eq:s-eom}--\eqref{eq:Q-eom} as a description of finite-$N$ SGD training of the SBM. Four representative cases are shown, spanning $K\in\{1,2\}$ and covering both phases; in each case we plot the signal overlaps $s_{k}(t)$ and the spherical-constraint multiplier $\kappa(t)$. Parameters are chosen so that the transient dip of $s_{1}(t)$, caused by the weight matrix $W$ needing some time to build the signal eigenvector before $\mathbf{x}$ can condense onto it, remains larger than the $1/\sqrt{N}$ noise floor; below that floor, finite-$N$ fluctuations dominate the trajectory and the comparison is obscured by thermal noise. Paramagnetic channels and the pure-PM case stay at the $1/\sqrt{N}$ floor in finite-$N$ realisations, consistent with DMFT setting them exactly to zero. The agreement confirms that the large-$N$ DMFT reduction reproduces the macroscopic observables of finite-$N$ SGD training.
We also tested discrete SGD with one randomly sampled training example per update and no added weight noise, finding good agreement with the $\eta\to\infty$ DMFT at small learning steps (not shown).

\subsection{Upper bound on the outlier count}\label{app:outlier-count}

The decomposition~\eqref{eq:W-integrated-main} implies that $W(t)$ carries at most $K$ upper outliers at every fixed time. The negative-phase integral
\begin{equation}\label{eq:negphase-psd}
  \int_{0}^{t} e^{-\gamma(t-u)/2} \frac{\mathbf{x}(u)\mathbf{x}(u)^{\top}}{N} \mathrm{d}u \;\succeq\; 0
\end{equation}
is positive semidefinite, so
\begin{equation}\label{eq:W-loewner-upper}
  W(t) \;\preceq\; W_{\mathrm{GOE}}(t)+\frac{1-e^{-\gamma t/2}}{\gamma}\,C.
\end{equation}
Monotonicity of ordered eigenvalues under the Loewner order then gives $\lambda_{j}(W(t))\le\lambda_{j}(W_{\mathrm{GOE}}(t)+\tfrac{1-e^{-\gamma t/2}}{\gamma}C)$ for every $j\ge1$, where $\lambda_{j}(\,\cdot\,)$ labels eigenvalues in decreasing order. Since $C$ has rank $K$ and is positive semidefinite, the min-max principle yields
\begin{equation}\label{eq:minmax-Kshift}
  \lambda_{K+j}\left(W_{\mathrm{GOE}}(t)+\frac{1-e^{-\gamma t/2}}{\gamma}C\right) \le \lambda_{j} \left(W_{\mathrm{GOE}}(t)\right),\qquad (j\ge1),
\end{equation}
and in particular $\lambda_{K+1}(W(t))\le\lambda_{1}(W_{\mathrm{GOE}}(t))$. With the stationary GOE initialization $W(0)\sim e^{-N\gamma\eta\Trace W^{2}/4}$, the process $W_{\mathrm{GOE}}(t)$ remains GOE at every time, with semicircle edge $2/\sqrt{\gamma\eta}$; hence for every $\varepsilon>0$, with high probability at large $N$,
\begin{equation}\label{eq:outlier-count-bound}
  \lambda_{K+1}(W(t)) \le \frac{2}{\sqrt{\gamma\eta}}+\varepsilon.
\end{equation}
Therefore, no more than $K$ eigenvalues of $W(t)$ are detached above the bulk edge at any given time.

\subsection{MAP limit of the dynamical threshold}\label{app:map-limit}

In the MAP regime $\eta\to\infty$, the learning-noise feedback drops out of both kernels in~\eqref{eq:MD-kernels}: $M(t,u)\to -\tfrac{\nu}{2}e^{-\gamma(t-u)/2}Q(t,u)$ and $D(t,u)\to 2\nu\,\delta(t-u)$. On the stationary uncondensed bath, the one-sided Laplace transform $\hat{f}(s)\equiv\int_{0}^{\infty}\!e^{-s\tau}f(\tau)\,\mathrm{d}\tau$ yields $\chi_{P}=1/[\kappa_{P}+(\nu/2)\hat{Q}_{P}(\gamma/2)]$, while the spherical constraint gives $\kappa_{P}=\nu-(\nu/2)\hat{Q_{P}^{2}}(\gamma/2)$. We work at $K=1$ with the unit-eigenvalue normalization $c_1=1$ inherited from $\|\mathbf{c}\|^{2}=N$, so the leading-mode condensation condition $\nu c_1\chi_{P}=\gamma$ reduces to $\nu\chi_{P}=\gamma$ and rearranges into
\begin{equation}\label{eq:map-bound}
  \tfrac{\gamma}{2}\bigl[\hat{Q}_{P}(\gamma/2)-\hat{Q_{P}^{2}}(\gamma/2)\bigr] \;=\; 1-\gamma.
\end{equation}
Because the LHS is the integral of $e^{-\gamma\tau/2}Q_{P}(\tau)[1-Q_{P}(\tau)]\geq 0$, it is non-negative; hence $\gamma>1$ admits no solution and the system is uncondensed for \emph{every} $\nu$. For $\gamma<1$ the LHS depends on $\nu$ through the bath auto-correlation $Q_{P}(\tau;\nu,\gamma)$, and a genuine $\nu$-dependent boundary survives. The resulting MAP phase diagram is shown in Fig.~\ref{fig:dynamics}D of the main text: the boundary $\nu_{c}(\gamma)$ occupies only a narrow sliver below $\gamma=1$, diverging vertically as $\gamma\to 1^{-}$ and disappearing below $\gamma_{\min}\approx 0.84$ where the uncondensed bath turns condensed for all $\nu$.

\begin{figure}
\centering
\includegraphics[width=0.95\textwidth]{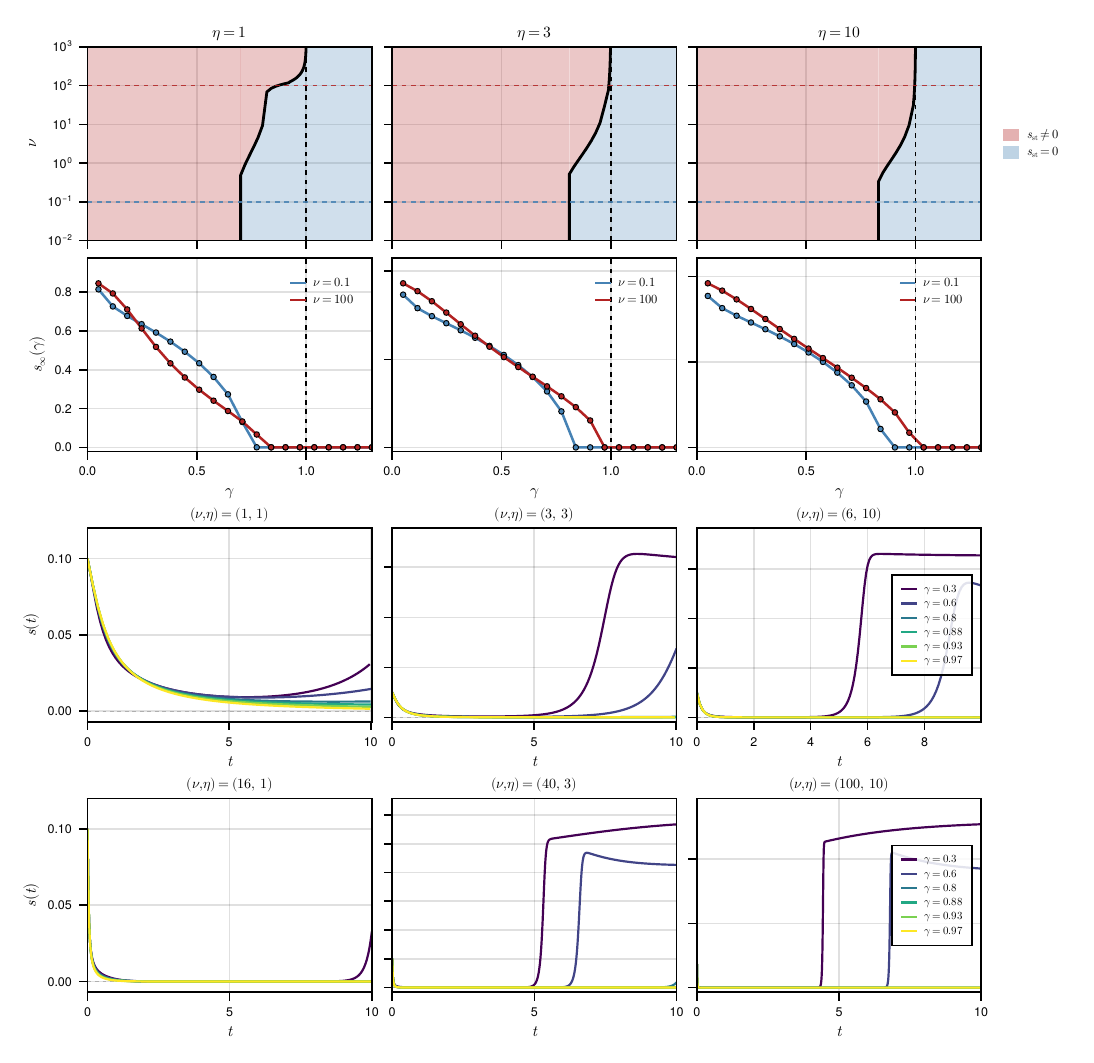}
\caption{\label{fig:finite-eta-phase-diagrams}%
\textbf{Dynamical phase structure of the $K{=}1$ DMFT at finite $\eta$.} Three values $\eta\in\{1,3,10\}$ are shown in columns. \textbf{Row 1}: dynamical phase diagram $\nu_{c}(\gamma)$ (solid black), obtained by warm-started path continuation on a fine $\tau$-grid ($\Delta\tau\simeq 4\times 10^{-3}$) with the guard $\nu_{c}\Delta\tau<0.6$; red shading is the condensed branch ($s_{\mathrm{st}}\neq 0$) and blue shading the uncondensed branch ($s_{\mathrm{st}}\to 0$). The dashed vertical line marks $\gamma=1$, the vertical asymptote of the boundary at every finite $\eta$. Horizontal dashed lines at $\nu=0.1$ (blue) and $\nu=100$ (red) mark the two fixed sampling rates swept in row~2. \textbf{Row 2}: stationary $s_{\mathrm{st}}(\gamma)$ from the time-translation invariant DMFT branch-follower at $\nu=0.1$ (blue) and $\nu=100$ (red), evaluated on a 20-point $\gamma$-grid (circles) and connected by straight segments. The blue curve vanishes at a strictly smaller $\gamma$ than the red curve, quantifying the widening of the condensed window as $\nu$ increases; both collapse to zero at $\gamma=1$. \textbf{Rows 3--4}: families of causal MSR trajectories $s(t)$ starting from $s(0)=0.1$, integrated to $t=10$, at six log-spaced integer sampling rates $\nu\in\{1,3,6,16,40,100\}$, paired per $\eta$ column as $(\nu_{\mathrm{lo}},\nu_{\mathrm{hi}})=(1,16),(3,40),(6,100)$ (titled in each panel) so each pair brackets the numerical $\nu_{c}$ range of its column, with a sweep of six $\gamma$ values bracketing the transition (low-$\gamma$ curves in dark purple, high-$\gamma$ in yellow, viridis gradient). Trajectories with $\gamma$ below the boundary condense to a finite plateau; those with $\gamma$ above decay to zero. The $t$ axis is identical across columns within each row.}
\end{figure}

\textbf{Finite-$\eta$ dynamical phase diagram.}
The MAP-limit structure survives at finite $\eta$: a grid-convergence study at fixed $\gamma=2$ finds the apparent crossing $\nu\chi_{P}=\gamma$ scaling as $\nu_{c}\propto 1/\Delta\tau$ for every $\eta$ tested, confirming that no physical boundary exists for $\gamma>1$. Figure~\ref{fig:finite-eta-phase-diagrams} shows the grid-converged sliver at $\eta\in\{1,3,10\}$: $\gamma_{\min}(\eta)$ approaches the MAP value $\approx 0.84$ and the boundary steepens toward $\gamma=1$ as $\eta$ grows. The stationary $s_{\mathrm{st}}(\gamma)$ sweeps (row~2) and causal MSR trajectories (rows~3--4) confirm this structure over two decades of $\nu$; the ceiling $\nu_{c}\lesssim 10^{2}$ visible in row~1 reflects the convergence guard $\nu_{c}\Delta\tau<0.6$.

\clearpage
\section{Nearly finite-dimensional data}\label{app:ooe-low-dim-data}
This Appendix shows that the static and dynamical solutions obtained here are valid descriptions of SBM training on $K\to\infty$ data living in a $K'$-dimensional slab embedded in $N$-dimensional space, $N\to\infty$. The condition is that $C$ has $K'=\mathcal{O}(1)$ eigenvalues which are $\BigO(K)$ while the remaining $\min\{N,K\}-K'$ stay $o(K)$. Depending on the scaling of $t$ and $\gamma,\eta,\nu$ with $K$, different regimes arise, some of which we characterize below.

\subsection{Statics} 

\subsubsection{Bare large-$K$ regime}
We consider the regime $\gamma,\eta=\mathcal{O}(1)$. This regime was first characterized through a replica analysis in \citep{tulinski2026}. We summarize their main findings in the following paragraph.

A typical $W\sim \Prob_\eta(\cdot\vert\mathcal{D})$ after training is found to exhibit the following properties. Its eigenvectors accurately recover the manifold directions, $\mathbf{v}_{1,\dots,K'}=\mathbf{c}_{1,\dots,K'}$, provided extensive modes are non-degenerate, and $h_{1,\dots,K'}^2=\widetilde{c}_{1,\dots,K'}$ where $\widetilde{c}_{1,\dots,K'}=c_{1,\dots,K'}/K$. Generated data $\mathbf{x}\sim \Prob_W$ have vanishing projections along the other directions as long as \begin{equation}
\gamma>\gamma_{\text{bulk}} \quad \text{where}\quad
\gamma_\text{bulk}=c_{K'+1}\left(\delta_K+\frac{K'}{K}c_{K'+1}\right),\quad \text{with} \quad \delta_K= 1 - \frac 1K \sum_{\ell=1}^{K'} c_\ell \,.
\end{equation}
Here $\delta_K$ is the fraction of variance unexplained by the $K'$ projections of the generated data along $\mathbf{v}_{1,\dots,K'}$. When $\gamma>\gamma_\mathrm{bulk}$, the $\eta_c(\gamma,c_{1,\dots,K})$ delimiting the boundary of the edge phases with the others, scales like $K'/K$ for thin slabs, $\delta_K\ll K^{-1/2}$, and like $\delta_K^2/\gamma$ for thick slabs, $\delta_K\gg K^{-1/2}$.

\subsection{Dynamics}

\subsubsection{Bare large-$K$ regime}\label{app:bare-large-k}
We consider the regime \begin{align}
    t,\gamma,\eta,\nu=\mathcal{O}(1).
\end{align}
\paragraph{Large-$K$ limit of the DMFT} We take $K\to\infty$ in the DMFT derived sending $N\to\infty$ first while holding $K$ fixed. For large $K$, the noise correlator stays $\BigO(1)$. The memory kernel behaves as
\begin{align}\label{eq:bare-large-k-memory}
    \widetilde{M}(t,u)\equiv\frac{M(t,u)}{K}=-\frac{\nu}{2}e^{-\frac{\gamma}{2}(t-u)}Q(t,u)+\mathcal{O}(K^{-1}), \qquad t>u,
\end{align}dominated by the negative phase.

Using that $\kappa(t)$ satisfies $\partial_tQ(t,t)=0$, we get that $\kappa(t)$ also scales linearly with $K$, as \begin{align}\label{eq:bare-large-k-multiplier}
    \widetilde{\kappa}(t)\equiv\frac{\kappa(t)}{K\nu}=\frac{1-e^{-\gamma t/2}}{\gamma}\sum_{\ell=1}^{K'}\widetilde{c}_\ell s_\ell(t)^2-\frac{1}{2}\int_0^t\mathrm{d}u\,e^{-\gamma(t-u)/2}Q(t,u)^2+\mathcal{O}(K^{-1}).
\end{align}

On $\BigO(1)$ timescales, the signals $s_\ell(t),s_k(t)$ ($\ell\le K'<k$) are quasistatic, with\begin{align}
    \!\!\!\begin{cases}\label{eq:bare-large-k-signal}
        0=-\widetilde{\kappa}(t)s_\ell(t)+(1-e^{-\gamma t/2})\frac{\widetilde{c}_\ell}{\gamma}s_\ell(t)-\frac{1}{2}\int_0^t\mathrm{d}u\,e^{-\gamma(t-u)/2}Q(t,u)s_\ell(u)+o(1),\\
        0=-\widetilde{\kappa}(t)s_k(t)-\frac{1}{2}\int_0^t\mathrm{d}u\,e^{-\gamma(t-u)/2}Q(t,u)s_k(u)+o(1).
    \end{cases}
\end{align}The correlation function also behaves quasistatically on finite timescales,
\begin{align}\label{eq:bare-large-k-correlation}
   \!\!0&=-\widetilde{\kappa}(t)Q(t,t')+(1-e^{-\gamma t/2})\sum_\ell\frac{\widetilde{c}_\ell}{\gamma}s_\ell(t)s_\ell(t')-\frac{1}{2}\int_0^t\mathrm{d}u\,e^{-\gamma(t-u)/2}Q(t,u)Q(u,t'),
\end{align}while the response function obeys \begin{align}\label{eq:bare-large-k-response}
\partial_tR(t,t')=-K\nu\widetilde{\kappa}(t)R(t,t')-\frac{K\nu}{2}\int_{t'}^t\mathrm{d}u\,e^{-\gamma(t-u)/2}Q(t,u)R(u,t')+O(1), \quad t>t',
\end{align}together with $R(t,t)=1$. Since the restoring force is linear in $K\nu$, we expect small perturbations at a time $t'<t$ to be suppressed on fast time scales $t-t'\sim1/K\nu$. Inspecting $R(t,t')$ close to the diagonal, $t=t'+\theta/K\nu$, we see that the integrand of the memory term is $\BigO(1)$, so the integral is $\sim 1/K\nu$, and $R(t,t')\sim e^{-K\nu\widetilde{\kappa}(t')(t-t')}\downarrow0$ ($t>t'$).

\paragraph{Large $K$ non-equilibrium stationary state} Now take $t,t'\to \infty$, with $\tau=t-t'=\mathcal{O}(1)$, and write $Q(t,t'),R(t,t')\to Q_\text{st}(\tau),R_\text{st}(\tau)$, $s_\ell(t),s_k(t),\kappa(t)\to s_{\ell,\text{st}},s_{k,\text{st}},\kappa_{\text{st}}$.

The $\BigO(1)$ bulk modes stay uncondensed, $s_k=0$ ($k>D)$, meanwhile extensive modes satisfy \begin{align}\label{eq:s_ell-large-K-stationarity}
    0=\left[\widetilde{\kappa}_{\text{st}}+\frac{1}{2}\int_0^\infty\mathrm{d}\tau\,e^{-\gamma\tau/2}Q_{\text{st}}(\tau)-\frac{\widetilde{c}_\ell}{\gamma}\right]s_{\ell,\text{st}}, \qquad \ell=1,\dots,K'.
\end{align}Hence, only modes with the same maximum rescaled eigenvalue $\widetilde{c}_{\max}$ can condense.

The stationary response function satisfies \begin{align}
\partial_\tau R_\text{st}(\tau)=-K\nu\left[\widetilde{\kappa}_{\mathrm{st}}R_{\mathrm{st}}(\tau)+\frac{1}{2}\int_{0}^\tau\mathrm{d}\sigma\,e^{-\gamma\sigma/2}Q_\text{st}(\sigma)R_\text{st}(\tau-\sigma)\right]
\end{align}together with $R_{\mathrm{st}}(0^+)=1$. Taking a one-sided Laplace transform $\check{R}(p)=\int_0^\infty\mathrm{d}\tau\,e^{-\tau p}R(\tau)$ yields \begin{align}
    \check{R}_\text{st}(p)=\frac{1}{p+K\nu(\widetilde{\kappa}_{\text{st}}+\frac{1}{2}\int_0^\infty\mathrm{d}\tau\,e^{-(p+\gamma/2)\tau}Q_\text{st}(\tau))}.
\end{align}In particular, $\int_0^\infty\mathrm{d}\tau\,R_\text{st}(\tau)=\check R(0)=\frac{\gamma}{K\nu\widetilde{c}_{\max}}\downarrow0$.

The equation for the stationary correlation function is \begin{align}
    0=-\widetilde{\kappa}_{\text{st}}Q_{\text{st}}(\tau)+\frac{1}{\gamma}\sum_\ell\widetilde{c}_\ell s_{\ell,{\text{st}}}^2-\frac{1}{2}\int_0^\infty\mathrm{d}\sigma\,e^{-\gamma\sigma/2}Q_{\text{st}}(\sigma)Q_{\text{st}}(\tau-\sigma).
\end{align}
Writing $\mathbf{x}(t)=\langle\mathbf{x}(t)\rangle+\delta\mathbf{x}(t)$ with $\langle\delta\mathbf{x}(t)\rangle=0$, one has $Q_{\text{st}}(\tau)=q+C(\tau)$, $q=\sum_{\ell\le K'}\delta_{c_\ell,c_{\max}}s_\ell^2$, $C(\tau)=\langle\delta\mathbf{x}(t)\cdot\delta\mathbf{x}(t-\tau)\rangle/N$. Combining the multiplier and signal equations,\begin{align}\label{eq:Q-large-K-stationary}
    \frac{1}{2}\int_0^\infty\mathrm{d}\tau\,e^{-\gamma\tau/2}C(\tau)^2=\frac{1}{\gamma}(1-q)(q-\widetilde{c})+\frac{1-2q}{2}\int_0^\infty\mathrm{d}\tau\,e^{-\gamma\tau/2}C(\tau).
\end{align}From Eq. \eqref{eq:x-effective}, fluctuations obey the linear equation
\begin{align}
    \frac{\mathrm{d}}{\mathrm{d}t}\delta\mathbf{x}=-K\nu\left[\widetilde{\kappa}_{\text{st}}\delta\mathbf{x}(t)+\frac{1}{2}\int_0^\infty\mathrm{d}\sigma\,e^{-\gamma \sigma/2}Q_{\text{st}}(\sigma)\delta\mathbf{x}(t-\sigma)\right]+\boldsymbol{\zeta}(t).
\end{align}When $\delta\mathbf{x}(t)$ varies slowly, $\delta\mathbf{x}(t-\sigma)\approx\delta\mathbf{x}(t)$ over the memory window $\sigma\sim2\gamma^{-1}$ and then $\frac{\mathrm{d}}{\mathrm{d}t}\delta\mathbf{x}\approx-K\nu[\widetilde{\kappa}_{\text{st}}+\frac{1}{2}\int_0^\infty\mathrm{d}\sigma\,e^{-\gamma\sigma/2}Q_{\text{st}}(\sigma)]\delta\mathbf{x}(t)$. Stationarity implies that the bracket equals $\widetilde{c}_{\max}/\gamma>0$, hence slow fluctuations are suppressed, and the overlap of $C(\tau)$ with kernels of a fixed $\BigO(1)$ width vanishes. In particular, $\int_0^\infty\mathrm{d}\tau\,e^{-\gamma\tau/2}C(\tau)=0$. Using $\int_0^\infty\mathrm{d}\tau\,e^{-\gamma\tau/2}C(\tau)=0$, Eq. \eqref{eq:Q-large-K-stationary} gives $q\ge\widetilde{c}_{\max}$ since $q\le 1$ and LHS $>0$, while Eq. \eqref{eq:s_ell-large-K-stationarity} gives $\gamma\widetilde{\kappa}_{\text{st}}=\widetilde{c}_{\max}-q$, whence $q\le \widetilde{c}_{\max}$. As a result, $q=\widetilde{c}_{\max}$ and $\Delta=\widetilde{\kappa}_{\text{st}}=0$.  We conclude that $Q_{\text{st}}(\tau)=q$ with $\tau>0$ for $K\to\infty$. Contrast this with the finite-$K$ case, where we expect $Q_{\text{st}}(\tau)$ to plateau only when $\tau\to\infty$.

The non-equilibrium stationary state is thus characterized by \begin{align}
    q=\sum_{\ell}\delta_{c_\ell,c_{\max}}s_{\ell}^2=\min\{1,\widetilde{c}_{\max}\}, \qquad \widetilde{\kappa}_{\text{st}}=\frac{(\widetilde{c}_{\max}-1)_+}{\gamma},
\end{align}independently of $\eta$ and $\nu$.  To leading order in $K$, then, the negative phase cancels the contribution to $W$ of the top extensive mode in the positive phase, see Eq. \eqref{eq:W-integrated-main}.  As a result, at long times, the top mode of $W$ is not the sampled direction, but the largest extensive unsampled direction.

\paragraph{Action in the bare large-$K$ regime}

The solution above was derived by sending $K\to\infty$ after having first taken $N\to\infty$ holding $K$ fixed. Using $\kappa=K\nu\widetilde{\kappa}$ and $c_\ell=K\widetilde{c}_\ell$ for the extensive modes, the general-$K$ form of Eq. \eqref{eq:Ax-action} decomposes as $N\mathcal{A}_{\mathbf{x}}=NK\nu\mathcal{A}_\text{eff}+\mathcal{A}_\text{rem}$, where
\begin{align}
    \mathcal{A}_\text{eff}
    &=\int\mathrm{d}t\,\widetilde{\kappa}(t)\frac{\hat{\mathbf{x}}(t)\cdot\mathbf{x}(t)}{N}\nonumber\\
    &\quad-\int\mathrm{d}t\,(1-e^{-\gamma t/2})\sum_{\ell\le K'}\frac{\widetilde{c}_\ell}{\gamma}\frac{\mathbf{x}(t)\cdot\mathbf{c}_\ell}{N}\frac{\hat{\mathbf{x}}(t)\cdot\mathbf{c}_\ell}{N}\nonumber\\
    &\quad+\frac{1}{2}\int_{t>u}\mathrm{d}t\,\mathrm{d}u\,e^{-\gamma(t-u)/2}Q(t,u)\frac{\hat{\mathbf{x}}(t)\cdot\mathbf{x}(u)}{N}.
\end{align}
The remainder contains the kinetic term, the spin noise, the weight noise, and the subextensive data modes:
\begin{align*}
    \mathcal{A}_\text{rem}
    &=\int\mathrm{d}t\,\left[\hat{\mathbf{x}}(t)\cdot\dot{\mathbf{x}}(t)-\nu\Vert\hat{\mathbf{x}}(t)\Vert^2\right]\\
    &\quad-\frac{\nu^2}{\eta\gamma N}\int_{t>u}\mathrm{d}t\,\mathrm{d}u\,e^{-\gamma(t-u)/2}(\mathbf{x}(t)\cdot\mathbf{x}(u))(\hat{\mathbf{x}}(t)\cdot\hat{\mathbf{x}}(u))\\
    &\quad-\frac{\nu^2}{\eta\gamma N}\int_{t>u}\mathrm{d}t\,\mathrm{d}u\,e^{-\gamma(t-u)/2}(\hat{\mathbf{x}}(t)\cdot\mathbf{x}(u))(\mathbf{x}(t)\cdot\hat{\mathbf{x}}(u))\\
    &\quad-\frac{\nu}{\gamma N}\int\mathrm{d}t\,(1-e^{-\gamma t/2})\sum_{k=K'+1}^{\operatorname{rank}C}c_k(\mathbf{x}(t)\cdot\mathbf{c}_k)(\hat{\mathbf{x}}(t)\cdot\mathbf{c}_k).
\end{align*}
The formal leading action $\mathcal{A}_\text{eff}$ gives the quasistatic signal and correlation equations above.

\begin{figure}[h]
    \centering
    \includegraphics[width=1.0\linewidth]{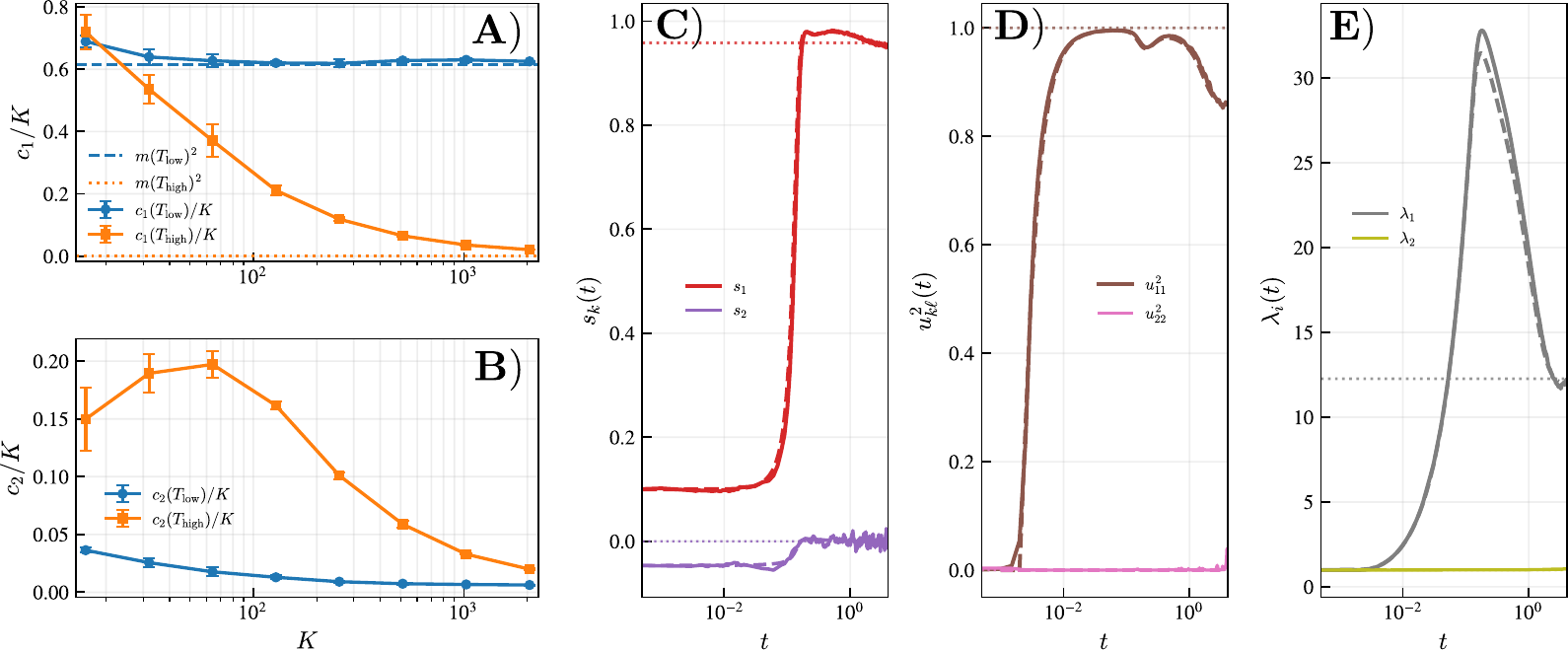}
    \caption{\textbf{Validation of the large-$K$ DMFT equations for nearly finite-dimensional Ising data.} \textbf{A, B)} Spectrum of the empirical covariance matrix $C$ for $K$ configurations of the two-dimensional periodic Ising model at fixed linear size $L=32$, $N=L\times L$($=1024$). Below the critical temperature, $T_{\rm low}=2.2<T_c\simeq 2.269$, the leading eigenvalue is extensive, $c_1\simeq K m(T_{\rm low})^2$, while $c_2/K$ decreases with $K$, consistent with effectively $K'=1$ data. Above the transition, $T_{\rm high}=3.2>T_c$, no extensive mode is present and both $c_1/K$ and $c_2/K$ vanish with $K$. Points and error bars show the mean and standard deviation over three independently sampled datasets. \textbf{C--E)} Comparison between finite-$N$ coupled Langevin simulations and the DMFT prediction for a representative below-critical dataset with $K=512$. Solid curves show finite-$N$ simulations, dashed curves show the DMFT solution using the empirical spectrum of $C$, and dotted horizontal lines show the large-$K$ stationary predictions. \textbf{C)} Signal overlaps $s_k(t)=c_k\cdot x(t)/N$. The dynamics condenses onto the leading Ising mode, while the second mode remains negligible. \textbf{D)} Squared eigenvector overlaps $u_{k}^2(t)$, showing alignment of the top learned eigenvector with the leading data direction. \textbf{E)} Top eigenvalues of $W(t)$, showing the detachment of $\lambda_1$ from the Wigner bulk edge while $\lambda_2$ remains near the edge. } \label{fig:ising-dmft-largeK}
\end{figure}

\subsubsection{Invariant large-$K$ regime}\label{app:invar-large-k}
We now show that the DMFT \eqref{eq:s-eom}-\eqref{eq:MD-kernels} is left invariant under the rescaling
\begin{align}
    \overline{t}=\frac{K}{K'}t, \quad \gamma=\frac{K}{K'}\overline{\gamma}, \quad  \eta=\frac{K'}{K}\overline{\eta},\quad \nu=\frac{K}{K'}\overline{\nu}, \qquad \text{with}\qquad \overline{t},\overline{\gamma},\overline{\eta},\overline{\nu}=\mathcal{O}(1),
\end{align} and describes the large-$K$ Langevin training dynamics Eq. \eqref{eq:W-langevin}-\eqref{eq:x-langevin} on nearly finite-$K'$ training data.

\begin{figure}[h]
    \centering
    \includegraphics[width=0.9\linewidth]{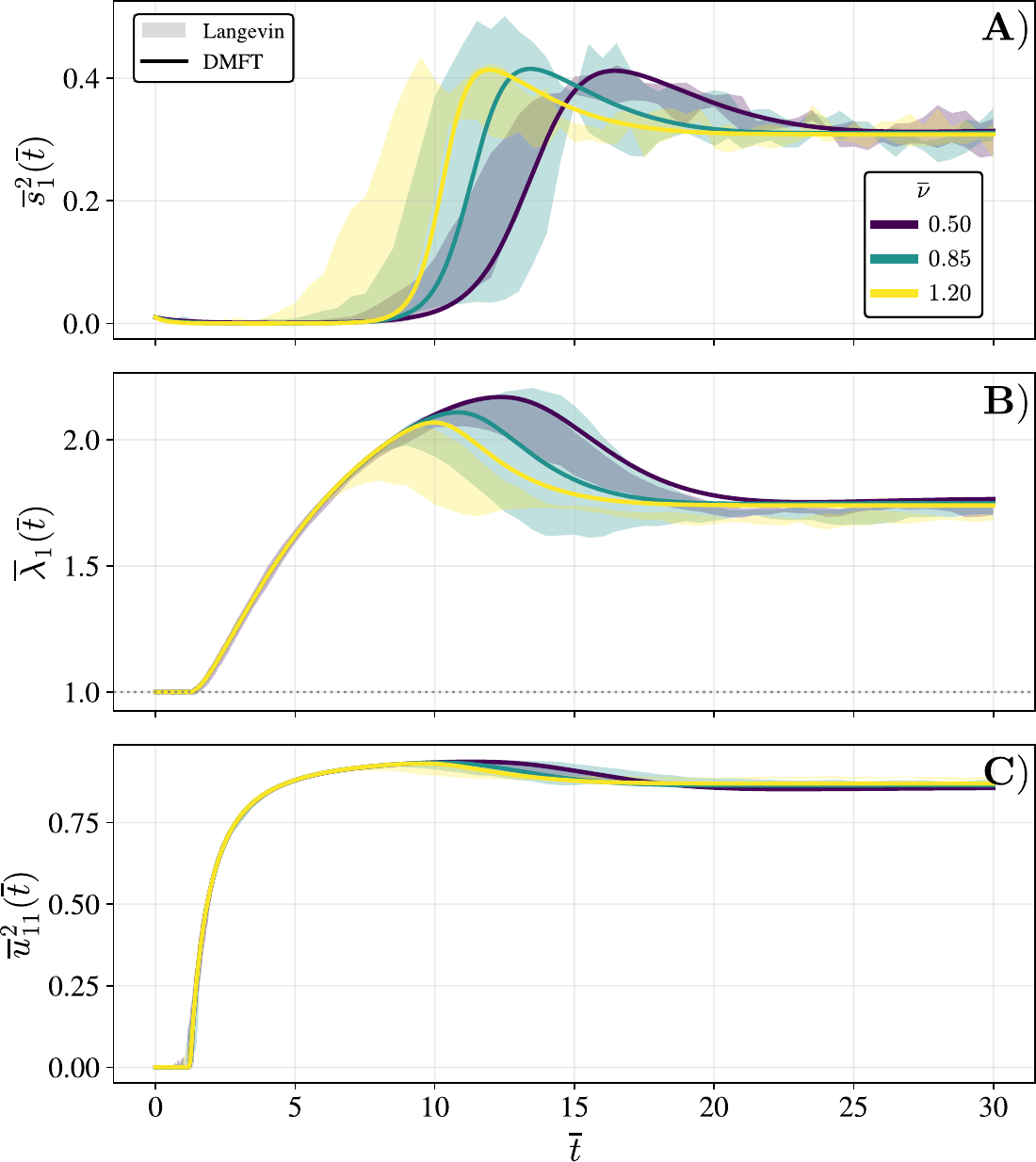}
    \caption{\textbf{Large-$K$ training dynamics for nearly finite-dimensional Ising data.} Comparison between the effective $K'=1$ large-$K$ DMFT prediction and finite-$N$ coupled Langevin simulations for below-critical two-dimensional Ising data at $T=1.8$, with $K=512$ and $N=2025$. Time is shown in the rescaled variable $\overline t = Kt/K'$, and colors indicate the rescaled persistent-chain update rate $\overline\nu$. Solid curves show the effective $K'=1$ DMFT solution, while shaded regions show finite-$N$ Langevin simulations over independent noise seeds, plotted as mean $\pm$ one standard deviation; the simulation mean is not separately drawn. \textbf{A)} Squared signal overlap $\overline s_1^2(\overline t)$, showing condensation of the persistent chain onto the leading Ising covariance mode. \textbf{B)} Top eigenvalue $\overline\lambda_1(\overline t)$ of the learned weight matrix, with the dotted line marking the bulk edge. \textbf{C)} Squared eigenvector overlap $\overline u_{11}^2(\overline t)$, showing alignment of the top eigenvector with the leading data mode. Increasing $\overline\nu$ accelerates the onset of the negative phase, reducing the eigenvalue overshoot while leaving the learned direction largely unchanged.}
\label{fig:largeK_nu_sweep}
\end{figure}

We introduce \begin{align}
    \overline{c}_{1,\dots,K'}=K'\widetilde{c}_{1,\dots,K'}\Big(\!\!=\frac{K'}{K}c_{1,\dots,K'}\Big), \qquad \overline{C}=\frac{K'}{K}C,
\end{align}together with the scaled-time dynamical variables \begin{align}
    \overline{W}(\overline{t})=W(t), \qquad \overline{\mathbf{x}}(\overline{t})=\mathbf{x}(t), \qquad \overline{\kappa}(\overline{t})=\frac{K'}{K}\kappa(t).
\end{align}The scaled covariance $\overline{C}$ has eigenvalues $\overline{c}_{1,\dots,K}=\frac{K'}{K}c_{1,\dots,K}$, for which $\overline{c}_{1,\dots,K'}=\mathcal{O}(1)$ while $\overline{c}_{K'+1,\dots,K}=o(1).$ Under these scalings, \begin{align}
  \frac{\mathrm{d}\overline{W}}{\mathrm{d}\overline{t}}&= 
  \frac{1}{2}\left(\overline{C} - \frac{K'}{N}\overline{\mathbf{x}}(\overline{t})\overline{\mathbf{x}}(\overline{t})^\top\right) - \frac{\overline{\gamma}}{2}\overline{W}(\overline{t}) + \frac{\overline{\Omega}(\overline{t})}{\sqrt{N\overline{\eta}}},\\
  \frac{\mathrm{d}\overline{\mathbf{x}} }{\mathrm{d}\overline{t}}&= 
  \overline{\nu} \overline{W}(\overline{t})\overline{\mathbf{x}}(\overline{t}) - \overline{\kappa}(\overline{t})\overline{\mathbf{x}}(\overline{t}) + \sqrt{2\overline{\nu}}\,\overline{\boldsymbol{\xi}}(\overline{t}),
\end{align}where $\overline{\Omega}(\overline{t}),\overline{\boldsymbol{\xi}}(\overline{t})$, denote white noises in the scaled time $\overline{t}$. Following the same steps as in App. \ref{app:dynamics-msr}, \begin{align}
\frac{\mathrm{d}\overline{s}_k}{\mathrm{d}\overline{t}}
&=
-\overline{\kappa}(\overline{t})\,\overline{s}_k(\overline{t})
+
\frac{\overline{\nu}}{\overline{\gamma}}
\left(1-e^{-\overline{\gamma}\,\overline{t}/2}\right)
\overline{c}_k\,\overline{s}_k(\overline{t})
+
\int_{0}^{\overline{t}}
\overline{M}(\overline{t},\overline{u})\,
\overline{s}_k(\overline{u})\,
\mathrm{d}\overline{u},
\label{eq:scaled-signal}\\
\partial_{\overline{t}}\overline{R}(\overline{t},\overline{t}')
&=
-\overline{\kappa}(\overline{t})\,
\overline{R}(\overline{t},\overline{t}')
+
\int_{\overline{t}'}^{\overline{t}}
\overline{M}(\overline{t},\overline{u})\,
\overline{R}(\overline{u},\overline{t}')\,
\mathrm{d}\overline{u},
\label{eq:scaled-response}\\
\partial_{\overline{t}}\overline{Q}(\overline{t},\overline{t}')
&=
\frac{\overline{\nu}}{\overline{\gamma}}
\left(1-e^{-\overline{\gamma}\,\overline{t}/2}\right)
\sum_{k=1}^{K}
\overline{c}_k\,
\overline{s}_k(\overline{t})\,
\overline{s}_k(\overline{t}')
-
\overline{\kappa}(\overline{t})\,
\overline{Q}(\overline{t},\overline{t}')
\\
&\quad
+
\int_{0}^{\overline{t}}
\overline{M}(\overline{t},\overline{u})\,
\overline{Q}(\overline{u},\overline{t}')\,
\mathrm{d}\overline{u}
+
\int_{0}^{\overline{t}'}
\overline{D}(\overline{t},\overline{u})\,
\overline{R}(\overline{t}',\overline{u})\,
\mathrm{d}\overline{u}.
\end{align}where
\begin{align}
\overline{M}(\overline{t},\overline{u})
&=
\frac{K'^2}{K^{2}} M(t,u)
=
e^{-\overline{\gamma}(\overline{t}-\overline{u})/2}
\left\{
\frac{\overline{\nu}^{2}}{\overline{\eta}\,\overline{\gamma}}\,
\overline{R}(\overline{t},\overline{u})
-
\frac{K'\overline{\nu}}{2}\,
\overline{Q}(\overline{t},\overline{u})
\right\},
\label{eq:scaled-memory-kernel}\\
\overline{D}(\overline{t},\overline{u})
&=
\frac{K'^2}{K^{2}} D(t,u)
=
2\overline{\nu}\,\delta(\overline{t}-\overline{u})
+
\frac{\overline{\nu}^{2}}{\overline{\eta}\,\overline{\gamma}}\,
e^{-\overline{\gamma}|\overline{t}-\overline{u}|/2}
\overline{Q}(\overline{t},\overline{u}),
\end{align}and we introduced the scaled-time dynamical order parameters \begin{align}
\overline{s}_k(\overline{t}) = s_k(t),
\qquad
\overline{Q}(\overline{t},\overline{t}') = Q(t,t'),
\qquad
\overline{R}(\overline{t},\overline{t}') = R(t,t').    
\end{align}To complete the reduction, we show that \begin{align}
\sum_{k=1}^{K}
\overline{c}_k\,
\overline{s}_k(\overline{t})\,
\overline{s}_k(\overline{t}')
\approx
\sum_{k=1}^{K'}
\overline{c}_k\,
\overline{s}_k(\overline{t})\,
\overline{s}_k(\overline{t}').
\end{align}
Let \begin{align}
\Delta(\overline{t},\overline{t}')
=
\frac{1}{N}
\overline{\mathbf{x}}(\overline{t})^{\top}
\overline{B}
\overline{\mathbf{x}}(\overline{t}'),
\qquad
\text{with} \qquad 
\overline{B}
=
\frac{1}{N}
\sum_{a=K'+1}^{K}
\overline{c}_a\,
\mathbf{c}_a
\mathbf{c}_a^{\top},
\end{align}
be the remainder of the above approximation. By the spherical constraint: \begin{align}
    |\Delta(\overline{t},\overline{t}')|
\le
\|\overline{B}\|_{\mathrm{op}}
\sqrt{
\frac{\|\overline{\mathbf{x}}(\overline{t})\|^{2}}{N}
\frac{\|\overline{\mathbf{x}}(\overline{t}')\|^{2}}{N}
}
=
\|\overline{B}\|_{\mathrm{op}}=\overline{c}_{K'+1}=o(1).
\end{align} Thus $\Delta(\overline{t},\overline{t}')\to0$ as $K$ grows, uniformly in $\overline{t},\overline{t}'$ by the spherical constraint, and the claim follows. This reduction requires no restriction on the relative scaling of $N$ and $K$ beyond the stated spectral assumptions. Fig.~\ref{fig:largeK_nu_sweep} validates this scaled DMFT against finite-$N$ Langevin simulations on nearly finite-dimensional training data, sampled from the two-dimensional Ising model below the critical temperature ($K'=1$).

\subsubsection{Recovering the bare from the invariant large-$K$ regime\label{app:large-K-recovering}}

We now check that the bare large-$K$ regime of Sec.~\ref{app:bare-large-k} is recovered as a boundary limit of the invariant scaling of Sec.~\ref{app:invar-large-k}. 

Let
\[
    \varepsilon=\frac{K'}{K}\to0.
\]
If the bare variables $t,\gamma,\nu,\eta$ are kept $O(1)$, then in the invariant variables one has
\[
    \overline\gamma=\varepsilon\gamma\to0,
    \qquad
    \overline\nu=\varepsilon\nu\to0,
    \qquad
    \overline\eta=\frac{\eta}{\varepsilon}\to\infty,
    \qquad
    \overline t=\frac{t}{\varepsilon}\to\infty.
\]
This is the boundary of the invariant scaling corresponding to the bare regime.

Since the scaled covariance is $\overline C=\varepsilon C$, its leading eigenvalues translate across the two regimes as
\[
    \overline c_\ell
    =
    \varepsilon c_\ell
    =
    K'\widetilde{c}_\ell,
    \qquad
    \ell=1,\ldots,K' .
\]

We use
\[
    \overline s_\ell(\overline t)=s_\ell(t),
    \qquad
    \overline Q(\overline t,\overline t')=Q(t,t'),
    \qquad
    \overline R(\overline t,\overline t')=R(t,t'), \qquad 
    \overline\kappa(\overline t)=\varepsilon\kappa(t).
\]

Injecting \[
    \overline\gamma(\overline t-\overline u)
    =
    \gamma(t-u), 
    \qquad
    \frac{\overline\nu^2}{\overline{\eta}\,\overline\gamma}
    =
    \frac{\varepsilon^2\nu^2}{\eta\gamma}
    =
    O(\varepsilon^2),
    \qquad
    K'\overline\nu=K'\varepsilon\nu,
\] 
in the scaled memory kernel~\eqref{eq:scaled-memory-kernel} gives
\[
    \overline M(\overline t,\overline u)
    =
    -\frac{K'\varepsilon\nu}{2}
    e^{-\gamma(t-u)/2}Q(t,u)
    +O(\varepsilon^2).
\]
Since $\mathrm{d}\overline u=\mathrm{d}u/\varepsilon$, the memory integral has a finite limit
\[
    \int_0^{\overline t}\mathrm{d}\overline u\,
    \overline M(\overline t,\overline u)\overline s_\ell(\overline u)
    =
    -\frac{K'\nu}{2}
    \int_0^t\mathrm{d}u\,
    e^{-\gamma(t-u)/2}
    Q(t,u)s_\ell(u)
    +o(1).
\]
Using $\partial_{\overline t}=\varepsilon\partial_t$ and $\varepsilon\to0$ in Eq.~\eqref{eq:scaled-signal} gives the bare large-$K$ quasistatic signal equation, Eq.~\eqref{eq:bare-large-k-signal}.

The same reduction applies to the correlation equation. The smooth part of
$\overline D$ is $O(\varepsilon^2)$, and the white-noise term does not contribute
to the off-diagonal equation $t>t'$. From there, using $\partial_{\overline t}=\varepsilon\partial_t$ and $\varepsilon\to0$ once again gives back the bare large-$K$ quasistatic correlation equation, Eq.~\eqref{eq:bare-large-k-correlation}.

Applying similar manipulations turns Eq.~\eqref{eq:scaled-response} into Eq.~\eqref{eq:bare-large-k-response}, thereby concluding the reduction.

\clearpage
\section{Teacher-student scenario}\label{app:kl-divergences}

This appendix collects the four teacher--student KL divergences used in the main text for the rank-one teacher of Sec.~\ref{sec:teacher-student}.

\begin{table}[h]
  \caption{KL divergences. Columns (students): typical $W\sim\Prob_\eta$ vs.\ posterior predictive $\PredPost$. Rows (metrics): forward ($\Pteacher\|\cdot$) vs.\ reverse ($\cdot\|\Pteacher$) KL.}
  \label{tab:kl-table}
  \centering
  \begin{tabular}{@{}rcc@{}}
    \toprule
    & Typical $W\sim\Prob_\eta$ & Post. pred. $\PredPost$ \\
    \midrule
    Forward KL & $\langle D_{\mathrm{KL}}(\Pteacher\|\Prob_W)\rangle$ & $D_{\mathrm{KL}}(\Pteacher\|\PredPost)$ \\
    Reverse KL & $\langle D_{\mathrm{KL}}(\Prob_W\|\Pteacher)\rangle$ & $D_{\mathrm{KL}}(\PredPost\|\Pteacher)$ \\
    \bottomrule
  \end{tabular}
\end{table}

\subsection{Teacher-generated training data}\label{app:teacher-data}

The teacher is the rank-one SBM $\Wteacher=\omega^{\ast}\mathbf{w}^{\ast}\mathbf{w}^{\ast\top}-(\omega^{\ast}/N)I$ of Sec.~\ref{sec:teacher-student}, against which all four KL divergences are evaluated. For $\omega^{\ast}>1$, the teacher generates data that condenses along $\mathbf{w}^{\ast}$. For $K=2$, we have almost surely that $C$ has the two eigenvalues:
\begin{equation}\label{eq:teacher-eigenvalues}
c_{1} = 2-\frac{1}{\omega^{\ast}},\qquad c_{2} = \frac{1}{\omega^{\ast}}.
\end{equation}
The corresponding eigenvectors $\mathbf{c}_{1,2}$ satisfy:
\begin{equation}\label{eq:teacher-eigenvectors}
\frac{(\mathbf{c}_{1}\cdot\mathbf{w}^{\ast})^{2}}{N} = 1 - \frac{1}{2\omega^{\ast}-1},
\quad
\frac{(\mathbf{c}_{2}\cdot\mathbf{w}^{\ast})^{2}}{N} = 0
\end{equation}
to leading order in $N$. These deterministic limits also imply self-averaging of the four KL divergences per spin (that we compute below) over the training set. The teacher partition function and entropy read:
\begin{equation}\label{eq:lnZ-and-entropy-teacher-sbm-rank-1}
\frac{1}{N}\ln Z(\Wteacher) = \frac{1}{2}\ln(2\pi e) + 
\frac{\omega^{\ast}-1-\ln\omega^{\ast}}{2},\quad
\frac{1}{N}\mathcal{H}[\Pteacher] = \frac{1}{2}\ln(2\pi e) - \frac{1}{2}\ln\omega^{\ast}
\end{equation}
where we used $\ln Z(0)=\frac{N}{2}\ln(2\pi e)$.

\subsection{Posterior predictive as a ratio of evidence integrals}\label{app:pred-post-ratio}

We compute:
\begin{equation}
\begin{aligned}
&\PredPost(\mathbf{x}|C) = \langle\Prob(\mathbf{x}|W)\rangle_{W\sim \Prob_{\eta}} = \int \Prob(\mathbf{x}|W)\Prob_{\eta}(W|C)\mathrm{d}W \\
&= \frac{1}{\Evidence(C)}
\int
\exp \left\{ \frac{N}{2}\Trace\left[\left( \eta C + \frac{\mathbf{x}\mathbf{x}^{\top}}{N} \right)W\right] - (K\eta+1)\ln Z(W) - \frac{N}{4}\gamma\eta\Trace W^{2} \right\} \mathrm{d}W
\end{aligned}
\end{equation}
Defining $C_{\mathbf{x}}=C+\frac{1}{\eta}\frac{\mathbf{x}\mathbf{x}^{\top}}{N}$, so that $\eta\Trace C_{\mathbf{x}}=K\eta+1$, the integrand is $e^{\mathcal{L}(W;C_{\mathbf{x}})}$. Then
\begin{equation}\label{eq:pred-post-ratio}
  \PredPost(\mathbf{x}|C)
  = \frac{\Evidence(C_{\mathbf{x}})}{\Evidence(C)},
  \qquad
  C_{\mathbf{x}} = C + \frac{1}{\eta}\frac{\mathbf{x}\mathbf{x}^\top}{N}.
\end{equation}
This identity, noticed in \citep{tulinski2026}, reduces the posterior predictive to evaluating the evidence at a low-rank perturbation of $C$, so we can reuse the evidence formulas derived above rather than computing a separate $W$ integral.

\subsection{MAP and flat limits of the posterior predictive}
$P_{\mathrm{pp},\eta}(\mathbf{x}\vert C)$ interpolates between the flat measure on $\mathcal{S}_N$ for $\eta\downarrow0$ and the MAP likelihood for $\eta\uparrow\infty$.

\paragraph{Flat posterior predictive} In the $\eta\to0^+$ limit, the posterior predictive is uniform on the sphere, \begin{align}
    \lim_{\eta\to0^+}\frac{1}{N}\log P_{\mathrm{pp},\eta}(\mathbf{x}\vert C)=-\frac{1}{2}\log(2\pi e)+o_N(1).
\end{align}$P_{\mathrm{pp},\eta}(\mathbf{x}\vert C)$ depends on $\eta$ through $P_\eta(W\vert C)\propto e^{-\frac{N\gamma\eta}{4}\Trace W^2+\frac{N\eta}{2}\Trace CW-K\eta\ln Z(W)}$. 

We argue that, under the rescaling $W\mapsto \eta^{-1/2}\widetilde W$, and assuming that $\eta^{1/2}\Trace C\to0^+$, the dominant contribution to the exponent is $-\frac{N\gamma}{4}\Trace \widetilde{W}^2+o(1)$. To start, since $\vert\Trace C\widetilde{W}\vert\le\Trace C\,\Vert \widetilde{W}\Vert_{\mathrm{op}}$, we have \begin{align}
    \eta\Trace CW=\eta^{1/2}\Trace\widetilde{W}C=o(1)
\end{align} for a typical $\widetilde{W}$ with $\Vert\widetilde{W}\Vert_{\mathrm{op}}=\mathcal{O}(1)$. To bound $\eta\ln Z(\eta^{-1/2}\widetilde{W})$, use \begin{align}
    N\lambda_{\min}(W) \le \mathbf{x}^\top  W\mathbf{x}\le N\lambda_{\max}(W),
\end{align} which holds pointwise on $\mathcal{S}_N$. Exponentiating and integrating over $\mathcal{S}_N$ gives \begin{align}
    0<Z(0)e^{\frac{N}{2}\lambda_{\min}}\le Z(W)\le Z(0)e^{\frac{N}{2}\lambda_{\max}}.
\end{align}Taking logarithms and using $\vert\lambda_{\min}\vert,\vert\lambda_{\max}\vert\leq\Vert W\Vert_{\mathrm{op}}$, we obtain \begin{align}
    \vert \ln Z(W)-\ln Z(0)\vert\le\frac{N}{2}\Vert W\Vert_{\mathrm{op}}.
\end{align} Injecting $W=\eta^{-1/2}\widetilde{W}$, multiplying by $\eta$ and using $\eta \ln Z(0)=\mathcal{O}(\eta)$, \begin{align}
    \frac{\eta}{N}\log Z(\eta^{-1/2}\widetilde{W})=\mathcal{O}(\eta^{1/2}).
\end{align} Then $P_{0^+}(\widetilde{W}\vert C)\propto e^{-\frac{N\gamma}{4}\Trace\widetilde{W}^2+o(1)}$. The Jacobian from $W\mapsto \widetilde{W}$ is independent of $C$ and $\mathbf{x}$, and absorbed into the normalization of the posterior over $\widetilde{W}$. Using $P_{0^+}(O^\top\widetilde{W}O\vert C)=P_{0^+}(\widetilde{W}\vert C)$ and $P_{W}(O\mathbf{x})=P_{O^\top WO}(\mathbf{x})$ for any orthogonal matrix $O$, we have $P_{\mathrm{pp},0^+}(O\mathbf{x})=P_{\mathrm{pp},0^+}(\mathbf{x})$. Any normalized rotationally invariant density on $\mathcal{S}_N$ is constant, whence the claimed \begin{align}
    P_{\mathrm{pp},0^+}(\mathbf{x}\vert C)=\frac{1}{Z(0)}.
\end{align}

\paragraph{MAP posterior predictive} In the $\eta\to+\infty$ limit, $P_\eta(W\vert C)\Longrightarrow \delta(W-W_{\mathrm{MAP}})$, where $W_{\mathrm{MAP}}$ is the unique maximizer of $P_\eta(W\vert C)$. Hence \begin{align}
    \lim_{\eta\to+\infty}P_{\mathrm{pp},\eta}(\mathbf{x}\vert C)=P_{W_{\mathrm{MAP}}}(\mathbf{x}).
\end{align} Stationarity of the log-posterior in $W$ shows that $W_{\mathrm{MAP}}$ solves \begin{align}
    \gamma W_{\mathrm{MAP}}=C-\frac{K}{N}\langle\mathbf{x}\mathbf{x}^\top\rangle_{P_{W_\mathrm{MAP}}}.
\end{align} 

\subsection{Posterior-predictive entropy}\label{app:entropy-pp}

Applying~\eqref{eq:pred-post-ratio} to the posterior-predictive entropy gives
\begin{equation}\label{eq:entropy-pp}
\mathcal{H}[\PredPost] = \ln\Evidence(C) - \left\langle \ln\Evidence(C_{\mathbf{x}}) \right\rangle_{\mathbf{x}\sim\PredPost}.
\end{equation}
Under $\mathbf{x}\sim\PredPost$, we have the overlaps (by~\eqref{app:eq:x-statistics-avg-x})
\begin{equation}\label{eq:m2-pp}
m_{k}^{2} = \frac{(\mathbf{c}_{k}\cdot\mathbf{x})^{2}}{N^{2}} = \frac{c_{k}}{2} - \frac{\gamma}{2}\left(\chi_{k}-\frac{1}{\eta c_{k}}\right)
,\quad k=1,2,
\end{equation}
The three nonzero eigenvalues $\tilde c_{1,2,3}$ of $C_{\mathbf{x}}$ \eqref{eq:pred-post-ratio} are the roots of
\begin{equation}\label{eq:cubic-pp}
m_{1}^{2}\tilde c(\tilde c-c_{2}) + m_{2}^{2}\tilde c(\tilde c-c_{1}) + (1-m_{1}^{2}-m_{2}^{2}-\eta\tilde{c})(\tilde c-c_{1})(\tilde c-c_{2}) = 0,
\end{equation}
with $\tilde c_{1}+\tilde c_{2}+\tilde c_{3}=2+1/\eta$. Evaluating $\ln\Evidence$ in~\eqref{eq:entropy-pp} at leading order (App.~\ref{app:evidence}),
\begin{equation}\label{eq:entropy-pp-result}
\frac{1}{N}\mathcal{H}[\PredPost] = \Phi(c_{1},c_{2}) - \Phi(\tilde c_{1},\tilde c_{2},\tilde c_{3}).
\end{equation}

\subsection{Reverse cross-entropies}\label{app:kl-reverse}

Since the cross-entropy is linear in its first argument, the reverse cross-entropies of a typical $\Prob_{W}$ and the $\PredPost$ coincide:
\begin{equation}\label{eq:rev-cross-entropy-linear}
\left\langle \mathcal{H}[\Prob_{W};\Pteacher] \right\rangle_{W\sim\Prob_{\eta}}
= \mathcal{H}[ \langle \Prob_{W} \rangle_{W\sim\Prob_{\eta}} ;\Pteacher]
= \mathcal{H}[\PredPost;\Pteacher],
\end{equation}
so the posterior-average and posterior-predictive reverse cross-entropies coincide. Expanding,
\begin{equation}\label{eq:xent-rev}
\mathcal{H}[\PredPost;\Pteacher] = \left\langle \Energy(\mathbf{x};\Wteacher) \right\rangle_{\PredPost} + \ln Z(\Wteacher).
\end{equation}
For $K=2$ with the rank-one teacher $\Wteacher$, the sample statistics Eq. \eqref{app:eq:x-statistics-avg-x} combined with the leading-order identities $\frac{1}{N}\mathbf{c}_{2}^{\top}\Wteacher\mathbf{c}_{2}=0$ and $\frac{1}{N}\mathbf{c}_{1}^{\top}\Wteacher\mathbf{c}_{1}=2\omega^{\ast}(\omega^{\ast}-1)/(2\omega^{\ast}-1)$ yield
\begin{equation}\label{eq:avg-energy-teacher}
\frac{1}{N}\left\langle \Energy(\mathbf{x};\Wteacher) \right\rangle_{\PredPost}
= \left( \frac{\gamma}{\eta}\frac{\omega^{\ast 2}}{(2\omega^{\ast}-1)^{2}}(\eta c_{1}\chi_{1}-1) - 1 \right)\frac{\omega^{\ast}-1}{2},
\end{equation}
where $\chi_{1}$ is the saddle-point value from the spherical integral~\eqref{eq:hciz-N}.

\subsection{Forward cross-entropies}\label{app:kl-forward}

\textbf{Posterior-average forward cross-entropy.}
The forward cross-entropy of a typical sample $W$ is
\begin{equation}\label{eq:xent-fwd-typ}
\left\langle \mathcal{H}[\Pteacher;\Prob_{W}] \right\rangle_{W\sim\Prob_{\eta}}
=
\left\langle\left\langle \Energy(\mathbf{x};W) \right\rangle_{\mathbf x\sim\Pteacher}  + \ln Z(W) \right\rangle_{W\sim\Prob_{\eta}}.
\end{equation}
For $K=2$ with the rank-one teacher, the weight statistics Eq. \eqref{eq:W-mean} combined with \eqref{eq:teacher-eigenvectors} give
\begin{equation}\label{eq:fwd-avg-energy}
\frac{1}{N}\left\langle \left\langle \Energy(\mathbf{x};W) \right\rangle_{\mathbf x\sim\Pteacher} \right\rangle_{W\sim\Prob_{\eta}}
= -\left(c_{1}\chi_{1} - \frac{1}{\eta}\right)\frac{(\omega^{\ast}-1)^{2}}{(2\omega^{\ast}-1)^{2}}.
\end{equation}

\textbf{Posterior-predictive forward cross-entropy.}
From \eqref{eq:pred-post-ratio},
\begin{equation}\label{eq:forward-xent-pp}
\mathcal{H}[\Pteacher;\PredPost] = \ln\Evidence(C) - \left\langle \ln\Evidence(C_{\mathbf{x}}) \right\rangle_{\mathbf{x}\sim \Pteacher}.
\end{equation}
The eigenvalues of $C_{\mathbf{x}}$ when $\mathbf{x}\sim\Pteacher$ concentrate (for $K=2$) to three deterministic values $(\lambda_{+},\lambda_{-},c_{2})$ where
\begin{equation}\label{eq:lambda-pm}
\lambda_{\pm}
= \frac{1}{2}\left\{ c_{1}+\frac{1}{\eta}\pm\sqrt{ \left( c_{1}-\frac{1}{\eta} \right)^{2}+\frac{2}{\eta}(c_{1}-c_{2})^{2} } \right\},
\end{equation}
with $\lambda_{+}+\lambda_{-}+c_{2}=2+1/\eta$.

\subsection{KL divergences}\label{app:kl-results}

We now assemble the four KL divergences from the cross-entropies above and the entropies of $\Pteacher$ and $\Prob_{W}$ in \eqref{eq:lnZ-and-entropy-teacher-sbm-rank-1} and \eqref{eq:SBM-entropy}. Differences of $\ln\Evidence$ are taken at leading $\BigO(N)$ from App.~\ref{app:evidence}.

\textbf{Posterior-average reverse KL.} Combining~\eqref{eq:xent-rev} with~\eqref{eq:SBM-entropy},
\begin{equation}\label{eq:reverse-kl-result}
\frac{1}{N}\left\langle D_{\mathrm{KL}}(\Prob_{W}\|\Pteacher) \right\rangle_{W\sim\Prob_{\eta}} =
\frac{1}{N}\left\langle \Energy(\mathbf{x};\Wteacher) \right\rangle_{\PredPost} + \frac{1}{2}\left( \omega^{\ast}-1-\ln\omega^{\ast}+\LogStieltjes(\mu) \right).
\end{equation}
with $\langle\Energy(\mathbf{x};\Wteacher)\rangle_{\PredPost}$ from~\eqref{eq:avg-energy-teacher}.

\textbf{Posterior-predictive reverse KL.} Combining~\eqref{eq:xent-rev} with~\eqref{eq:entropy-pp-result},
\begin{equation}\label{eq:reverse-kl-pp}
\begin{aligned}
\frac{1}{N}D_{\mathrm{KL}}(\PredPost\|\Pteacher)
&=\frac{1}{N}\left\langle \Energy(\mathbf{x};\Wteacher) \right\rangle_{\PredPost} + \frac{\ln(2\pi)+\omega^{\ast}-\ln\omega^{\ast}}{2}\\
&\quad+\Phi(\tilde c_{1},\tilde c_{2},\tilde c_{3}) - \Phi(c_{1},c_{2}),
\end{aligned}
\end{equation}
with $\langle\Energy(\mathbf{x};\Wteacher)\rangle_{\PredPost}$ from~\eqref{eq:avg-energy-teacher}.

\textbf{Posterior-average forward KL.}
Combining~\eqref{eq:fwd-avg-energy} with the teacher entropy $\mathcal{H}[\Pteacher]/N=\tfrac{1}{2}\ln(2\pi e)-\tfrac{1}{2}\ln\omega^{\ast}$ from~\eqref{eq:lnZ-and-entropy-teacher-sbm-rank-1} and the posterior-average $\langle\ln Z(W)\rangle_{W\sim\Prob_{\eta}}/N$ from~\eqref{eq:SBM-log-Z} (the $\tfrac{1}{2}\ln(2\pi)$ base-measure constant cancels between the two),
\begin{equation}\label{eq:forward-kl-typ}
\frac{1}{N}\left\langle D_{\mathrm{KL}}(\Pteacher\|\Prob_{W}) \right\rangle_{W\sim\Prob_{\eta}} =
-\frac{\eta c_{1}\chi_{1}-1}{\eta}\frac{(\omega^{\ast}-1)^{2}}{(2\omega^{\ast}-1)^{2}} + \frac{\mu-\LogStieltjes(\mu)}{2} - \frac{1-\ln\omega^{\ast}}{2}.
\end{equation}

\textbf{Posterior-predictive forward KL.}
Subtracting the teacher entropy from~\eqref{eq:forward-xent-pp} and evaluating $\ln\Evidence$ at the eigenvalues~\eqref{eq:lambda-pm},
\begin{equation}\label{eq:forward-kl-pp}
\frac{1}{N}D_{\mathrm{KL}}(\Pteacher\|\PredPost) =
-\frac{\ln(2\pi e)-\ln\omega^{\ast}}{2} - \Phi(\lambda_{+},\lambda_{-},c_{2}) + \Phi(c_{1},c_{2}),
\end{equation}
with $\Phi$ the $\BigO(N)$ coefficient of the evidence (App.~\ref{app:evidence}). The optimum $\eta_{\mathrm{pp}}^{\mathrm{fwd}}=\arg\min_{\eta>0}D_{\mathrm{KL}}(\Pteacher\Vert\PredPost)$ admits closed-form thresholds in the $(\omega^{\ast},\gamma)$ plane derived in App.~\ref{app:fwd-pp-temperature}.

\subsection{Posterior-predictive gap identities}\label{app:kl-gaps}

We show that:
\begin{align}
  \left\langle D_{\mathrm{KL}}(\Prob_W\|\Pteacher)\right\rangle_{W\sim \Prob_{\eta}} - D_{\mathrm{KL}}(\PredPost\|\Pteacher) &= \mathrm{MI}(W;\mathbf{x}|\mathcal{D}) \ge 0,\label{eq:reverse-gap} \\
  \left\langle D_{\mathrm{KL}}(\Pteacher\|\Prob_W)\right\rangle_{W\sim \Prob_{\eta}} - D_{\mathrm{KL}}(\Pteacher\|\PredPost) &= \bigl\langle D_{\mathrm{KL}}(\Prob_\eta(W|\mathcal{D})\|\Prob_\eta(W|\mathcal{D},\mathbf{x}))\bigr\rangle_{\mathbf{x}\sim\Pteacher} \ge 0,\label{eq:forward-gap}
\end{align}
proving the inequalities stated in Sec.~\ref{sec:teacher-student} of the main text. These gaps vanish in the MAP limit.

\textbf{Reverse KL gap.} Using \eqref{eq:rev-cross-entropy-linear}:
\begin{equation}
\left\langle D_{\mathrm{KL}}(\Prob_{W}\|\Pteacher)\right\rangle - D_{\mathrm{KL}}(\PredPost\|\Pteacher) = \mathcal{H}[\PredPost] - \left\langle \mathcal{H}[\Prob_{W}]\right\rangle_{W\sim\Prob_{\eta}}
\end{equation}
The right-hand side equals the mutual information $\mathrm{MI}(W;\mathbf{x}|\mathcal{D})$ between the posterior weights $W$ and student-generated data $\mathbf{x}$:
\begin{equation}
\begin{aligned}
\mathrm{MI}(W;\mathbf{x}|\mathcal{D}) &= \int \Prob_{\eta}(W|\mathcal{D})\Prob(\mathbf{x}|W)\ln\frac{\Prob(\mathbf{x}|W)}{\PredPost(\mathbf{x}|\mathcal{D})}\mathrm{d}W\mathrm{d}\mathbf{x} \\
&= \mathcal{H}[\PredPost] - \left\langle \mathcal{H}[\Prob_{W}]\right\rangle_{W\sim\Prob_{\eta}} \geq 0
\end{aligned}
\end{equation}
establishing \eqref{eq:reverse-gap}.

\textbf{Forward KL gap.} We compute:
\begin{equation}
\left\langle D_{\mathrm{KL}}(\Pteacher\|\Prob_{W})\right\rangle_{W\sim\Prob_{\eta}} - D_{\mathrm{KL}}(\Pteacher\|\PredPost) 
= \int\mathrm{d}W\, \Prob_{\eta}(W|\mathcal{D})\int \mathrm{d}\mathbf{x}\,\Pteacher(\mathbf{x})\ln\frac{\PredPost(\mathbf{x}|\mathcal{D})}{\Prob(\mathbf{x}|W)}
\end{equation}
Using the Bayes update $\Prob_{\eta}(W|\mathcal{D},\mathbf{x})=\Prob(\mathbf{x}|W)\Prob_{\eta}(W|\mathcal{D})/\PredPost(\mathbf{x}|\mathcal{D})$, the inner integral over $W$ becomes $D_{\mathrm{KL}}\!\left(\Prob_{\eta}(W|\mathcal{D})\,\middle\|\,\Prob_{\eta}(W|\mathcal{D},\mathbf{x})\right)$, yielding \eqref{eq:forward-gap}.

\textbf{Discussion.} Both identities are classical Bayesian facts from two largely disjoint literatures. The reverse gap is the mutual information $\mathrm{MI}(W;\mathbf{x})$, equivalent to Lindley's expected-information gain~\citep{lindley1956measure} and to the epistemic/aleatoric decomposition of predictive entropy used in Bayesian deep learning~\citep{houlsby2011bayesian,depeweg2018decomposition,lakshminarayanan2017simple}. The forward gap goes back to~\citep{aitchison1975goodness}, who showed that $\PredPost$ dominates typical $\Prob_W$ under KL loss. The distinction between posterior predictive and a typical posterior sample is mainstream in feed-forward Bayesian models but, to our knowledge, has not been emphasized for generative models.

\subsection{Sampling temperature tuning\label{app:temperature-tuning-theory}}

We focus on the effect of sampling temperature tuning on the forward KL. We have:
\begin{equation}\label{eq:D_KL_beta}
D_{\mathrm{KL}}(\Pteacher\|\Prob_{\beta W}) = 
\beta\mathcal{H}[\Pteacher;\Prob_{W}] - \mathcal{H}[\Pteacher] + \ln Z(\beta W) - \beta\ln Z(W)
\end{equation}
This is a convex function of $\beta$, which therefore has a single minimum:
\begin{equation}\label{eq:beta_TT}
\beta_\mathrm{TT} = \operatorname*{argmin}_{\beta} D_{\mathrm{KL}}(\Pteacher\|\Prob_{\beta W})
\end{equation}
To determine whether the optimal $\beta_\mathrm{TT}$ is $>1$ or $<1$, it suffices to compute the sign of:
\begin{equation}\label{eq:D_KL_beta-der}
\left. \frac{ \partial D_{\mathrm{KL}}(\Pteacher\|\Prob_{\beta W}) }{ \partial \beta }  \right|_{\beta=1} = \mathcal{H}[\Pteacher;\Prob_W] - \mathcal{H}[\Prob_W]
\end{equation}
$\beta_\mathrm{TT}$ is $<1$ or $>1$ according to whether \eqref{eq:D_KL_beta-der} is positive or negative, respectively. Note that equations \eqref{eq:D_KL_beta} and \eqref{eq:D_KL_beta-der} are general identities, valid for any temperature-tuned energy-based model.

Now we specialize to the teacher-student scenario introduced in the main text, with $K=2$ and the rank-one teacher. Averaging over the posterior $W\sim\Prob_{\eta}$ and substituting the forward average energy~\eqref{eq:fwd-avg-energy} and the model average energy~\eqref{eq:SBM-avg-energy} gives, to leading order in $N$,
\begin{align}
\frac{1}{N}\left. \frac{ \partial D_{\mathrm{KL}}(\Pteacher\|\Prob_{\beta W}) }{ \partial \beta }  \right|_{\beta=1} 
&= \frac{\mu-1}{2} - \left( \frac{\omega^{\ast}-1}{2\omega^{\ast}-1} \right)^{2} \left( c_{1}\chi_{1}-\frac{1}{\eta} \right) \\
&= \begin{cases}
\frac{1}{2\gamma\eta} - \frac{1}{\gamma}\left( \frac{\omega^{\ast}-1}{\omega^{\ast}} \right)^{2} & h=0 \\
\frac{1}{\sqrt{\gamma\eta}} - \frac{1}{2} - \frac{1}{\gamma} \left( \frac{\omega^{\ast}-1}{\omega^{\ast}} \right)^{2} & h\ne0,u_{1}=0 \\
\frac{\lambda_{1}-1}{2} - \left( \frac{\omega^{\ast}-1}{2\omega^{\ast}-1} \right)^{2}
\left( c_{1}\lambda_{1}-\frac{1}{\eta} \right) & h,u_1\ne 0
\end{cases}
\end{align}
after substituting the saddle-point values for $\mu$ and $\chi_{1}$ in each phase.

After temperature tuning, the model can be transported from one phase to another.
\begin{itemize}
    \item If $h=u_{1}=0$, then $\beta_{\mathrm{TT}}$ is $<1$ or $>1$ according to whether $2\eta[(\omega^{\ast}-1)/\omega^{\ast}]^{2}$ is $<1$ or $>1$. In this case, the temperature-tuned model remains in the initial phase (with $h=u_{1}=0$).
    \item If $h\ne0$ and $u_{1}=0$, then $\beta_{\mathrm{TT}}<1$ is such that the temperature-tuned model lands in a phase with $h=u_{1}=0$, killing the condensed random direction.
    \item If $h=0$ and $u_{1}\ne0$, $\beta_\mathrm{TT}>1$ or $<1$ according to whether $\frac{\omega^{\ast 2}}{2(\omega^{\ast}-1)^{2}}$ is $<\eta$ or $>\eta$. If $\frac{(\omega^{\ast}-1)^{2}(2\omega^{\ast}-1)}{\omega^{\ast}}<\frac{\gamma}{2\eta}$ the model stays in the same phase ($h=0,u_{1}\ne0$), but otherwise it activates the dormant aligned mode, landing in $h,u_{1}\ne0$, which is beneficial because this mode was already aligned to the teacher.
    \item If $h,u_{1}\ne0$, both $\beta_\mathrm{TT}>1$ or $<1$ are possible, and the model can stay in the same phase or move to $h=0,u_{1}\ne0$. 
\end{itemize}
Generally, since $\beta$ only rescales the spectrum of $W$ without deforming it, temperature tuning cannot create new data-aligned outliers or merge two distinct outliers. Temperature tuning can only activate or deactivate an already present aligned mode. 

\subsection{Double descent}\label{app:double-descent-kl}

We derive the threshold $\eta_{\mathrm{DD}}(\omega^{\ast})$ above which the typical reverse KL of Fig.~\ref{fig:double-descent} acquires a local minimum inside the data-aligned condensed phase $h\ne0,u_{1}\ne0$, creating the first descent of the double-descent profile. The post-condensation $h=0$ branch is governed by
\begin{equation}\label{eq:rkl-para}
  \tfrac{1}{N}
  \left\langle D_{\mathrm{KL}}(\Prob_W\|\Pteacher)\right\rangle_{W\sim \Prob_{\eta}} = \frac{\omega^{\ast}-1-\ln\omega^{\ast}}{2} + \frac{1}{4\gamma\eta},
\end{equation}
with $\gamma$-derivative $-1/(4\eta\gamma^{2})<0$ strictly negative for any finite $\eta$; the second descent therefore exists trivially, and the analysis reduces to the condensed branch. We work in the $K=2$ rank-one teacher of App.~\ref{app:teacher-data} with eigenvalues $c_{1}=2-1/\omega^{\ast}$, $c_{2}=1/\omega^{\ast}$ from~\eqref{eq:teacher-eigenvalues}.

On the condensed branch the top-outlier Stieltjes transform $g_{1}=\Stieltjes(\lambda_{1})$ provides a smooth parametrization,
\begin{equation}\label{eq:dd-gamma-of-g}
  \gamma(g_{1})=2g_{1}^{2}-\frac{g_{1}}{\omega^{\ast}}, \qquad
  \lambda_{1}=\frac{1}{g_{1}} + \frac{g_{1}}{\gamma\eta}, \qquad
  \frac{1}{\omega^{\ast}} < g_{1} < 1,
\end{equation}
with the outlier-detachment condition $g_{1}<\sqrt{\gamma\eta}$ equivalent to $\eta>\eta_{F}(g_{1}):=\omega^{\ast}g_{1}/(2\omega^{\ast}g_{1}-1)$, and the branch terminating at the condensation threshold $g_{1}=1$, $\gamma=c_{1}$. Substituting~\eqref{eq:dd-gamma-of-g} into~\eqref{eq:reverse-kl-result} and using $\Stieltjes(\mu)=1-h^{2}=g_{1}$ on this branch yields the intensive reverse KL as a function of $g_{1}$,
\begin{multline}\label{eq:dd-rkl-cond}
\tfrac{1}{N}\bigl\langle D_{\mathrm{KL}}(\Prob_{W}\Vert\Pteacher)\bigr\rangle_{W\sim\Prob_{\eta}}
  = \\
\frac{\omega^{\ast}-1-\ln(\omega^{\ast}g_{1})}{2}
  -\frac{\omega^{\ast}(\omega^{\ast}-1)(1-g_{1})}{2\omega^{\ast}-1}\!\left[1-\frac{\omega^{\ast}g_{1}}{\eta(2\omega^{\ast}-1)}\right]
  +\frac{\omega^{\ast}g_{1}}{4\eta(2\omega^{\ast}g_{1}-1)},
\end{multline}
which matches~\eqref{eq:rkl-para} continuously at $g_{1}=1$.

Since $\mathrm{d}\gamma/\mathrm{d}g_{1}=4g_{1}-1/\omega^{\ast}>0$ on this branch, extrema in $\gamma$ and in $g_{1}$ coincide. With
\begin{equation}\label{eq:gMAP-def}
g_{\mathrm{MAP}}:=\frac{2\omega^{\ast}-1}{2\omega^{\ast}(\omega^{\ast}-1)},
\end{equation}
the $g_{1}$-derivative writes,
\begin{multline}\label{eq:dd-deriv-decomp}
  \frac{\mathrm{d}}{\mathrm{d}g_{1}}\!\left[\tfrac{1}{N}\bigl\langle D_{\mathrm{KL}}(\Prob_{W}\Vert\Pteacher)\bigr\rangle_{W\sim\Prob_{\eta}}\right]
  = \frac{\omega^{\ast}(\omega^{\ast}-1)(g_{1}-g_{\mathrm{MAP}})}{g_{1}(2\omega^{\ast}-1)}\\
  +\frac{1}{\eta}\!\left[\frac{\omega^{\ast 2}(\omega^{\ast}-1)(1-2g_{1})}{(2\omega^{\ast}-1)^{2}}-\frac{\omega^{\ast}}{4(2\omega^{\ast}g_{1}-1)^{2}}\right].
\end{multline}
In the MAP limit only the first term survives, and is positive on $(g_{\mathrm{MAP}},1)$ when $g_{\mathrm{MAP}}<1$, equivalently $2\omega^{\ast 2}-4\omega^{\ast}+1>0$, equivalently
\begin{equation}\label{eq:dd-strong-teacher}
  \omega^{\ast}>\omega_{\mathrm{DD}} := 1+\frac{1}{\sqrt{2}}
\end{equation}
Below $\omega_{\mathrm{DD}}$ the condensed branch is monotone for every $\eta$ and no first minimum exists.

When~\eqref{eq:dd-strong-teacher} holds, an interior local minimum appears at $g_{1}\in(g_{\mathrm{MAP}},1)$ as soon as the slope~\eqref{eq:dd-deriv-decomp} is positive and the outlier-detachment condition $\eta>\eta_{F}(g_{1})$ holds. The sharp threshold is the one-dimensional variational formula
\begin{multline}\label{eq:eta-dd}
  \eta_{\mathrm{DD}}(\omega^{\ast})
  =\inf_{g_{\mathrm{MAP}}<g_{1}<1}\,\frac{\omega^{\ast}g_{1}}{2\omega^{\ast}g_{1}-1} \times \\
  \max\!\left\{1,\;\frac{2\omega^{\ast}g_{1}-1}{(2\omega^{\ast}-1)(g_{1}-g_{\mathrm{MAP}})}\!\left[(2g_{1}-1)+\frac{(2\omega^{\ast}-1)^{2}}{4\omega^{\ast}(\omega^{\ast}-1)(2\omega^{\ast}g_{1}-1)^{2}}\right]\right\},
\end{multline}
with $\eta_{\mathrm{DD}}(\omega^{\ast})=\infty$ for $\omega^{\ast}\le\omega_{\mathrm{DD}}$. For $\eta_{\mathrm{DD}}<\eta<\infty$ the reverse KL has the double-descent profile of Fig.~\ref{fig:double-descent}; the MAP limit collapses the post-transition branch to a flat line, so the second descent ceases to be strict while the first minimum survives.

\subsection{Tempered-posterior effects}\label{app:tempered-posterior-analytics}

The optimal posterior temperature $\eta_{\mathrm{pp}}$ can be defined by minimising either the reverse KL $D_{\mathrm{KL}}(\PredPost\Vert\Pteacher)$ or the forward KL $D_{\mathrm{KL}}(\Pteacher\Vert\PredPost)$ over $\eta>0$. In both cases the $(\omega^{\ast},\gamma)$ plane splits into warm ($\eta_{\mathrm{pp}}<1$), cold ($\eta_{\mathrm{pp}}>1$), and MAP ($\eta_{\mathrm{pp}}\to\infty$) regions, and the two definitions agree at the qualitative level: increasing $\omega^{\ast}$ or $\gamma$ shifts the optimum from warm to cold under either KL. The phase boundaries and the structure of the optimum nevertheless differ quantitatively, as we work out in the following two subsections.

\subsubsection{Optimal tempered-posterior according to the reverse KL}\label{app:rev-pp-temperature}

We minimise~\eqref{eq:reverse-kl-pp} over $\eta>0$ at fixed $(\omega^{\ast},\gamma)$ and obtain the five regions of Fig.~\ref{fig:warm-cold-pp}A. On a smooth condensed branch, evaluating $\Phi(\tilde c_{1},\tilde c_{2},\tilde c_{3})$ along the $m_{1,2}^{2}(\eta)$-parametrisation determined by~\eqref{eq:cubic-pp} reduces~\eqref{eq:reverse-kl-pp} to
\begin{equation}\label{eq:rev-pp-reduced}
\tfrac{1}{N}D_{\mathrm{KL}}(\PredPost\Vert\Pteacher)=\frac{\omega^{\ast}-1-\ln\omega^{\ast}}{2}-\frac{m_{1}^{2}}{2g_{\mathrm{MAP}}}-\frac{1}{2}\ln(1-m_{1}^{2}-m_{2}^{2}),
\end{equation}
with $g_{\mathrm{MAP}}$ as in~\eqref{eq:gMAP-def}; all $(\gamma,\eta)$-dependence of the right-hand side is carried by $m_{1}^{2}, m_{2}^{2}$, while $\omega^{\ast}$ and $g_{\mathrm{MAP}}$ are teacher-only constants. The term $m_{1}^{2}/(2g_{\mathrm{MAP}})$ is the energetic gain along the teacher-aligned mode; $m_{2}^{2}$ enters only through the entropy because the second empirical mode is teacher-orthogonal. Differentiating in $\eta$ gives the stationarity condition
\begin{equation}\label{eq:rev-pp-stationarity}
\partial_{\eta}m_{1}^{2}=\frac{g_{\mathrm{MAP}}\bigl(\partial_{\eta}m_{1}^{2}+\partial_{\eta}m_{2}^{2}\bigr)}{1-m_{1}^{2}-m_{2}^{2}}.
\end{equation}

\textbf{Zero-overlap regions.} The bound $-\ln(1-s)\geq s$ on $s=m_{1}^{2}+m_{2}^{2}$ applied to~\eqref{eq:rev-pp-reduced} gives $D_{\mathrm{KL}}(\PredPost\Vert\Pteacher)/N-(\omega^{\ast}-1-\ln\omega^{\ast})/2\geq[(1-1/g_{\mathrm{MAP}})m_{1}^{2}+m_{2}^{2}]/2$, so for $g_{\mathrm{MAP}}\geq 1$, equivalently $\omega^{\ast}\leq\omega_{0}:=1+1/\sqrt{2}$, the leading reverse KL is minimised at $m_{1}^{2}=m_{2}^{2}=0$ and equals the uniform-student baseline $D_{\mathrm{KL}}(\Prob_{W=0}\Vert\Pteacher)/N=(\omega^{\ast}-1-\ln\omega^{\ast})/2$ throughout the zero-overlap sector. For $\gamma<c_{1}$ this sector is the warm interval $\eta\in(0,\gamma/c_{1}^{2}]$ and the optimum is degenerate and warm (\emph{warm deg.}\ in Fig.~\ref{fig:warm-cold-pp}A). For $\gamma\geq c_{1}$, at any teacher strength, both overlaps vanish for all $\eta>0$ and the leading reverse KL equals the uniform-student baseline throughout (\emph{deg.}).

\textbf{Single-mode branch.} For $\omega^{\ast}>\omega_{0}$ and $\gamma<c_{1}$ a positive-overlap optimum exists. On the $d=1$ outlier branch, $m_{2}^{2}\equiv 0$, and~\eqref{eq:rev-pp-stationarity} reduces to the scalar target
\begin{equation}\label{eq:rev-pp-q1star}
(m_{1}^{2})^{\ast}=1-g_{\mathrm{MAP}}.
\end{equation}
Tracking $m_{1}^{2}(\eta)$ along the smooth branch with the $g_{1}$-parametrisation of~\eqref{eq:fwd-pp-eta0}, the value $\eta_{\mathrm{pp}}^{\mathrm{rev}}=1$ is hit on
\begin{equation}\label{eq:rev-pp-gwc}
\gamma_{\rm wc}^{(1)}(\omega^{\ast})=2g_{\rm wc}^{2}-c_{2}\,g_{\rm wc},\qquad
g_{\rm wc}=\tfrac{1}{4}\!\left[(c_{2}+3c_{1})-\sqrt{(c_{2}+3c_{1})^{2}-16c_{1}g_{\mathrm{MAP}}}\right],
\end{equation}
and $\eta_{\mathrm{pp}}^{\mathrm{rev}}\to\infty$ on
\begin{equation}\label{eq:rev-pp-ginf}
\gamma_{\infty}^{\mathrm{rev}}(\omega^{\ast})=2g_{\mathrm{MAP}}^{2}-c_{2}\,g_{\mathrm{MAP}}=\frac{2\omega^{\ast}-1}{2\omega^{\ast}(\omega^{\ast}-1)^{2}}.
\end{equation}

\textbf{Two-mode correction.} The $d=2$ co-condensed branch exists only for $\gamma<c_{2}^{2}$ and pre-empts the $d=1$ optimum when its turn-on tangent is downhill, $g_{\mathrm{MAP}}<1-(c_{1}-c_{2})^{2}/(2c_{1})$, equivalently $\omega^{\ast}>\omega_{2}:=(3+\sqrt{3})/2$. Substituting the edge-branch overlaps $m_{i}^{2}=(c_{i}/2)(1-\sqrt{\gamma/\eta}/c_{i})^{2}$ into~\eqref{eq:rev-pp-stationarity} fixes the warm/cold boundary at $\gamma_{\rm wc}^{(2)}=\alpha^{2}$, with $\alpha\in(0,c_{2})$ the unique root of
\begin{equation}\label{eq:rev-pp-d2}
\alpha\,(c_{1}-\alpha)(2c_{1}c_{2}-\alpha)=2c_{1}\,g_{\mathrm{MAP}}\,(c_{1}c_{2}-\alpha).
\end{equation}
The cold/MAP boundary~\eqref{eq:rev-pp-ginf} is unchanged.

\textbf{Phase classification.} For $\omega^{\ast}\leq\omega_{0}$ the optimum is the zero-overlap plateau (\emph{warm deg.}\ for $\gamma<c_{1}$, \emph{deg.}\ for $\gamma\geq c_{1}$). For $\omega^{\ast}>\omega_{0}$, set $\gamma_{\rm wc}=\gamma_{\rm wc}^{(1)}$ when $\omega^{\ast}\leq\omega_{2}$ and $\gamma_{\rm wc}=\gamma_{\rm wc}^{(2)}$ when $\omega^{\ast}>\omega_{2}$; the regions are then \emph{warm} for $\gamma<\gamma_{\rm wc}$, \emph{cold} for $\gamma_{\rm wc}<\gamma<\gamma_{\infty}^{\mathrm{rev}}$, \emph{MAP} for $\gamma_{\infty}^{\mathrm{rev}}\leq\gamma<c_{1}$, and \emph{deg.}\ for $\gamma\geq c_{1}$. At the MAP optimum, the predictive reverse KL equals the $\eta\to\infty$ limit of the general typical-posterior expression~\eqref{eq:reverse-kl-result}.

\subsubsection{Optimal tempered-posterior according to the forward KL}\label{app:fwd-pp-temperature}

\begin{figure}[h]
  \centering
  \includegraphics[width=0.8\linewidth]{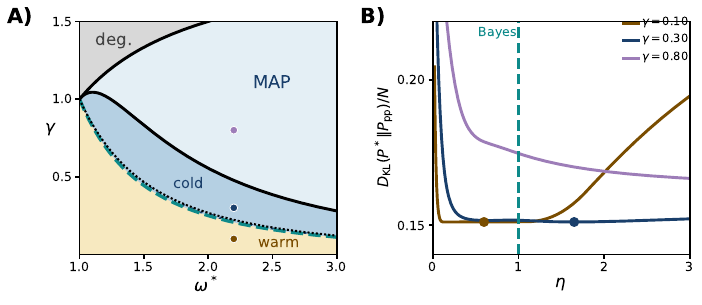}
  \caption{\textbf{Optimum $\eta_{\mathrm{pp}}^{\mathrm{fwd}}=\arg\min_{\eta>0}D_{\mathrm{KL}}(\Pteacher\Vert\PredPost)$ for the rank-one $K{=}2$ teacher.}
  \textbf{A)} Phases in the $(\omega^{\ast},\gamma)$ plane: warm flat (yellow), unique cold (medium blue), MAP (light blue), and degenerate (gray), where all $\eta\geq\eta_{\mathrm{PM}}$ attain the MAP value (App.~\ref{app:fwd-pp-temperature}); the hatched sliver is the mixed warm/cold tie. Boundaries: dashed teal $\gamma_{\rm wc}$ (Bayes-crossing, $\eta_{\mathrm{pp}}^{\mathrm{fwd}}=1$), dotted black $\gamma_{\rm flat}$ (upper edge of the warm flat interval), solid black $\gamma_{\infty}$ (cold/MAP) and $\gamma=c_{1}$ (MAP/degenerate).
  \textbf{B)} $D_{\mathrm{KL}}(\Pteacher\Vert\PredPost)/N$ vs.\ $\eta$ at $\omega^{\ast}=2.2$ for $\gamma\in\{0.10,0.30,0.80\}$ marked in panel~A; teal dashed at $\eta=1$.}
  \label{fig:warm-cold-pp-fwd}
\end{figure}

We minimise~\eqref{eq:forward-kl-pp} over $\eta>0$ at fixed $(\omega^{\ast},\gamma)$:
\begin{equation}
\eta_{\mathrm{pp}}^{\mathrm{fwd}}=\arg\min_{\eta>0}D_{\mathrm{KL}}(\Pteacher\Vert\PredPost)
\end{equation}
and obtain three closed-form thresholds, $\gamma_{\rm wc}<\gamma_{\rm flat}<\gamma_{\infty}$. Together with $\gamma=c_{1}$, they partition the $(\omega^{\ast},\gamma)$ plane into the five regions of Fig.~\ref{fig:warm-cold-pp-fwd}A.

\textbf{Smooth branch.} On the condensed branch outside the warm flat regime defined below, the perturbed and original sources carry the same number of outliers and the $\eta$-dependence of~\eqref{eq:forward-kl-pp} does not cancel. The interior stationary condition $\partial_{\eta}D_{\mathrm{KL}}(\Pteacher\Vert\PredPost)=0$ reduces to matching the evidence saddle variable $g_{1}$ between $\mathcal Y(C_{\mathbf{x}})$ and $\mathcal Y(C)$, namely $g_{1}^{(C_{\mathbf{x}})}=g_{1}^{(C)}$, which using~\eqref{eq:lambda-pm} simplifies to $\lambda_{-}=g_{1}/\eta$. Combined with the elementary identities $\lambda_{+}\lambda_{-}=\bigl[c_{1}-\tfrac{1}{2}(c_{1}-c_{2})^{2}\bigr]/\eta$ and $\lambda_{+}+\lambda_{-}=c_{1}+1/\eta$, this gives the smooth-branch stationary point $\eta_{0}(\gamma,\omega^{\ast})$ of the forward KL,
\begin{equation}\label{eq:fwd-pp-eta0}
\eta_{0}(\gamma,\omega^{\ast})=\frac{g_{1}\,(1-g_{1})}{c_{1}(1-g_{1})-\tfrac{1}{2}(c_{1}-c_{2})^{2}},\qquad
g_{1}(\gamma,\omega^{\ast})=\frac{c_{2}+\sqrt{c_{2}^{2}+8\gamma}}{4}.
\end{equation}
Since $g_{1}$ is monotone in $\gamma$, $\eta_{0}$ diverges at $g_{1}=1-(c_{1}-c_{2})^{2}/(2c_{1})$. Substituting back into $\gamma=2g_{1}^{2}-c_{2}g_{1}$ yields the cold/MAP boundary
\begin{equation}\label{eq:fwd-pp-gamma-inf}
\gamma_{\infty}(\omega^{\ast})=2\!\left[1-\tfrac{(c_{1}-c_{2})^{2}}{2c_{1}}\right]^{2}-c_{2}\!\left[1-\tfrac{(c_{1}-c_{2})^{2}}{2c_{1}}\right]=\frac{(3\omega^{\ast}-2)(4\omega^{\ast}-3)}{\omega^{\ast 2}(2\omega^{\ast}-1)^{2}}.
\end{equation}
The smooth branch crosses $\eta_{0}=1$ at $g_{1}=c_{2}$, equivalently
\begin{equation}\label{eq:fwd-pp-gamma-wc}
\gamma_{\rm wc}(\omega^{\ast})=c_{2}^{2}=1/\omega^{\ast 2}.
\end{equation}

\textbf{Warm flat interval.} For $\eta<1$ one has $\lambda_{-}(\eta)>c_{2}$, and when $\lambda_{-}(\eta)\geq\sqrt{\gamma/\eta}$ the perturbed source $(\lambda_{+},\lambda_{-},c_{2})$ acquires one outlier above the bulk edge in addition to those of $(c_{1},c_{2})$. In this regime the $\eta$-dependent pieces of $-\Phi(\lambda_{+},\lambda_{-},c_{2})+\Phi(c_{1},c_{2})$ cancel exactly, producing a flat plateau in $D_{\mathrm{KL}}(\Pteacher\Vert\PredPost)$. Setting $\lambda_{-}(\eta)=\sqrt{\gamma/\eta}$ yields a quadratic in $\sqrt{\gamma/\eta}$,
\begin{equation}\label{eq:fwd-pp-warm-quadratic}
\frac{\gamma}{\eta}-\bigl[\gamma+c_{1}-\tfrac{1}{2}(c_{1}-c_{2})^{2}\bigr]\sqrt{\gamma/\eta}+\gamma\,c_{1}=0,
\end{equation}
whose two positive roots delimit the warm interval
\begin{equation}\label{eq:fwd-pp-iwarm}
I_{\rm warm}(\gamma,\omega^{\ast})=\bigl[\eta_{-},\;\min(1,\eta_{+})\bigr],
\end{equation}
with endpoints
\begin{equation}\label{eq:fwd-pp-eta-warm}
\eta_{\pm}=\frac{4\gamma}{\Bigl[\gamma+c_{1}-\tfrac{1}{2}(c_{1}-c_{2})^{2}\,\mp\,\sqrt{\bigl[\gamma+c_{1}-\tfrac{1}{2}(c_{1}-c_{2})^{2}\bigr]^{2}-4\gamma c_{1}}\Bigr]^{2}}.
\end{equation}
Every $\eta\in I_{\rm warm}$ is a global minimiser. The discriminant of~\eqref{eq:fwd-pp-warm-quadratic} vanishes at
\begin{equation}\label{eq:fwd-pp-gamma-flat}
\gamma_{\rm flat}(\omega^{\ast})=\bigl(\sqrt{c_{1}}-(c_{1}-c_{2})/\sqrt{2}\bigr)^{2},
\end{equation}
above which the warm flat regime no longer exists. For $\omega^{\ast}>1$ the three thresholds satisfy $\gamma_{\rm wc}<\gamma_{\rm flat}<\gamma_{\infty}$, with equality only in the degenerate limit $\omega^{\ast}\downarrow 1$.

\textbf{Uncondensed plateau.} For $\gamma>c_{1}$, both the original and perturbed sources are uncondensed for $\eta\geq\eta_{\mathrm{PM}}$. Setting $\lambda_{+}(\eta)=\gamma$ in~\eqref{eq:lambda-pm} gives the threshold
\begin{equation*}
\eta_{\mathrm{PM}}=\frac{\gamma-c_{1}+\tfrac{1}{2}(c_{1}-c_{2})^{2}}{\gamma(\gamma-c_{1})}.
\end{equation*}
Substituting the uncondensed branch of~\eqref{eq:Phi1-eval} into~\eqref{eq:forward-kl-pp} gives, throughout this interval,
\begin{equation*}
\frac{1}{N}D_{\mathrm{KL}}(\Pteacher\Vert\PredPost)=\frac{\ln\omega^{\ast}}{2}-\frac{(c_{1}-c_{2})^{2}}{4\gamma}.
\end{equation*}
The leading forward KL therefore attains its MAP value at every $\eta\in[\eta_{\mathrm{PM}},\infty)$, giving a degenerate optimum (\emph{deg.}\ in Fig.~\ref{fig:warm-cold-pp-fwd}A). Unlike the reverse-KL plateau, this interval does not include all positive $\eta$.

\textbf{Phase classification.} The minimisation has five outcomes (Fig.~\ref{fig:warm-cold-pp-fwd}A). For $\gamma<\gamma_{\rm wc}$ the warm flat interval $I_{\rm warm}\subset(0,1]$ is the optimum-set, with smooth-branch representative $\eta_{0}<1$ inside; at $\gamma=\gamma_{\rm wc}$, $\eta_{0}=1$. In the sliver $\gamma_{\rm wc}<\gamma\leq\gamma_{\rm flat}$ a warm flat sub-interval $I_{\rm warm}\subset(0,1)$ and the cold isolated point $\eta_{0}>1$ are simultaneously global minimisers; $D_{\mathrm{KL}}(\Pteacher\Vert\PredPost)/N$ traces a non-convex W shape with the two troughs at the same value to leading order in $N$. At $\gamma=\gamma_{\rm flat}$ the warm interval collapses to a single point, still distinct from the cold minimiser. For $\gamma_{\rm flat}<\gamma<\gamma_{\infty}$ the unique global minimiser is the cold smooth-branch point $\eta_{\mathrm{pp}}^{\mathrm{fwd}}=\eta_{0}>1$. For $\gamma_{\infty}\leq\gamma\leq c_{1}$ the infimum is reached only as $\eta\to\infty$ (MAP), whereas for $\gamma>c_{1}$ every $\eta\geq\eta_{\mathrm{PM}}$ is a global minimiser.

\subsection{Training dynamics of the KL divergences}\label{app:kl-divergences-dyn}

We derive asymptotically exact expressions tracking the large-$N$ limit of the intensive instantaneous KL divergences over the course of training. We focus on the typical KL divergences \begin{align}
  \frac{1}{N} D_{\mathrm{KL}}( \Pteacher \Vert \Prob_{t})&=\frac{1}{N}\mathcal{H}[\Pteacher;\Prob_t]-\frac{1}{N}\mathcal{H}[\Pteacher], \\
  \frac{1}{N} D_{\mathrm{KL}}(\Prob_{t} \Vert \Pteacher)&=\frac{1}{N}\mathcal{H}[\Prob_t;\Pteacher]-\frac{1}{N}\mathcal{H}[\Prob_t],
\end{align}where we used the shorthand $\Prob_t=\Prob_{W(t)}$ and $\Pteacher=\Prob_{\Wteacher}$, with $\Wteacher$ the rank-one teacher of App.~\ref{app:teacher-data}.

\subsubsection{Student average energy and entropy}
Let $\lambda_1(t)$ be the top eigenvalue of $W(t)$ and $\mu(t)=\max\{\Stieltjes^{-1}(1),\lambda_1(t)\}$. Then, similar to Eqs.~\eqref{eq:SBM-log-Z}, \eqref{eq:SBM-avg-energy} and \eqref{eq:SBM-entropy},
\begin{equation}\begin{aligned}
\frac{1}{N}\ln Z(W(t)) &= \frac{\ln(2\pi) + \mu(t) - \LogStieltjes(\mu(t))}{2}
\\
\frac{1}{N}\langle\mathcal{E}(\mathbf{x}; W(t))\rangle_{\mathbf x\sim\Prob_t} &= \frac{1-\mu(t)}{2},
\\
\frac{1}{N}\mathcal{H}[P_{W(t)}] &= \frac{\ln(2\pi e)-\LogStieltjes(\mu(t))}{2}
\end{aligned}\end{equation}

\subsubsection{Forward cross entropy}

We first compute
\begin{equation}
\frac{1}{N}\langle\mathcal{E}(\mathbf{x}; W(t))\rangle_{\mathbf x\sim\Pteacher} =
- \frac12\left(1-\frac{1}{\omega^{\ast}}\right)\mathbf{w}^{\ast}\!^\top \Theta(t)\mathbf{w}^{\ast}
\end{equation}
where
\begin{equation}
\Theta(t) = \frac{1-e^{-\gamma t/2}}{\gamma}C-\frac{1}{N}\int_0^t\mathrm{d}u\,e^{-\gamma (t-u)/2}\mathbf{x}(u)\mathbf{x}(u)^\top
\end{equation}
and where we used
$\frac{\langle\mathbf{x}^\top W(t)\mathbf{x}\rangle_{\Pteacher}}{N} = \frac{\langle(\mathbf{w}^{\ast}\cdot\mathbf{x})^2\rangle_{\Pteacher}}{N}\mathbf{w}^{\ast}\!^\top W(t)\mathbf{w}^{\ast}$ and $\frac{\langle(\mathbf{w}^{\ast}\cdot\mathbf{x})^2\rangle_{\Pteacher}}{N} = 1-1/\omega^{\ast}$, to leading order, since $\mathbf{w}^{\ast}\!^\top W_\mathrm{GOE}(t)\mathbf{w}^{\ast}=o(1)$. Here $\Theta(t)$ has at most $K$ positive eigenvalues $\theta_k(t)$, with eigenvectors $\mathbf{t}_k$. Decomposing $\mathbf{w}^{\ast}$ onto $\mathbf{c}_1,\dots,\mathbf{c}_K$, one gets\begin{align}
  \mathbf{w}^{\ast}\!^\top \Theta(t)\mathbf{w}^{\ast}=\left(1-\frac{1}{2\omega^{\ast}-1}\right)\left(\frac{1-e^{-\gamma t/2}}{\gamma}c_1-\int_0^t\mathrm{d}u\,e^{-\gamma (t-u)/2}s_1(u)^2\right),
\end{align}
where we used Eq. \eqref{eq:teacher-eigenvectors}.

We then find:
\begin{align}\label{eq:dyn-fwd-ce}
\frac1N\mathcal{H}[\Pteacher;\Prob_t] 
=
\frac{\ln2\pi +\mu(t) - \LogStieltjes(\mu(t))}{2}
- \frac12\left(1-\frac{1}{\omega^{\ast}}\right)\mathbf{w}^{\ast}\!^\top \Theta(t)\mathbf{w}^{\ast}.
\end{align}

Determining $\mu(t)$ or $g_t:=\Stieltjes(\mu(t))$ requires knowledge of $\lambda_1(t)$. To make progress, we use the approximation that $W_{\mathrm{GOE}}(t)$ is asymptotically independent of $\Theta(t)$. Note that, for $u\le t$, $\Theta(t)$ depends on $W_{\mathrm{GOE}}(u)$ through $\mathbf{x}(u)$. Therefore, this approximation is valid for $t$ small enough that the negative phase is small relative to the positive phase. 

Properties of finite-rank deformed Wigner matrices then imply
\begin{align}
    g_t=\begin{cases}
        \min\{1,\sqrt{\gamma\eta}\}& \theta_1(t)\le \theta_c,\\
        \theta_1(t)^{-1}, &\theta_1(t)>\theta_c,
    \end{cases} \qquad\text{with}\qquad 
    \theta_c=\max\{1,1/\sqrt{\gamma\eta}\},
\end{align}
$h_t^2=(1-g_t)_+$ and $\mu(t)=g_t/(\gamma\eta)+1/g_t$.

\subsubsection{Reverse cross entropy}
Injecting $\Wteacher$ gives $\frac1N\langle\mathcal{E}(\mathbf{x};\Wteacher)\rangle_{\mathbf{x}\sim \Prob_t}=-\frac{\omega^{\ast}h_t^2}{2}\frac{\langle(\mathbf{v}_1\cdot\mathbf{w}^{\ast})^2\rangle}{N}+o(1)$ upon using $\frac{\omega^{\ast}}{2}\frac{\Vert \mathbf{x}\Vert_2^2}{N}=\mathcal{O}(1)$ and $\frac{\langle(\mathbf{w}^{\ast}\cdot\mathbf{x})^2\rangle}{N}=h_t^2\frac{\langle(\mathbf{v}_1\cdot\mathbf{w}^{\ast})^2\rangle}{N}+o(1)$. Therefore,
\begin{align}\label{eq:dyn-rev-ce}
    \frac1N\mathcal{H}[\Prob_t;\Pteacher]=\frac12\ln2\pi+\frac{1}{2}(\omega^{\ast}-\ln\omega^{\ast})-\frac{1}{2}\omega^{\ast}h_t^2\frac{(\mathbf{v}_1(t)\cdot\mathbf{w}^{\ast})^2}{N}+o(1).
\end{align}

We again make the early-training approximation that $\Theta(t)$ is asymptotically independent of $W_{\mathrm{GOE}}(t)$. Once $\lambda_1(t)$ has detached one has, from spectral properties of finite-rank deformed Wigner matrices, \begin{align}
    \frac{(\mathbf{v}_1(t)\cdot\mathbf{t}_1(t))^2}{N^2}=\left(1-\frac{1}{\gamma\,\eta\,\theta_1(t)^2}\right)_++o(1).
\end{align}
The component of $\mathbf{v}_1(t)$ orthogonal to $\mathbf{t}_1(t)$ is asymptotically isotropic in the $(N-1)$-dimensional subspace orthogonal to $\mathbf{t}_1(t)$ and is therefore asymptotically orthogonal to $\mathbf{w}^{\ast}$. As a result, 
\begin{align}
    \frac{(\mathbf{v}_1(t)\cdot\mathbf{w}^{\ast})^2}{N}=\left(1-\frac{1}{\gamma\,\eta\,\theta_1(t)^2}\right)_+\frac{(\mathbf{t}_1(t)\cdot\mathbf{w}^{\ast})^2}{N}+o(1).
\end{align}
To evaluate $\mathbf{t}_1(t)\cdot\mathbf{w}^{\ast}$, expand $\mathbf{t}_1(t)$ in the eigenbasis of the empirical covariance of the training data and use $(\mathbf{c}_1\cdot \mathbf{w}^{\ast})^2/N=1-1/(2\omega^{\ast}-1)$, $(\mathbf{c}_k\cdot\mathbf{w}^{\ast})^2/N=o(1)$ ($k>d$). In this way, we find
\begin{align}\label{eq:dyn-teachr-student-overlap}
    \frac{(\mathbf{v}_1(t)\cdot\mathbf{w}^{\ast})^2}{N}=\left(1-\frac{1}{\gamma\eta\,\theta_1(t)^2}\right)\left(1-\frac{1}{2\omega^{\ast}-1}\right)\frac{\left(\mathbf{t}_1(t)\cdot\mathbf{c}_1\right)^2}{N^2}.
\end{align}

\subsubsection{Forward KL}
Combining Eqs.~\eqref{eq:lnZ-and-entropy-teacher-sbm-rank-1} and \eqref{eq:dyn-fwd-ce} yields
\begin{align}
&\frac{1}{N} D_{\mathrm{KL}}( \Pteacher \, \Vert\, \Prob_{t})
= \frac12\ln\omega^{\ast} - \frac{1}{2}+\frac12\left[\mu(t)+\ln g_t-\frac{g_t^2}{2\gamma\eta}\right]\nonumber\\&
\!-\frac12\left(1-\frac{1}{\omega^{\ast}}\right)\left(1-\frac{1}{2\omega^{\ast}-1}\right)\left(\frac{1-e^{-\gamma t/2}}{\gamma}c_1-\int_0^t\mathrm{d}u\,e^{-\gamma (t-u)/2}s_1(u)^2\right).
\end{align}

\subsubsection{Reverse KL}
Using Eqs.~\eqref{eq:dyn-rev-ce}, \eqref{eq:dyn-teachr-student-overlap}, \begin{align}
&\frac{1}{N} D_{\mathrm{KL}}(\Prob_{t} \, \Vert\, \Pteacher)=\frac{1}{2}(\omega^{\ast}-1-\ln\omega^{\ast})-\frac12\ln g_t+\frac{g_t^2}{4\gamma\eta}\nonumber\\
&-\frac{1}{2}\omega^{\ast}h_t^2\left(1-\frac{1}{\gamma\eta\,\theta_1(t)^2}\right)\left(1-\frac{1}{2\omega^{\ast}-1}\right)\frac{\left(\mathbf{t}_1(t)\cdot\mathbf{c}_1\right)^2}{N^2}.
\end{align}
At very early times, $t<t_p=\inf\{t:\theta_1(t)>\max\{1,1/\sqrt{\gamma\eta}\}\}$, until the student condenses along an outlier direction, the reverse KL stays flat, with plateau height \begin{align}
    \frac{1}{N} D_{\mathrm{KL}}(\Prob_{t} \, \Vert\, \Pteacher)=\frac{1}{2}(\omega^{\ast}-1-\ln\omega^{\ast})-\frac{1}{2}\ln g_0+\frac{g_0^2}{4\gamma\eta}, \qquad && g_0=\begin{cases}
        1,  & \gamma\eta\ge 1\\
        \sqrt{\gamma\eta}, &\gamma\eta<1.
    \end{cases}
\end{align}

\paragraph{Optimal stopping time.} For $t>t_p$, the saddle is controlled by the data-aligned spike, $g_t=\theta_1(t)^{-1}$, and the behavior of the reverse KL results from the trade-off between  the (i) marginal gain of increasing the student's alignment with the teacher and (ii) the marginal cost of decreasing the student's diversity. The optimal stopping time $t^*$ is reached when (i) and (ii) are balanced,\begin{align}
    &\frac{\mathrm{d}}{\mathrm{d}t}\Big(\omega^{\ast}h_t^2\Big(1-\tfrac{1}{\gamma\eta\,\theta_1(t)^2}\Big)\Big(1-\tfrac{1}{2\omega^{\ast}-1}\Big)\tfrac{\left(\mathbf{t}_1(t)\cdot\mathbf{c}_1\right)^2}{N^2}\Big)\Big\vert_{t=t^*}=\frac{\mathrm{d}}{\mathrm{d}t}\left(\ln \theta_1(t)+\tfrac{1}{2\gamma\eta\theta_1(t)^2}\right)\Big\vert_{t=t^*}.\nonumber
\end{align} 
When $u_1^2(t^*)\simeq 1$, setting $\partial_\theta D_{\mathrm{KL}}(\Prob_{t} \, \Vert\, \Pteacher)\vert_{\theta=\theta^*}=0$ gives a quadratic in $\theta^*$, whose positive root is \begin{align}
\theta^\star=\frac{1}{2}\left[\omega^{\ast}\left(1-\frac{1}{2\omega^{\ast}-1}\right)+\sqrt{\omega^{\ast 2}\left(1-\frac{1}{2\omega^{\ast}-1}\right)^2+\frac{4}{\gamma\eta}}\right].
\end{align}
The early-training approximation, $\theta_1(t)\simeq \frac{c_1}{\gamma}(1-e^{-\gamma t/2})$, leads to the $\nu$-independent estimate
\begin{align}\label{eq:early-stopping-time}
    t^*\simeq-\frac{2}{\gamma}\ln\left(1-\frac{\gamma\,\theta^*}{c_1}\right).
\end{align}Injecting the parameter values of Fig. \ref{fig:ooe-wrap} yields $t^*\simeq 5\ln 2\simeq 3.47$, in excellent agreement with the numerical value.

\clearpage
\subsection{Posterior-predictive forward KL}\label{app:cross-entropies}

\clearpage
\section{Comparison of the SBM with a multivariate Gaussian model}\label{app:comparison-gaussian}

The SBM and a multivariate Gaussian both have quadratic energies, but the spherical constraint permits sample condensation, in which a finite number of modes carries a nonzero fraction of the total squared norm. Spectral outlier transitions also occur in standard spiked Gaussian covariance models~\citep{bbp2005}, where fixed spike strengths do not produce sample condensation. Spectral separation and sample condensation are therefore distinct phenomena.

\begin{figure}[h]
\centering
\includegraphics[width=0.65\linewidth]{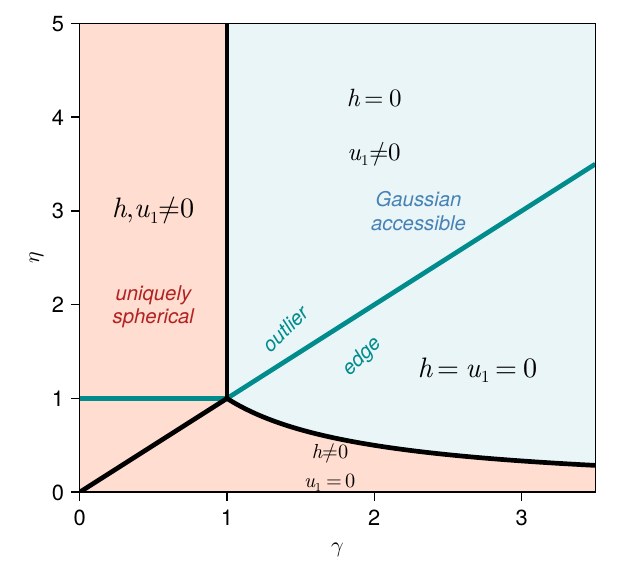}
  \caption{\textbf{Spectral outlier transitions and spherical condensation.}
  Phase diagram of the $K=1$ SBM, with phases labeled by $(h,u_1)$.
  Blue shading identifies uncondensed sectors whose spectral behavior has analogues in spiked Gaussian models.
  Salmon shading identifies sectors with sample condensation, absent from the Gaussian comparison with fixed spike strengths.
  \label{fig:sbm-vs-gaussian}}
\end{figure}

Figure~\ref{fig:sbm-vs-gaussian} illustrates this distinction using the $K=1$ SBM phase diagram. The uncondensed sectors ($h=0$), with or without a spectral outlier, have Gaussian analogues. The condensed sectors ($h\ne0$) exhibit the additional condensation mechanism induced by the spherical constraint.

\clearpage

\section{First-order SBM reduction of a general spherical EBM}
\label{app:first-order-sbm-reduction}

We describe a simple sense in which the SBM arises as a leading quadratic approximation of a more general EBM on the sphere. Let $\mu_{0}$ denote the uniform probability measure on
\begin{equation}
\mathcal{S}_{N}
= \left\{
\mathbf{x}\in\mathbb{R}^{N}:
\|\mathbf{x}\|^{2}=N
\right\},
\end{equation}
and consider an arbitrary spherical EBM
\begin{equation}
\frac{\mathrm{d}\Prob_{\epsilon}}{\mathrm{d}\mu_{0}}(\mathbf{x})
=
\frac{\exp[-\epsilon\Energy(\mathbf{x})]}
{\left\langle \exp[-\epsilon\Energy]\right\rangle_{0}},
\label{eq:general-spherical-ebm}
\end{equation}
where $\langle\cdot\rangle_{0}$ denotes averaging under $\mu_{0}$ and $\Energy(\mathbf x)$ is an arbitrary function, for example parameterized by a neural network.
The parameter $\epsilon$ is a bookkeeping parameter controlling the departure from the uniform distribution. Here $\mathbf x\in\mathcal{S}_N$ may represent either original data on the sphere, or a normalized
feature representation of a more general EBM.

Writing
\begin{equation}
\widetilde{\Energy}(\mathbf{x})
=
\Energy(\mathbf{x})-\langle\Energy\rangle_{0},
\end{equation}
the normalized density has the expansion
\begin{equation}
\frac{\mathrm{d}\Prob_{\epsilon}}{\mathrm{d}\mu_{0}}(\mathbf{x})
=
1-\epsilon\widetilde{\Energy}(\mathbf{x})
+\BigO(\epsilon^{2}).
\label{eq:general-spherical-ebm-expansion}
\end{equation}
Equivalently,
\begin{equation}
\log\frac{\mathrm{d}P_{\epsilon}}{\mathrm{d}\mu_0}(\mathbf{x})
=
-\epsilon\widetilde{\mathcal E}(\mathbf{x})
+\mathcal O(\epsilon^2).
\end{equation}
At order $\epsilon$, the response of any centered quadratic observable
depends only on the degree-two spherical-harmonic component of this
log-density.

The degree-two spherical harmonics are precisely the traceless quadratic
forms
\[
\mathbf{x}\mapsto \frac{1}{2}\mathbf{x}^{\top}W\mathbf{x},
\qquad
W\in\mathrm{Sym}_{N},
\qquad
\Trace W=0.
\]
The degree-two component is uniquely represented as
$\frac12\mathbf{x}^{\top}W_{\mathrm{eff}}\mathbf{x}$, where
$\operatorname{tr}W_{\mathrm{eff}}=0$. Equivalently,
\begin{equation}
W_{\mathrm{eff}}
=
\underset{\substack{W\in\mathrm{Sym}_{N}\\ \Trace W=0}}
{\arg\min}\,
\left\langle
\left[
-\epsilon\widetilde{\Energy}(\mathbf{x})
-\frac{1}{2}\mathbf{x}^{\top}W\mathbf{x}
\right]^{2}
\right\rangle_{0}.
\label{eq:Weff-projection-definition}
\end{equation}
This projection can be evaluated explicitly. The uniform spherical
moments are
\begin{equation}
\left\langle x_{i}x_{j}\right\rangle_{0}
=
\delta_{ij},
\qquad
\left\langle x_{i}x_{j}x_{k}x_{l}\right\rangle_{0}
=
\frac{N}{N+2}
\left(
\delta_{ij}\delta_{kl}
+\delta_{ik}\delta_{jl}
+\delta_{il}\delta_{jk}
\right).
\label{eq:uniform-sphere-moments}
\end{equation}
Consequently, for any traceless $U\in\mathrm{Sym}_{N}$,
\begin{equation}
\left\langle
\left(\mathbf{x}\mathbf{x}^{\top}-I\right)
\frac{1}{2}\mathbf{x}^{\top}U\mathbf{x}
\right\rangle_{0}
=
\frac{N}{N+2}U.
\label{eq:quadratic-sphere-identity}
\end{equation}
The normal equations associated with
Eq.~\eqref{eq:Weff-projection-definition} therefore give
\begin{equation}
W_{\mathrm{eff}}
=
-\epsilon\frac{N+2}{N}
\left\langle
\left(\mathbf{x}\mathbf{x}^{\top}-I\right)
\Energy(\mathbf{x})
\right\rangle_{0}.
\label{eq:Weff-explicit}
\end{equation}
Notice that $\Trace W_{\mathrm{eff}}=0$ automatically, since
$\|\mathbf{x}\|^{2}=N$. Equivalently, for every traceless
$U\in\mathrm{Sym}_{N}$,
\begin{equation}
\left\langle
\left[
-\epsilon\widetilde{\Energy}(\mathbf{x})
-\frac{1}{2}\mathbf{x}^{\top}W_{\mathrm{eff}}\mathbf{x}
\right]
\mathbf{x}^{\top}U\mathbf{x}
\right\rangle_{0}
=
0.
\label{eq:quadratic-projection-orthogonality}
\end{equation}
Thus $\frac{1}{2}\mathbf{x}^{\top}W_{\mathrm{eff}}\mathbf{x}$ is exactly
the degree-two component of the first-order log-density. If the EBM is
invariant under $\mathbf{x}\mapsto-\mathbf{x}$, as in the zero-field
setting studied here, its degree-one spherical harmonic vanishes.
Otherwise, the degree-one component remains orthogonal to all the
quadratic observables considered below.

The distribution obtained by retaining only this quadratic component is
\begin{equation}
\Prob_{\epsilon}^{(2)}(\mathbf{x})
=
\frac{1}{Z(W_{\mathrm{eff}})}
\exp\left(
\frac{1}{2}\mathbf{x}^{\top}W_{\mathrm{eff}}\mathbf{x}
\right),\qquad
\mathbf{x}\in\mathcal{S}_{N}
\label{eq:projected-sbm}
\end{equation}
which is precisely an SBM. 
The general EBM and its projected SBM have identical quadratic observables to first order.
In particular, defining the sampling covariance
\[
\Sigma_{\epsilon}
=
\left\langle
\mathbf{x}\mathbf{x}^{\top}
\right\rangle_{\Prob_{\epsilon}},
\]
Eqs.~\eqref{eq:general-spherical-ebm-expansion},
\eqref{eq:Weff-explicit}, and
\eqref{eq:quadratic-sphere-identity} give
\begin{equation}
\Sigma_{\epsilon}
=
I+\frac{N}{N+2}W_{\mathrm{eff}}
+\BigO(\epsilon^{2})
=
\left\langle
\mathbf{x}\mathbf{x}^{\top}
\right\rangle_{\Prob_{\epsilon}^{(2)}}
+\BigO(\epsilon^{2}).
\label{eq:covariance-first-order-reduction}
\end{equation}
Therefore, to first order around the isotropic distribution,
$W_{\mathrm{eff}}$ and $\Sigma_{\epsilon}-I$ have the same eigenvectors and proportional eigenvalues. Spectral alignment and the corresponding quadratic sample-overlap observables of a general spherical EBM thus have direct counterparts in the SBM.

This reduction is of course a first-order approximation only. It does not imply that the complete distribution is an SBM: higher spherical harmonics can affect higher-order observables already at order $\epsilon$. A distribution-level approximation additionally requires the non-quadratic components of the log-density to be of order $\epsilon^{2}$, and the perturbative argument by itself does not establish the non-perturbative condensation transition occurring at finite coupling.

We acknowledge an anonymous reviewer for suggesting this derivation.

\clearpage

\section{Computational experiments in generative models}\label{app:comp-exp}

We test the four SBM phenomena in tractable small models and larger generative architectures. Table~\ref{tab:experiment-summary} summarizes the observables and their correspondence with the SBM predictions.

\begin{table}[!ht]
\caption{Summary of the experiments and their correspondence with the SBM predictions.}
\label{tab:experiment-summary}
\centering
\footnotesize
\setlength{\tabcolsep}{4pt}
\renewcommand{\arraystretch}{1.1}
\begin{tabular}{@{}p{\dimexpr0.24\linewidth-6pt\relax}p{\dimexpr0.20\linewidth-6pt\relax}p{\dimexpr0.28\linewidth-6pt\relax}p{\dimexpr0.28\linewidth-6pt\relax}@{}}
\toprule
\raggedright\textbf{System / data} & \raggedright\textbf{Observables} & \raggedright\textbf{Phenomenology} & \raggedright\textbf{Comparison with SBM} \tabularnewline
\midrule
\raggedright Small BM on Curie--Weiss data (App.~\ref{app:experiments-small-bm}) &
\raggedright All state observables (exact enumeration) &
\raggedright Sampling-temperature tuning; double descent; warm and cold posteriors; out-of-equilibrium training &
\raggedright Qualitative agreement for all four effects \tabularnewline
\addlinespace
\raggedright Small RBM on two-pattern teacher data (App.~\ref{app:experiments-small-rbm}) &
\raggedright All state observables (exact enumeration) &
\raggedright Sampling-temperature tuning; double descent; warm and cold posteriors; out-of-equilibrium training &
\raggedright Qualitative agreement for all four effects \tabularnewline
\addlinespace
\raggedright Potts BM on Pfam and lattice-protein data (Apps.~\ref{app:temperature-tuning}, \ref{app:ooe-potts}) &
\raggedright Coupling and covariance matrices &
\raggedright Sampling-temperature tuning; out-of-equilibrium training &
\raggedright Qualitative spectral signatures \tabularnewline
\addlinespace
\raggedright FVSBN on Curie--Weiss and Ising data (App.~\ref{app:fvsbn-K-dd}) &
\raggedright Reverse KL; sample moments &
\raggedright Double descent &
\raggedright Qualitative agreement for double descent \tabularnewline
\addlinespace
\raggedright RBM on financial returns (App.~\ref{app:rbm-finance}) &
\raggedright Estimated NLL; weight matrix &
\raggedright Double descent &
\raggedright Qualitative spectral correspondence \tabularnewline
\addlinespace
\raggedright BayesGAN on a 2D Gaussian target (App.~\ref{app:tempered-posterior-gen}) &
\raggedright Estimated KL divergences &
\raggedright Warm and cold posteriors &
\raggedright Qualitative agreement with posterior-temperature dependence \tabularnewline
\bottomrule
\end{tabular}
\end{table}
\clearpage

\subsection{Experiments in small tractable BMs\label{app:experiments-small-bm}}
We numerically study fully connected zero-field binary Boltzmann machines in a controlled teacher--student setting. Configurations are vertices of the $N$-cube,
$s \in \{-1,+1\}^{N},$
and the teacher is a rank-one Curie--Weiss model,
\begin{equation}
P^\star(s)
=
\frac{1}{Z^\star}
\exp\!\left[
\frac{\omega^\star}{2N}
\left(\sum_{i=1}^{N} s_i\right)^2
\right].
\end{equation}
The parameter $\omega^\star$ controls the strength of the teacher interaction. Equivalently, up to an additive constant (independent of $s$), the teacher coupling matrix has entries $J^\star_{ij}=\frac{\omega^\star}{N}$ ($i\neq j$) and $J^\star_{ii}=0$, and $Z^\star(=Z(J^\star))$. Its finite-$N$ leading eigenvalue is therefore $\omega^\star(1-1/N)$.

The student is a fully connected binary Boltzmann machine,
\begin{equation}
P_J(s)
=
\frac{1}{Z(J)}
\exp\!\left[
\sum_{i<j} J_{ij}s_is_j
\right].
\end{equation}
The coupling matrix $J$ is symmetric, with $J_{ii}=0$, and its $N(N-1)/2$ off-diagonal entries are the free parameters. There are no external fields, and the couplings are subject to quadratic regularisation of intensity $\gamma$, as in Eq.~\eqref{eq:tilted-posterior}.

Training datasets consist of $K$ independent configurations sampled from the
teacher. %

Because the systems are small, all $2^N$ configurations are enumerated. For any fixed coupling matrix $J$, partition functions, equilibrium negative phases, state probabilities, forward and reverse Kullback--Leibler divergences, covariance matrices, and teacher-aligned observables are then evaluated exactly in configuration space. Integration over couplings remains numerical. For posterior tempering, we use four independent Metropolis-adjusted Hamiltonian Monte Carlo chains per dataset and parameter pair, retaining $4000$ draws per chain after $1000$ warmup steps, extended to $16000$ draws at $\gamma=0.1$ and $\eta\in\{0.05,0.075,0.1,0.15\}$. Across the tested grid, the maximum rank-normalised split-$\widehat R$ is below $1.005$ and the minimum bulk effective sample size exceeds $2300$. For the $K=1$ regularisation sweep, we average exactly over teacher-generated datasets and spin-sign gauges, using independent Gaussian-prior gauge-orbit draws at $\eta=1$ and importance sampling at the other temperatures, supplemented by adjusted Hamiltonian Monte Carlo and Laplace importance proposals in the cold regime. State enumeration is exact to numerical precision; posterior averages retain Monte Carlo uncertainty.

The common performance measures are
\begin{equation}
D_{\mathrm{KL}}(P^\star \Vert Q)
\qquad\text{and}\qquad
D_{\mathrm{KL}}(Q \Vert P^\star),
\end{equation}
where $Q$ is either an individual trained model or a posterior-predictive distribution. For posterior prediction, we average $P_J$ over the coupling posterior before evaluating the KL. The typical reverse KL in the regularisation sweep instead averages $D_{\mathrm{KL}}(P_J\Vert P^\star)$ over posterior draws and teacher-generated datasets.

\begin{figure}
    \centering
\includegraphics[width=1.0\linewidth]{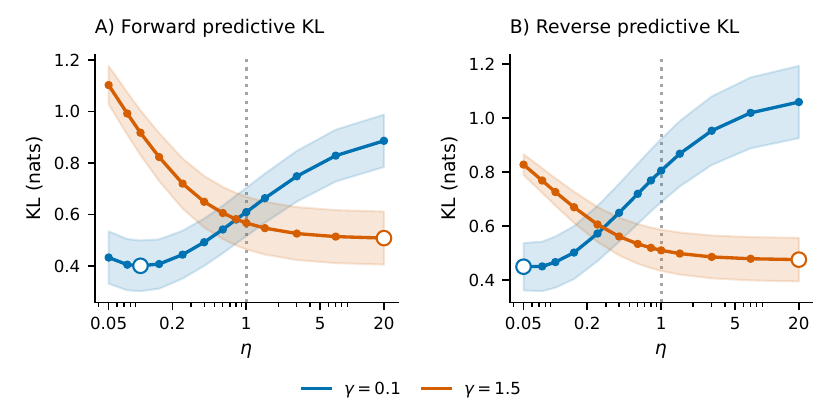}
\caption{
\textbf{Posterior tempering in a zero-field binary BM student with a Curie--Weiss teacher.} The forward \textbf{A)} and reverse \textbf{B)} KL losses of the posterior predictive are shown versus posterior inverse temperature $\eta$, for weak ($\gamma=0.1$, blue) and strong ($\gamma=1.5$, orange) regularisation. Curves are means over $100$ independently sampled datasets; shading gives pointwise $99\%$ confidence intervals over datasets. The vertical dotted line marks the ordinary Bayes posterior, $\eta=1$, and open circles mark minima on the tested grid. Under weak regularisation, the forward and reverse minima occur at $\eta=0.1$ and $0.05$, respectively; under strong regularisation, both occur at the largest tested value, $\eta=20$. Shared parameters: $N=8$, $K=4$, $\omega^\star=1.5$; posterior convention as in Eq.~\eqref{eq:tilted-posterior}.
    }
    \label{fig:binar-bm-posterior-tempering}   
\end{figure}

\begin{figure}
    \centering
    \includegraphics[width=1.0\linewidth]{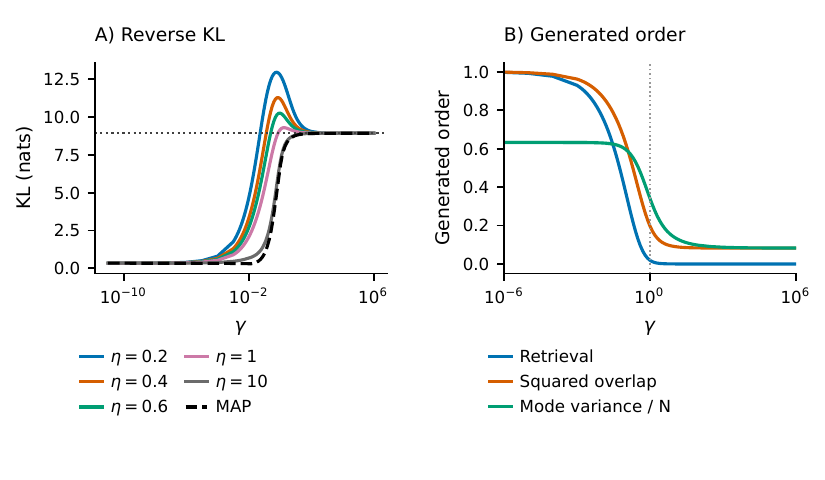}
    \caption{
    \textbf{Double descent with regularization strength in a zero-field binary BM student with a Curie--Weiss teacher.}
    \textbf{A)} Posterior-average reverse KL versus regularisation $\gamma$ for various inverse temperatures $\eta$. For $\eta=0.2,0.4,0.6,1$, the loss rises from its weak regularisation minimum, overshoots the uniform student limit $8.9265$ nats (horizontal dotted line), and then decreases towards that limit. Cooling reduces the overshoot, which is absent at $\eta=10$. The dashed black curve gives the exact MAP loss, obtained from the strictly convex objective. It has a single minimum, $0.2846$ nats at $\gamma\simeq0.006604$, and then increases towards the uniform limit from below, without a second descent.
    \textbf{B)} Retrieval probability $P_J(\pm\xi)$, squared overlap with the observed sample $\langle(s\cdot\xi/N)^2\rangle$, and variance along the leading coupling eigenvector divided by $N$, at $\eta=1$. Their decline accompanies the loss of the data-aligned mode; the vertical dotted line marks $\gamma=1$ as a reference in panel \textbf{B)}. Averages over all possible one-example datasets are exact. Shading gives approximate $95\%$ Monte Carlo intervals for the coupling integrals. Common parameters: $N=12$, $K=1$, $\omega^\star=3$, with Eq.~\eqref{eq:tilted-posterior} defining $\gamma$ and $\eta$.
    }
    \label{fig:small-bm-double-descent}
\end{figure}

\begin{figure}
\centering    
\includegraphics[width=1.0\linewidth]{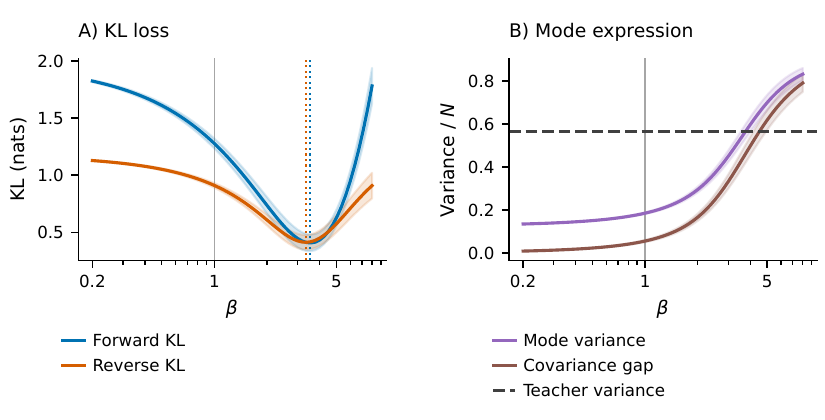}
\caption{\textbf{Generative temperature tuning in a zero-field binary BM student with a Curie--Weiss teacher.}
We hold the exact MAP couplings fixed while varying the generation inverse temperature $\beta$. Curves average $200$ independently sampled datasets; shading gives pointwise $99\%$ confidence intervals. \textbf{A)} Forward (blue) and reverse (orange) KL losses have cold optima at $\beta_{\mathrm{opt}}^{\mathrm{fwd}}\simeq3.54$ and $\beta_{\mathrm{opt}}^{\mathrm{rev}}\simeq3.34$, marked by the corresponding dotted lines. \textbf{B)} Generated variance along $v_1(\widehat J)$ (purple), generated covariance eigenvalue gap (brown), and teacher variance along $v_1(\widehat J)$ (horizontal dashed), all divided by $N$. The learned direction is underexpressed at $\beta=1$; cooling restores its variance towards the teacher value and improves generation. Grey vertical lines mark $\beta=1$. Common parameters: $N=8$, $K=4$, $\omega^\star=1.5$, $\gamma=6$ in Eq.~\eqref{eq:tilted-posterior}.
}
\label{fig:tt}
\end{figure}

\begin{figure}
    \centering
    \includegraphics[width=1.0\linewidth]{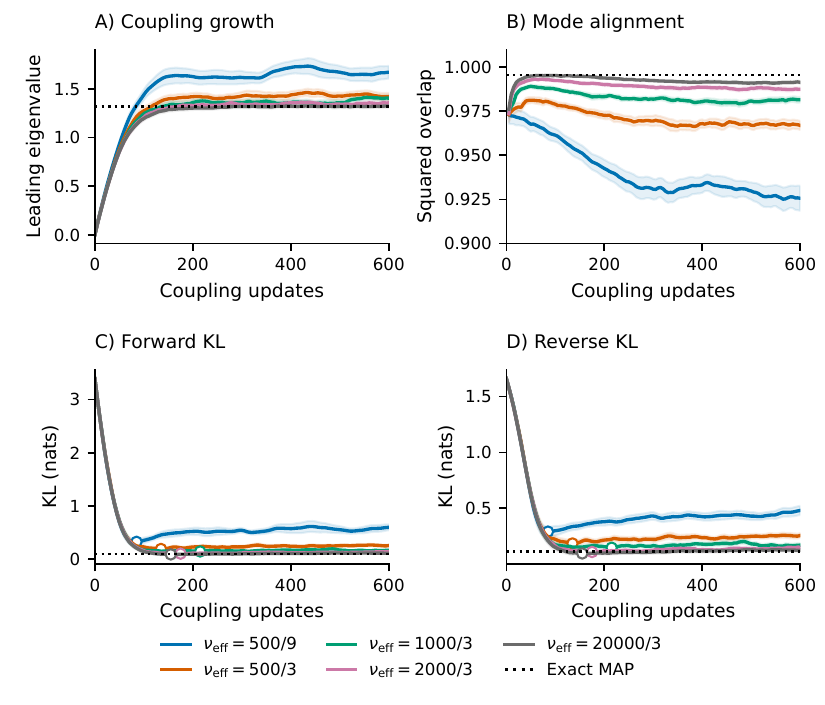}
    \caption{
    \textbf{Out-of-equilibrium learning in a zero-field binary BM student with a Curie--Weiss teacher.}
    Persistent contrastive-divergence training uses effective sampling rates $\nu_{\mathrm{eff}}\in\{500/9,500/3,1000/3,2000/3,20000/3\}$, with one persistent particle per replica. We define $\nu_{\mathrm{eff}}=m/(N\Delta t)$ for $m=1,3,6,12,120$ random single-spin Gibbs updates per coupling update. In the time convention of Eq.~\eqref{eq:W-langevin}, the learning rate $\epsilon=0.003$ gives $\Delta t=2N\epsilon/K=0.0015$. Thus $\nu_{\mathrm{eff}}$ is the Gibbs refresh rate per spin per unit learning time. Couplings start at zero and particles start uniformly; random-scan heat-bath updates precede each coupling update. Curves average $50$ independent datasets after averaging three training replicas per dataset; shading gives pointwise $95\%$ dataset confidence intervals. Horizontal dotted lines denote exact MAP values. \textbf{A)} Slow sampling produces an overshoot of the leading coupling eigenvalue, reaching $1.670$ at step $600$, compared with $1.322$ for the fastest sampler and $1.319$ at MAP. \textbf{B)} Squared overlap of the leading learned eigenvector with the principal empirical direction, showing reduced late-time alignment under slower sampling. \textbf{C,D)} Exact forward and reverse KL losses; open circles mark minima on the five-update checkpoint grid. Faster sampling reduces late-time degradation, while the precise optimum times need not be ordered by sampling rate. Common parameters: $N=12$, $K=48$, $\omega^\star=1.5$, and $\gamma=3.2$ in Eq.~\eqref{eq:tilted-posterior}. The learning rate $0.003$ multiplies the gradient of the negative log-posterior per training example.
    }    \label{fig:ooe}
\end{figure}

\clearpage

\subsection{Experiments in small tractable RBMs\label{app:experiments-small-rbm}}

We use a zero-field binary RBM with visible spins $\mathbf v\in\{\pm1\}^{N}$, hidden spins $\mathbf h\in\{\pm1\}^{M}$, and joint distribution
\begin{equation}
\Prob_\beta(\mathbf v,\mathbf h\mid W)
=\frac{\exp(\beta\,\mathbf v^\top W\mathbf h)}{Z_\beta(W)},
\qquad
\Prob_\beta(\mathbf v\mid W)
=\sum_{\mathbf h}\Prob_\beta(\mathbf v,\mathbf h\mid W),
\label{eq:small-rbm-model}
\end{equation}
where $W\in\mathbb R^{N\times M}$, $Z_\beta(W)$ is the partition function, and $\beta$ is the generation inverse temperature. The student has $N=16$ visible and $M=3$ hidden spins. Each training set contains $K=16$ independent visible configurations sampled from a teacher with two hidden spins and weights $W^\star=(\boldsymbol\xi_1,\boldsymbol\xi_2/2)$, where the $\boldsymbol\xi_r$ are orthogonal $\pm1$ patterns. Both teacher sampling and training use $\beta=1$. For training data $\mathcal D$, the posterior is
\begin{equation}
\Prior_{\eta,\gamma}(W\mid\mathcal D)
\propto\exp\!\left\{\eta\left[
\sum_{\mathbf v\in\mathcal D}\log\Prob_1(\mathbf v\mid W)
-\frac{\gamma}{2}\|W\|_F^2
\right]\right\}.
\label{eq:small-rbm-posterior}
\end{equation}
Here $\gamma$ sets the weight penalty and $\eta$ is the posterior inverse temperature. We sum over spin configurations exactly, use exact-gradient optimization for MAP fits, and use replica-exchange HMC for averages over weights.

\begin{figure}[!htbp]
\centering
\includegraphics[width=\linewidth]{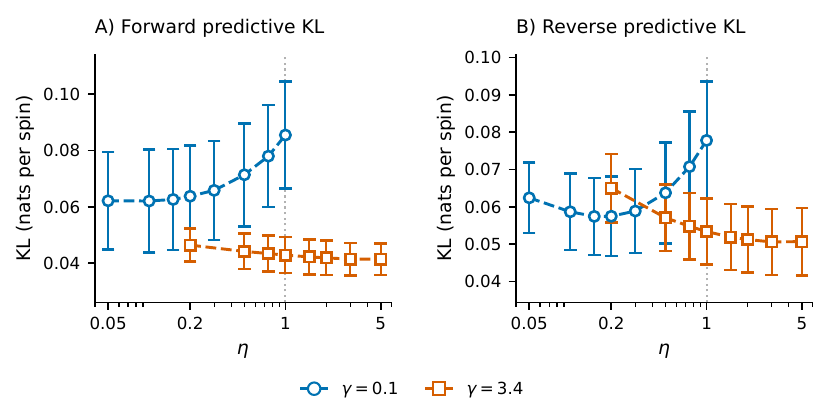}
\caption{\textbf{Posterior tempering in the small RBM.}
\textbf{A,B)} Forward and reverse KL of the posterior-predictive mixture, in nats per visible spin, at weak ($\gamma=.1$, blue) and stronger ($\gamma=3.4$, orange) regularization. Each curve contains eight tested temperatures, spanning $.05\le\eta\le1$ and $.2\le\eta\le5$, respectively. Points average over eight independent training sets from the teacher defined in the text; bars give $95\%$ dataset confidence intervals, and dashed segments guide the eye. The vertical dotted line marks the ordinary posterior, $\eta=1$. Warming ($\eta<1$) improves prediction under weak regularization, while cooling ($\eta>1$) improves it under stronger regularization. Some posterior-predictive estimates retain unresolved Monte Carlo precision limitations; this uncertainty is not included in the dataset intervals. Common parameters: $N=16$, $M=3$, $K=16$.}
\label{fig:small-rbm-eta}
\end{figure}

\begin{figure}[!htbp]
\centering
\includegraphics[width=\linewidth]{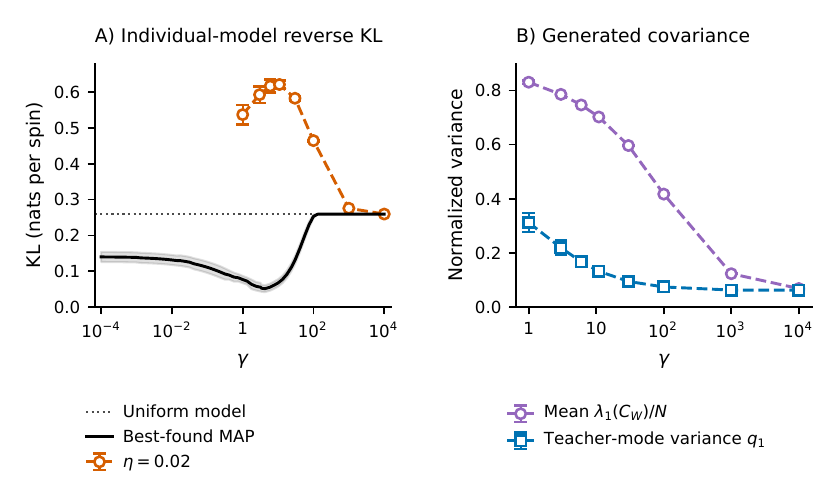}
\caption{\textbf{Double descent and generated covariance in the small RBM.}
Posterior averages use eight regularization strengths from $1$ to $10^4$ at $\eta=.02$; the MAP comparison uses $70$ values from $10^{-4}$ to $10^4$. Both average over the same eight independently drawn training sets, each containing $16$ visible configurations from the teacher defined at the start of App.~\ref{app:experiments-small-rbm}.
\textbf{A)} Mean individual-model reverse KL per spin. Orange markers show posterior averages; the solid black curve shows best-found MAP fits selected by the training objective from multiple exact-gradient initializations and bidirectional continuation. Their sampled mean minimum is $0.05134$ nats per spin at $\gamma\simeq4.217$; the subsequent high-regularization peak and second descent are absent. The horizontal dotted line is the uniform-model limit.
\textbf{B)} At $\eta=.02$, mean leading eigenvalue of the individual-model covariance $C_W=\mathbb E[\mathbf v\mathbf v^\top\mid W]$, divided by $N$ (purple), and mean variance along the dominant teacher pattern, $\mathbb E_W[\boldsymbol\xi_1^\top C_W\boldsymbol\xi_1]/N^2$ (blue). Error bars and grey shading give pointwise $95\%$ dataset intervals; dashed segments guide the eye between posterior estimates. The posterior reverse-KL rise accompanies weakened teacher expression while individual models remain anisotropic. Common parameters: $N=16$, $M=3$, $K=16$.}
\label{fig:small-rbm-gamma}
\end{figure}

\begin{figure}[!htbp]
\centering
\includegraphics[width=\linewidth]{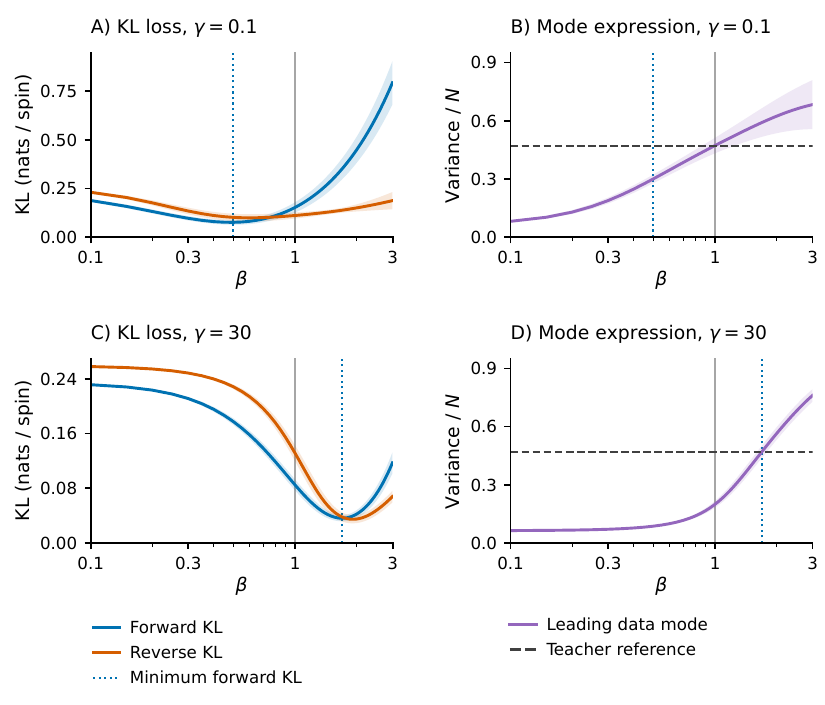}
\caption{\textbf{Generative temperature tuning in the small RBM.}
We vary the joint-energy inverse temperature $\beta$ at fixed fitted weights, with $\gamma=.1$ (top) and $\gamma=30$ (bottom).
\textbf{A,C)} Forward (blue) and reverse (orange) KL per spin; dotted lines mark the minima of the mean forward-KL curves.
\textbf{B,D)} Generated variance along the leading direction of the training-data second-moment matrix, divided by $N$ (purple); horizontal dashed lines give the teacher variance along that direction. Grey vertical lines mark $\beta=1$. Curves average over eight independent training sets; shading gives pointwise $95\%$ dataset intervals. Warming suppresses mode expression and cooling amplifies it; KL improvement need not reduce the mode-variance mismatch. The minima are descriptive selections on these datasets. Common parameters: $N=16$, $M=3$, $K=16$.}
\label{fig:small-rbm-beta}
\end{figure}

We observe warm and cold posterior effects: warming improves prediction under weak regularization, while cooling helps under stronger regularization (Fig.~\ref{fig:small-rbm-eta}). Increasing regularization produces double descent, with reverse KL rising to an intermediate maximum before falling again (Fig.~\ref{fig:small-rbm-gamma}). Near the maximum, the generated samples retain strong correlations while expressing the teacher pattern less strongly. Generation-temperature tuning also improves prediction by warming or cooling the distribution at fixed fitted weights (Fig.~\ref{fig:small-rbm-beta}). Some posterior-predictive estimates retain unresolved Monte Carlo uncertainty beyond the dataset error bars.

\begin{figure}[!htbp]
\centering
\includegraphics[width=\linewidth]{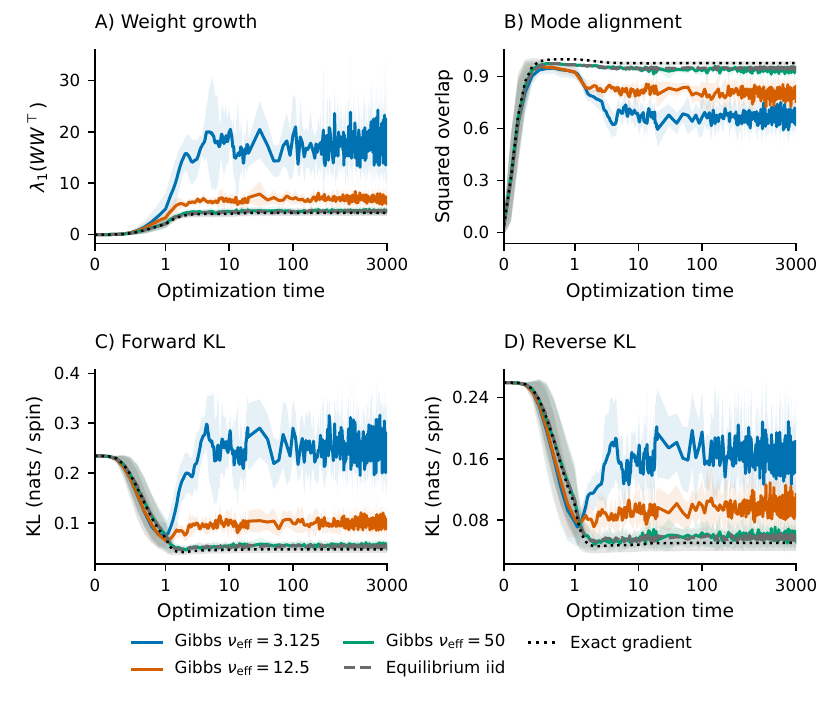}
\caption{\textbf{Out-of-equilibrium learning in the small RBM.}
One persistent particle receives a full blocked Gibbs sweep with probability $p$ per weight update. We measure training time as $t=n\Delta t$ after $n$ updates, with learning rate $\Delta t=.02$. The effective sampling rate $\nu_{\mathrm{eff}}=p/\Delta t$ counts expected sweeps per unit training time. The dashed grey curve uses an independent exact-equilibrium sample at every weight update; the dotted black curve uses exact gradients.
\textbf{A)} Leading eigenvalue of $WW^\top$.
\textbf{B)} Squared overlap of its leading eigenvector with the principal direction of the training-data second-moment matrix. Under slower sampling, alignment first rises and then declines.
\textbf{C,D)} Exact forward and reverse KL per spin. Curves average over eight additional independent training sets after averaging three stochastic repeats within each set; shading gives pointwise $95\%$ dataset intervals. The time axis is linear up to $t=1$ and logarithmic thereafter. Slower sampling amplifies weight modes and worsens late-time generation, favoring early stopping. Common parameters: $N=16$, $M=3$, $K=16$, $\gamma=3.4$.}
\label{fig:small-rbm-nu}
\end{figure}

To study out-of-equilibrium learning, we estimate the model term in the training gradient using one persistent visible configuration. We vary how often a Gibbs sweep refreshes it; its contribution is evaluated at the current weights at every update. Slow refresh amplifies the leading mode of $WW^\top$. Its eigenvector initially aligns with the principal data direction, then loses alignment as generation deteriorates; early stopping therefore helps (Fig.~\ref{fig:small-rbm-nu}). We compare with exact gradients and independent equilibrium sampling. Infrequent exact-equilibrium redraws reproduce much of the degradation, consistent with the persistent configuration lagging behind the changing model.

\clearpage

\subsection{Sampling temperature tuning in Potts BMs trained on protein sequences}\label{app:temperature-tuning}

A prediction of the SBM phase diagram is that $L_{2}$ regularization shrinks the coupling eigenvalues and pushes the trained model toward the $h=0$ phase, where generated samples are near-isotropic noise. Raising the sampling inverse temperature $\beta>1$ amplifies signal modes and can trigger an $h=0$ to $h\ne0$ condensation that rescues these modes, giving a principled mechanism for the widely-used practice of sampling energy-based models at temperatures below the training value~\citep{fields2025}. We now show that this picture quantitatively describes pairwise Potts Boltzmann machines trained on real protein sequence data.

Expanding the sampling covariance of a model trained at $\beta=1$ with regularization $\gamma$ around the training point and choosing $\beta$ to optimize the fit of the model covariance to the empirical data covariance, to leading order in $\gamma$, yields the closed-form optimum
\begin{equation}\label{eq:beta-star-unified}
  \beta_{\mathrm{TT}}^{\mathrm{cov}} = 1 + \gamma\,\frac{\sum_{k}\lambda_{k}^{2}\,c_{k}^{2}}{\sum_{k}\lambda_{k}(\lambda_{k}-\mu^{\prime})\,c_{k}^{4}},
\end{equation}
where $\lambda_{k}$ and $c_{k}$ are, respectively, the coupling eigenvalues of the trained model and the data covariance eigenvalues in a common eigenbasis, and $\mu^{\prime}$ encodes the on-site restoring force set by the normalization constraint: $\mu^{\prime}_{\mathrm{SBM}}=\sum_{k}\lambda_{k}c_{k}^{2}/\sum_{k}c_{k}^{2}$ for the spherical constraint, and $\mu^{\prime}=0$ for the per-site softmax of the Potts model. We use covariance matching rather than minimizing a KL divergence because any KL between the model and the data distribution requires the Potts partition function $Z(\beta J)$ and is intractable in this regime, while the model and data covariances are directly accessible from samples.

\paragraph*{Derivation.} Work in the joint eigenbasis of the trained coupling $J$ and the data covariance $C$, with eigenvalues $\lambda_{k}$ and $c_{k}$, and let $\sigma_{k}^{2}(\beta)$ denote the sampling-covariance eigenvalue at inverse temperature $\beta$. The $L_{2}$-regularized stationarity condition $\langle\mathbf{x}\mathbf{x}^{\top}\rangle_{J}=C-\gamma J$ at $\beta=1$ reads in this basis
\begin{equation}\label{eq:training-stationarity}
  \sigma_{k}^{2}(1)=c_{k}-\gamma\lambda_{k}+\BigO(\gamma^{2}).
\end{equation}
Sampling at $\beta$ rescales $J\mapsto\beta J$. For the spherical model $P_{\beta}(\mathbf{x})\propto e^{(\beta/2)\mathbf{x}^{\top}J\mathbf{x}}\,\delta(\|\mathbf{x}\|^{2}-N)$ the Lagrange multiplier $\mu_{\beta}$ gives $\sigma_{k}^{2}(\beta)=(\mu_{\beta}-\beta\lambda_{k})^{-1}$, hence
\begin{equation}\label{eq:linear-response-beta}
  \partial_{\beta}\sigma_{k}^{2}\big|_{\beta=1}=(\lambda_{k}-\mu^{\prime})\,\sigma_{k}^{2}(1)^{2}=(\lambda_{k}-\mu^{\prime})\,c_{k}^{2}+\BigO(\gamma),
\end{equation}
with $\mu^{\prime}\equiv\partial_{\beta}\mu_{\beta}|_{\beta=1}$. Differentiating the trace constraint $\sum_{k}\sigma_{k}^{2}(\beta)=N$ at $\beta=1$ fixes
\begin{equation}\label{eq:mu-prime}
  \mu^{\prime}_{\mathrm{SBM}}=\frac{\sum_{k}\lambda_{k}c_{k}^{2}}{\sum_{k}c_{k}^{2}}+\BigO(\gamma);
\end{equation}
the Potts softmax enforces normalization site-wise, leaving $\mu^{\prime}=0$. Expanding $\sigma_{k}^{2}(\beta)$ to first order around $\beta=1$ via~\eqref{eq:training-stationarity}--\eqref{eq:linear-response-beta} gives the residual
\begin{equation}\label{eq:beta-residual}
  \sigma_{k}^{2}(\beta)-c_{k}=-\gamma\lambda_{k}+(\beta-1)(\lambda_{k}-\mu^{\prime})\,c_{k}^{2}+\BigO(\gamma^{2}).
\end{equation}
A scalar $\beta$ cannot zero this mode-by-mode; we project onto the coupling direction weighted by the data covariance,
\begin{equation}\label{eq:beta-star-projection}
  \sum_{k}\lambda_{k}\,c_{k}^{2}\bigl(\sigma_{k}^{2}(\beta_{\mathrm{TT}}^{\mathrm{cov}})-c_{k}\bigr)=0.
\end{equation}
Substituting~\eqref{eq:beta-residual} into~\eqref{eq:beta-star-projection} yields
\begin{equation}\label{eq:beta-star-solve}
  -\gamma\sum_{k}\lambda_{k}^{2}c_{k}^{2}+(\beta_{\mathrm{TT}}^{\mathrm{cov}}-1)\sum_{k}\lambda_{k}(\lambda_{k}-\mu^{\prime})c_{k}^{4}=0,
\end{equation}
which solves to Eq.~\eqref{eq:beta-star-unified}.

\subsubsection{Response-regulator PF00072}

\begin{figure}[h]
\centering
\includegraphics[width=\textwidth]{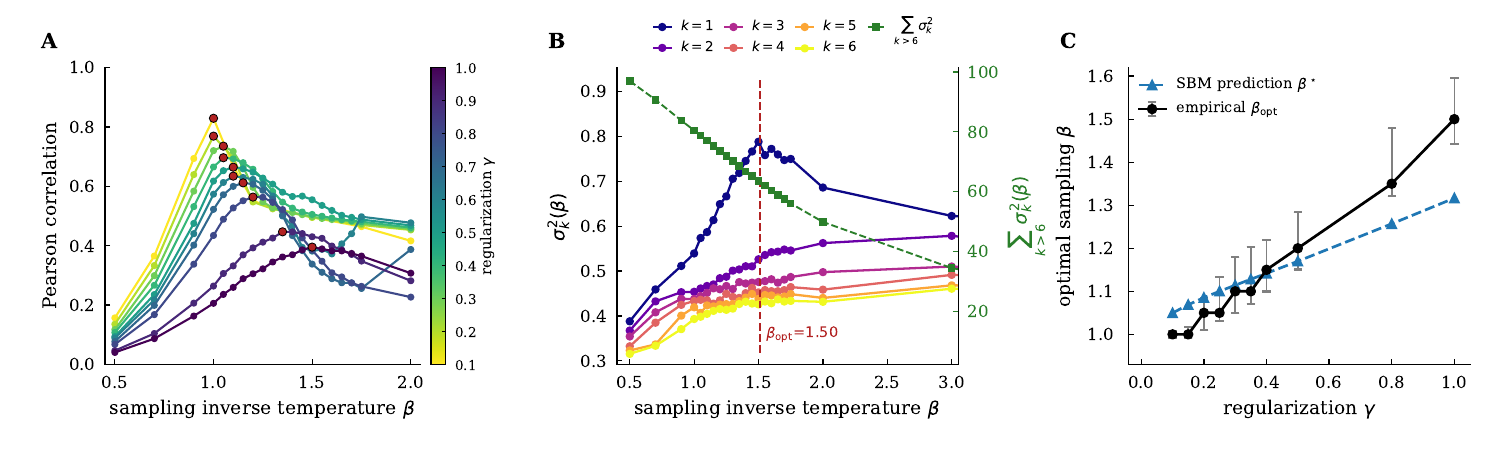}
\caption{\label{fig:temperature-tuning}\textbf{Temperature tuning on Potts BM trained on PF00072.} Pairwise Potts models on the Pfam response-regulator family PF00072 ($L=111$, $q=21$), trained via adabmDCA~\citep{rosset2026adabmdca}. \textbf{A)}~Pearson correlation between generated and data connected correlations as a function of sampling inverse temperature $\beta$, for sixteen $L_{2}$ regularizations $\gamma$ (color bar). Red dots mark the peak of each curve, defining $\beta_{\mathrm{opt}}(\gamma)$. \textbf{B)}~Top six covariance eigenvalues $\sigma_{k}^{2}(\beta)$ of sequences generated by the model trained at $\gamma=1$ (left axis, solid lines). The dominant modes amplify as $\beta$ increases, with the empirical optimum $\beta_{\mathrm{opt}}=1.50$ marked by the red dashed line. The green dashed line with square markers (right axis) shows the sum of the remaining eigenvalues, $\sum_{k>6}\sigma_{k}^{2}(\beta)$, which decreases monotonically with $\beta$ and approaches zero as $\beta\to\infty$. \textbf{C)}~Empirical $\beta_{\mathrm{opt}}(\gamma)$ (black circles, from panel A) compared against the SBM prediction $\beta_{\mathrm{TT}}^{\mathrm{cov}}$ of~\eqref{eq:beta-star-unified} (blue triangles, dashed). For each $\gamma$, $\beta_{\mathrm{opt}}$ is extracted from the Pearson curve on the combined original + refined $\beta$ grid; the error bar shows the range of $\beta$ over which the Pearson correlation stays within one standard error of its peak (standard error estimated from an independent replicate scan with a different random seed, yielding $\sigma_{P}\simeq 5\times 10^{-3}$).}
\end{figure}

We trained pairwise Potts models on the Pfam response-regulator family PF00072 ($L=111$, $q=21$) via adabmDCA~\citep{rosset2026adabmdca} at sixteen regularizations $\gamma\in[0.01,1.0]$. For each trained model we sampled sequences at thirteen inverse temperatures using parallel tempering with $10^{4}$ chains and recorded (i) the Pearson correlation between model and data connected correlations, whose maximum defines the empirical optimum $\beta_{\mathrm{opt}}(\gamma)$, and (ii) the covariance eigenvalue spectrum $\sigma_{k}^{2}(\beta)$ of generated samples. Figure~\ref{fig:temperature-tuning} summarizes the result. Panel~A shows the Pearson correlation as a function of $\beta$ for each regularization; the red dots mark the maximum of each curve and define $\beta_{\mathrm{opt}}(\gamma)$. Panel~B tracks the top six sampling covariance eigenvalues at $\gamma=1$: as $\beta$ increases, the dominant modes detach from the near-degenerate bulk and grow sharply near $\beta\simeq\beta_{\mathrm{opt}}=1.5$, the direct finite-$N$ signature of the SBM condensation. Past $\beta_{\mathrm{opt}}$, however, the top eigenvalue itself declines: as $\beta\to\infty$ the Gibbs measure concentrates on the ground state, so the covariance of generated samples collapses in \emph{every} direction and all $\sigma_{k}^{2}$ vanish. This is made explicit by the green dashed line in panel~B (right axis), which plots the bulk sum $\sum_{k>6}\sigma_{k}^{2}(\beta)$ and decays monotonically toward zero with $\beta$. The total trace $\sum_{k}\sigma_{k}^{2}$ is therefore \emph{not} conserved along $\beta$, unlike the idealized spherical model whose hard constraint $|\mathbf x|^{2}=N$ fixes it exactly. As a result, $\beta_{\mathrm{opt}}$ emerges from the competition between the condensation amplification of signal modes (which grows $\sigma_{1}^{2}$) and the freezing-driven overall shrinkage (which eventually overwhelms it), so that sampling at $\beta$ well past $\beta_{\mathrm{opt}}$ degrades the generative model. Panel~C shows $\beta_{\mathrm{opt}}(\gamma)$ rising monotonically from $1$ as $\gamma$ increases and tracking the SBM prediction~\eqref{eq:beta-star-unified} across the full regularization range; the naive $1/\lambda_{1}$ criterion (not shown) predicts values in the range $0.15$--$0.52$, wrong by factors of $3$--$7$. Residual deviations at large $\gamma$ reflect higher-order corrections in $\gamma$ and a per-site softmax rather than global spherical constraint in the Potts model, but do not affect the qualitative agreement. The same picture holds on a second family, the SH3 domain PF00018, with a more pronounced temperature-tuning effect owing to its smaller effective sample size; we show the corresponding data in App.~\ref{app:temperature-tuning-pf00018} below.

\subsubsection{SH3 domain PF00018}\label{app:temperature-tuning-pf00018}

\begin{figure}
\centering
\includegraphics[width=\textwidth]{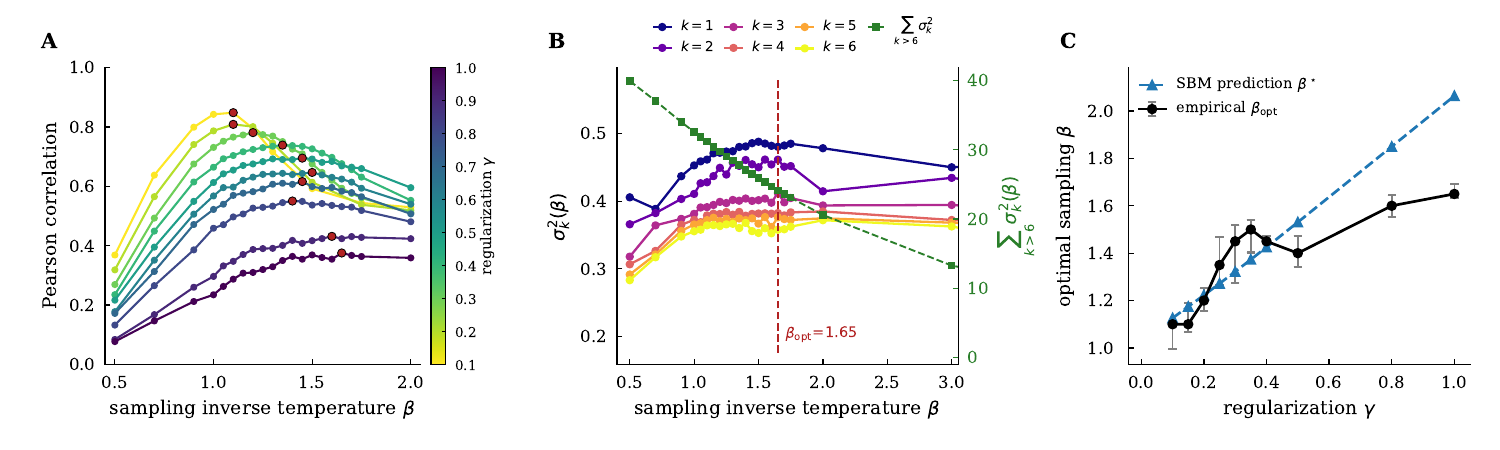}
\caption{\label{fig:temperature-tuning-pf00018}\textbf{Temperature tuning on Potts BM trained on PF00018.} Pairwise Potts models on the Pfam SH3 domain PF00018 ($L=48$, $q=21$), trained via adabmDCA. \textbf{A)}~Pearson correlation between generated and data connected correlations as a function of sampling inverse temperature $\beta$, for sixteen $L_{2}$ regularizations (color bar); red dots mark $\beta_{\mathrm{opt}}$. \textbf{B)}~Top six covariance eigenvalues $\sigma_{k}^{2}(\beta)$ at $\gamma=1$ (left axis, solid lines); green dashed line with square markers (right axis) shows the bulk sum $\sum_{k>6}\sigma_{k}^{2}(\beta)$. \textbf{C)}~Empirical $\beta_{\mathrm{opt}}(\gamma)$ compared against the SBM prediction $\beta_{\mathrm{TT}}^{\mathrm{cov}}$ of~\eqref{eq:beta-star-unified}; error bars show the Pearson-plateau half-width (see Fig.~\ref{fig:temperature-tuning} caption). Conventions as in Fig.~\ref{fig:temperature-tuning}.}
\end{figure}

The SH3 domain PF00018 ($L=48$, $q=21$) sits deeper in the undersampled regime than PF00072; protocols are identical to App.~\ref{app:temperature-tuning}. Fig.~\ref{fig:temperature-tuning-pf00018} shows the temperature-tuning effect is correspondingly stronger, with $\beta_{\mathrm{opt}}\simeq 1.75$ already at $\gamma=0.8$; Eq.~\eqref{eq:beta-star-unified} again tracks the empirical optimum across $\gamma$, with the same large-$\gamma$ bias attributable to finite-order expansion and the softmax constraint.

\subsection{Double descent}\label{app:double-descent-EBM}

\subsubsection{Double descent in a binary FVSBN, large-$K$ behavior and the rank-$1$-data condition}\label{app:fvsbn-K-dd}

We test double descent in a discrete student trained on two teachers whose covariances are dominated by a single mode. We vary the number of training samples from $K\ll N$ to $K=2N$. This tests the regime-(b) extension in Sec.~\ref{sec:phase-diagram}: the equilibrium theory remains valid as $K\to\infty$ provided the empirical covariance $C$ retains $\BigO(1)$ outliers above its bulk.

The student is a binary fully-visible sigmoid belief network (FVSBN)~\citep{frey1998graphical} on $\{\pm 1\}^N$, with $N=16$ and autoregressive log-likelihood
\begin{equation}\label{eq:fvsbn-log-likelihood}
  \ln q_\theta(s) = \sum_i \ln \sigma\!\bigl((2s_i)\bigl(b_i + \sum_{j<i} W_{ij}s_j\bigr)\bigr),
\end{equation}
where $\sigma$ is the logistic function. Its parameters are the strictly lower-triangular matrix $W\in\mathbb{R}^{N\times N}$ and biases $b\in\mathbb{R}^N$. We first maximize the likelihood with $1500$ Adam steps. We then collect SWAG snapshots every $15$ steps during $1500$ SGD-Langevin updates at constant learning rate and $T_{\mathrm{lang}}=0.3$. We fit these snapshots with a diagonal Gaussian, draw $30$ students, and average their reverse KL against the teacher. Setting the optimizer's weight decay to $\mathrm{wd}=N\gamma/(2K)$ makes the loss averaged over training samples match the SBM prior $\propto\exp(-N\gamma\Trace W^2/4)$, so $\gamma$ remains comparable across sample sizes.

\begin{figure}
\centering
\includegraphics[width=\linewidth]{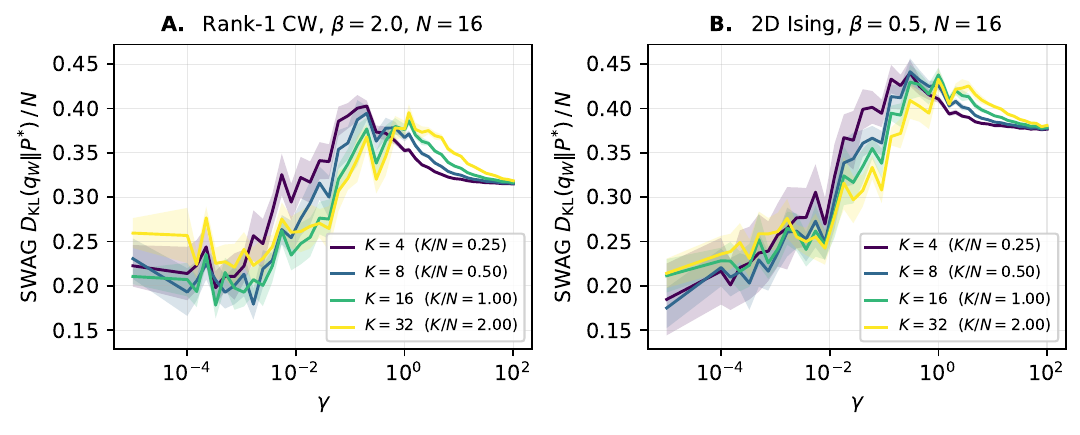}
\caption{%
\textbf{Double descent in a binary FVSBN persists across $K$ on rank-$1$-dominated teachers.}
SWAG posterior-averaged reverse KL per spin, $D_{\mathrm{KL}}(q_{W}\Vert\Pteacher)/N$, vs.\ the paper-convention regularization $\gamma$, for the binary FVSBN trained on \textbf{A)} a rank-$1$ Curie--Weiss / Hopfield teacher at $\beta=2.0$, $N=16$, and \textbf{B)} a 2D Ising teacher on the $L=4$ periodic square lattice ($N=16$) at $\beta=0.5$, deep in the ferromagnetic phase. Each panel overlays four sample sizes $K\in\{4,8,16,32\}$, i.e.\ $K/N\in\{0.25,0.5,1,2\}$, $10$ seeds each; shaded bands are $\pm 1\,\mathrm{SEM}$. The optimizer weight decay is $\mathrm{wd}=N\gamma/(2K)$, matching the SBM prior $\propto\exp(-N\gamma\Trace W^2/4)$ at every $K$. Across both teachers the SWAG-DD bump above the $\gamma=0$ baseline persists at every $K$ ($+0.14$ to $+0.18$ nats/$N$ on CW, $+0.22$ to $+0.27$ on Ising), while the empirical-covariance gap $\lambda_1(C)/\lambda_2(C)$ grows monotonically with $K$ (from $9.7$ to $27.3$ on CW and from $9.5$ to $18.8$ on Ising). The peak position $\gamma_{\mathrm{peak}}$ shifts with $K$ in the direction predicted by the retarded-learning threshold $\gamma\sim\eta c_1^2$ with $c_1\propto K$. The replacement of the geometry-free rank-$1$ Hopfield teacher by the 2D Ising NN lattice leaves the signature qualitatively unchanged, confirming that the relevant condition is a low-effective-rank empirical covariance and not the spatial structure of the teacher.}
\label{fig:fvsbn-K-dd}
\end{figure}

The two teachers have different interaction structures. The first is a rank-$1$ Curie--Weiss / Hopfield model with $\mathcal{E}(s)=-\tfrac{1}{2N}(\xi\cdot s)^2$, $\beta=2.0$, and a random $\xi\in\{\pm 1\}^N$ (Fig.~\ref{fig:fvsbn-K-dd}A). The second is a 2D Ising model on the $L=4$ periodic square lattice ($N=L^2=16$) at $\beta=0.5$, deep in the ferromagnetic phase (Fig.~\ref{fig:fvsbn-K-dd}B). We compute $\ln Z$ exactly by exhaustive enumeration and a column transfer matrix, respectively; for the first teacher, we also evaluate the reverse KL exactly. We use $K\in\{4,8,16,32\}$ with $10$ seeds each and record the spectrum of $C=\tfrac{1}{N}\sum_k x^k x^{k\top}$. As predicted, $\lambda_1(C)$ grows linearly with $K$, while $\lambda_2(C)$ grows much more slowly. From $K=4$ to $32$, the ratio $\lambda_1/\lambda_2$ rises from $9.7$ to $27.3$ for CW and from $9.5$ to $18.8$ for Ising: the empirical covariance remains dominated by its leading mode.

Figure~\ref{fig:fvsbn-K-dd} shows the SWAG-averaged reverse KL per spin. For both teachers, the double-descent bump persists across the full $K$ range, rising above the $\gamma=0$ baseline by $0.14$--$0.18$ nats/$N$ (CW) and $0.22$--$0.27$ nats/$N$ (Ising). Its height depends little on $K$, but its position $\gamma_{\mathrm{peak}}$ moves from $\gamma\approx 0.30$ at $K\le 8$ to $\gamma\approx 1$ at $K\ge 16$, tracking the retarded-learning threshold $\gamma\sim\eta c_1^2$ with $c_1=\lambda_1(C)\propto K$. Thus the same qualitative signature appears for the rank-$1$ Hopfield teacher without spatial structure and the Ising teacher with nearest-neighbor lattice interactions.

\subsubsection{Double descent in an RBM trained on financial data}\label{app:rbm-finance}

The double-descent phenomenon predicted by the theory is not limited to the SBM. A qualitatively similar behavior as a function of regularization strength also appears in a restricted Boltzmann machine (RBM) trained on financial data, where the peak in test loss coincides with a spectral transition in the learned weight matrix, qualitatively consistent with the spectral picture underlying double descent in the SBM.

\paragraph{Data.} We use the Kenneth R.\ French 49 industry portfolio daily value-weighted returns~\citep{fama1997french,laloux1999noise}, spanning July 1926 to February 2026 (14\,286 trading days after removing dates with missing entries). Each industry's returns are standardized to zero mean and unit variance on the training set, and winsorized at $\pm 5\sigma$ to limit the effect of extreme events. The empirical covariance of the standardized training returns has a dominant eigenvalue $\lambda_{1}\approx 25$, well separated from $\lambda_{2}\approx 1.6$ ($\lambda_{1}/\lambda_{2}\approx 15$), reflecting the single ``market mode'' that drives correlated daily fluctuations across sectors. The data is split chronologically into 80\%\ training (11\,428 days) and 20\%\ test (2\,858 days).

\paragraph{Model and training.}
We train a Gaussian-visible, Bernoulli-hidden RBM with $N=49$ visible units (one per industry) and $M=20$ hidden units, using persistent contrastive divergence (PCD)~\citep{tieleman2008training} with $L_{2}$ weight decay of strength~$\gamma$ applied to the weight matrix~$W\in\mathbb{R}^{N\times M}$. The visible biases and precisions are unregularized. Training consists of 6\,000 Adam~\citep{kingma2015adam} burn-in iterations followed by 2\,000 SGD iterations at constant learning rate (the latter phase used for SWAG snapshot collection~\citep{maddox2019simple}, though posterior averaging turns out to have a negligible effect here). The log-partition function is estimated by annealed importance sampling~\citep{salakhutdinov2008learning} with 10\,000 inverse-temperature steps and 15 chains. We sweep $\gamma$ from $10^{-6}$ to $10$ and repeat each run for 8 independent training seeds at fixed data subset; error bars show $\pm 1$ standard deviation across seeds.

\begin{figure}
\centering
\includegraphics[width=0.7\columnwidth]{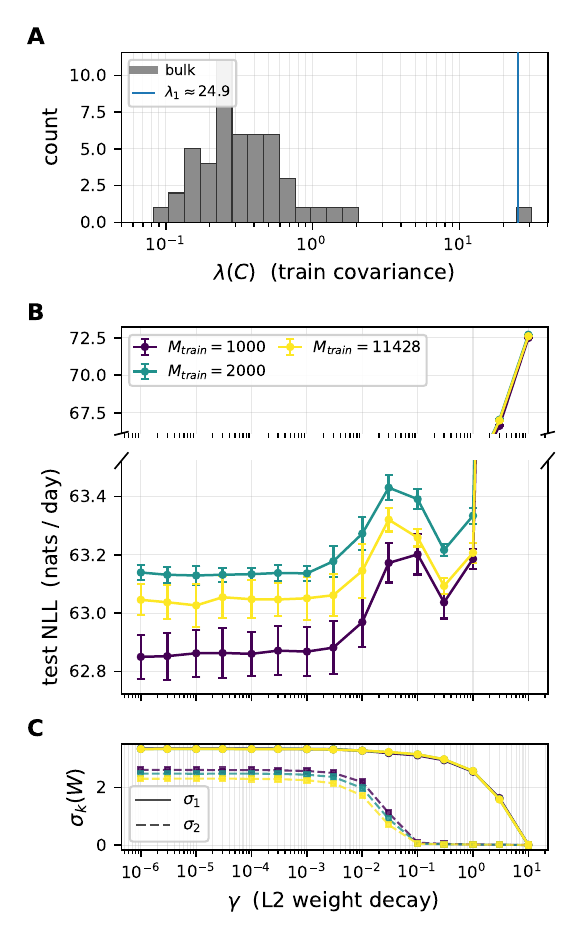}
\caption{%
\textbf{Double descent in a Gaussian-visible RBM trained on financial return data.}
\textbf{A)} Eigenvalue spectrum of the empirical training-set covariance $C$ on a log-$\lambda$ axis. The bulk (gray histogram) spans $\lambda\!\sim\!10^{-1}$ to $\lambda\!\sim\!2$, dominated by sectoral correlations; a single rank-one outlier sits at $\lambda_{1}\approx 25$ (blue line), more than an order of magnitude above the bulk maximum $\lambda_{2}\approx 1.6$, the ``market mode'' that drives the retarded-learning transition seen in panel~\textbf{B}.
\textbf{B)} Test negative log-likelihood (NLL) per day as a function of $L_{2}$ weight decay~$\gamma$, for training-set sizes $M_{\mathrm{train}}=1000$, $2000$, and $11428$ (full training set). A broken $y$-axis separates the detailed view of the double-descent bump (lower part, $62.8$--$63.5$ nats/day) from the high-$\gamma$ collapse to the uncorrelated-Gaussian floor (upper part, $\approx 67$--$73$ nats/day).
\textbf{C)} Top two singular values $\sigma_{1}$ (solid) and $\sigma_{2}$ (dashed) of the trained weight matrix~$W$. At $\gamma\lesssim 10^{-2}$, both are comparable ($\sigma_{1}\approx 3$, $\sigma_{2}\approx 2.5$); at $\gamma\approx 3\times 10^{-2}$--$10^{-1}$, $\sigma_{2}$ drops to zero while $\sigma_{1}$ persists, marking the spectral transition that coincides with the NLL bump.
Data: Kenneth French 49 industry daily value-weighted returns (1926--2026), standardized and winsorized at $\pm 5\sigma$. Model: Gaussian--Bernoulli RBM ($N{=}49$, $M{=}20$), trained by PCD with Adam burn-in; log-partition function estimated by AIS. Error bars: $\pm 1\,\sigma$ over 8 training seeds at fixed data subset.
}
\label{fig:rbm-finance}
\end{figure}

\paragraph{Results.}
Figure~\ref{fig:rbm-finance} displays the eigenvalue spectrum of the empirical training-set covariance $C$ (panel~\textbf{A}), the test negative log-likelihood (NLL) per day (panel~\textbf{B}), and the two largest singular values $\sigma_{1}$, $\sigma_{2}$ of the trained weight matrix~$W$ (panel~\textbf{C}), for three training-set sizes $M_{\mathrm{train}}\in\{1000, 2000, 11428\}$.

At small~$\gamma$, the test NLL reaches a low plateau ($\approx 63$~nats/day) where the RBM captures both the market-mode correlation and sub-leading sector structure: both $\sigma_{1}$ and $\sigma_{2}$ are of comparable magnitude ($\sigma_{1}\approx 3$, $\sigma_{2}\approx 2.5$). As $\gamma$ increases past $\sim 10^{-2}$, $\sigma_{2}$ drops sharply toward zero while $\sigma_{1}$ remains stable: $L_{2}$ regularization kills the secondary modes before the dominant one. At this transition, the test NLL rises to a local maximum (the ``bump''), with amplitude $\Delta\mathrm{NLL}\approx 0.3$~nats/day, reproducible across runs (5--6\,$\sigma$). Beyond the transition ($\gamma\approx 0.3$), the NLL partially recovers as the model commits to a clean rank-one representation of the market mode (only $\sigma_{1}$ survives), before finally collapsing to the uncorrelated-Gaussian floor ($\approx 73$~nats/day) when $\gamma\gtrsim 1$ kills all weights.

The non-monotonic NLL profile (plateau, bump, partial recovery, then collapse) mirrors the double-descent structure predicted by the SBM theory (Fig.~\ref{fig:double-descent}): the left minimum corresponds to the $h\ne 0$ phase (signal captured), the bump to the spectral transition where secondary modes merge into the effective noise floor, and the right-side recovery to the regime where the model operates with a single clean outlier. As $M_{\mathrm{train}}$ decreases, the bump shifts to larger~$\gamma$ (from $\gamma\approx 3\times 10^{-2}$ at $M_{\mathrm{train}}=11428$ to $\gamma\approx 10^{-1}$ at $M_{\mathrm{train}}=1000$), consistent with a noisier empirical covariance requiring stronger regularization to trigger the spectral transition.

The bump is visible already in the MAP (point-estimate) test NLL, without requiring Bayesian posterior averaging. The reported MAP is the endpoint of the $2\,000$-step constant-learning-rate SGD phase rather than a true posterior argmax, and so retains residual minibatch noise; the SWAG posterior-predictive NLL, averaging over snapshots from this same phase, follows the single-endpoint curve within error bars. The spectral mechanism underlying the double descent, $L_{2}$-driven merging of a data eigenvalue into the noise bulk, is therefore robust enough to affect even the point estimate, provided the data has a clean rank-one structure.

\subsection{Tempered posterior effects}\label{app:tempered-posterior-gen}

\subsubsection{BayesGAN}

The closed-form analysis of Fig.~\ref{fig:warm-cold-pp} predicts that the posterior-predictive reverse KL of an SBM student has a $\gamma$-dependent optimum $\eta_{\mathrm{pp}}^{\mathrm{rev}}(\gamma)$: warm ($\eta_{\mathrm{pp}}^{\mathrm{rev}}<1$) at small $\gamma$ and cold ($\eta_{\mathrm{pp}}^{\mathrm{rev}}>1$) at large $\gamma$. We test whether the same warm-to-cold migration is visible in a deep generative model, taking the Bayesian GAN of \citep{saatchi2017bayesian} as a convenient example. Tempering is straightforward, even though the posterior is genuinely multimodal in weight space, as the original sampler (SGHMC \citep{chen2014stochastic}) admits the tempered target without modification.

\paragraph{Setting.}
We replace the joint posterior of \citep{saatchi2017bayesian} by its tempered counterpart $p_\eta(\theta)\propto[\mathcal{L}(\theta)\,p(\theta)]^{\eta}$, with $\eta{=}1$ the Saatchi--Wilson Bayesian setting and $\eta\to\infty$ the classical maximum-likelihood GAN. The SGHMC update of \citep{chen2014stochastic} with gradient term scaled by $\eta$ targets $p^{\eta}$ at stationarity. The generator $G(z;\theta_g)$ and discriminator $D(x;\theta_d)$ are two-layer ReLU MLPs (hidden $32{\times}32$, $z\in\mathbb{R}^4$). Algorithm~1 of \citep{saatchi2017bayesian} is run with $J_g{=}4$ generator chains and $J_d{=}1$ discriminator chain, $M{=}2$ inner steps, friction $\alpha{=}0.1$, step size $\varepsilon{=}5\times 10^{-5}$, an isotropic Gaussian prior $\mathcal{N}(0,\sigma_p^2 I)$ on every weight, and a $500$-iteration Adam warm-up before SGHMC sampling, as recommended in \citep{saatchi2017bayesian}. The predictive density $\PredPost$ is estimated by isotropic Gaussian KDE ($\sigma_{\mathrm{KDE}}{=}0.1$) on samples pooled across all generator chains and post-burn-in snapshots, which marginalizes weights and noise jointly.

\begin{figure}
\centering
\includegraphics[width=0.9\linewidth]{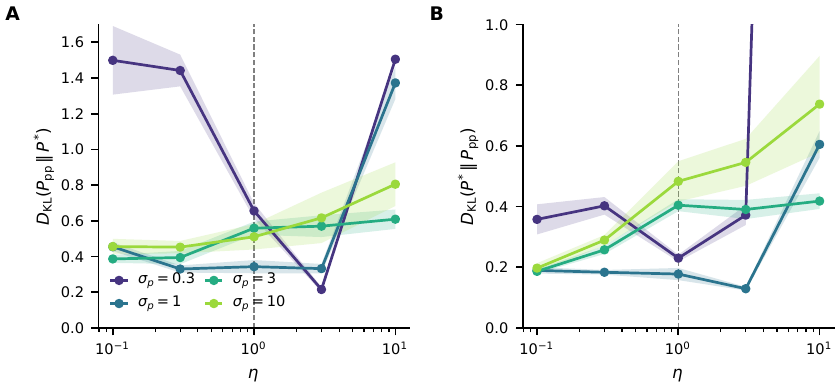}
\caption{%
\textbf{Tempered Bayesian GAN on a unimodal target: warm-to-cold migration with prior strength.}
\textbf{A)}~Posterior-predictive reverse KL, $D_{\rm KL}(\PredPost\Vert \Pteacher)$, versus $\eta$, for four prior widths $\sigma_p\in\{0.3,1,3,10\}$.
\textbf{B)}~Forward $D_{\rm KL}(\Pteacher\Vert \PredPost)$.
Target: single isotropic Gaussian at the origin with $\sigma_{\rm target}{=}0.5$. Up to twelve seeds per $(\sigma_p,\eta)$ cell. Bands are $\pm 1\,\mathrm{SEM}$. Vertical dashed line: $\eta=1$ (Bayes).
}
\label{fig:bayesgan-unimodal}
\end{figure}

\paragraph{Target and sweep.}
The target is a single 2-D isotropic Gaussian at the origin with $\sigma_{\rm target}{=}0.5$. Figure~\ref{fig:bayesgan-unimodal} sweeps $(\sigma_p, \eta)\in\{0.3,1,3,10\}\times\{0.1,0.3,1,3,10\}$ at up to twelve seeds per cell. The Gaussian prior $\mathcal{N}(0,\sigma_p^{2}I)$ on the BayesGAN weights identifies $\sigma_p$ with the SBM prior standard deviation $\sqrt{2/(N\gamma)}$, so smaller $\sigma_p$ corresponds to larger $\gamma$. The closed-form prediction of Fig.~\ref{fig:warm-cold-pp} (warm at small $\gamma$, cold at large $\gamma$) therefore maps onto warm at large $\sigma_p$, cold at small $\sigma_p$. In $D_{\rm KL}(\PredPost\Vert \Pteacher)$ (panel~A), the argmin shifts as predicted: from $\eta_{\mathrm{pp}}^{\mathrm{rev}}{=}0.3$ at weak priors ($\sigma_p\geq 1$) to $\eta_{\mathrm{pp}}^{\mathrm{rev}}{=}3$ at the strongest prior ($\sigma_p{=}0.3$). The forward $D_{\rm KL}(\Pteacher\Vert \PredPost)$ (panel~B) has a cold optimum at $\sigma_p{=}1$. Both forward and reverse KLs place the optimum cold for small $\sigma_p$ and warm for large $\sigma_p$. At $\sigma_p{=}0.3$, $\eta{=}10$ the tight prior freezes the chain near $\theta\approx 0$, off the target; the forward KL in panel~B then diverges because the KDE of $\PredPost$ leaves the support of $\Pteacher$ uncovered, while the reverse KL in panel~A stays bounded since $\Pteacher$ is positive everywhere.

\subsection{Out-of-equilibrium training dynamics in a Potts Boltzmann machine}\label{app:ooe-potts}

The training dynamics of Sec.~\ref{sec:dynamics} predicts that the stationary sample--data overlap $s_{\mathrm{st}}$ undergoes a phase transition at the critical sampling rate $\nu_{c}=\gamma/(c_{1}\chi_{P})$ [Fig.~\ref{fig:dynamics}C]: above $\nu_{c}$ the persistent chain aligns with the data spike, while below $\nu_{c}$ the chain remains uncondensed ($s=0$) even though the weight matrix has already acquired the data signal. This out-of-equilibrium (OOE) phenomenon mirrors the empirical findings of~\citep{decelle2021equilibrium}, who demonstrated that RBMs trained by persistent contrastive divergence operate in qualitatively distinct equilibrium or OOE regimes depending on whether the per-step MCMC budget exceeds or falls below the model's mixing time.

We test whether this OOE mechanism is observable in a discrete Potts Boltzmann machine trained on real sequence data. Specifically, we train a pairwise Potts model on sequences sampled from a lattice-protein teacher at $\beta_{\mathrm{sel}}=1000$ ($L=27$ sites, $q=20$ amino acids), using PCD in the MAP limit ($\eta\to\infty$, i.e.\ no weight noise), with $L_{2}$ regularization $\gamma=0.01$ and learning rate $\eta_{\mathrm{lr}}=0.01$. We control the sampling rate via $k$, the number of single-site Gibbs update attempts per gradient step (one full sweep corresponds to $k=L=27$).

\textbf{Order parameters.}
We monitor the chain variances $\sigma_{a}^{2}=\langle[(\mathrm{oh}(\mathbf{x})-f_{1})^{\top}\mathbf{c}_{a}]^{2}\rangle$ along the data-spike directions, the Potts analog of the SBM signal-overlap variance $\langle s_{a}^{2}\rangle$, and the top coupling eigenvalue $\max_{k}|\lambda_{k}(J)|$, the analog of $\lambda_{1}(W)$. Here $f_{1}\equiv\langle\mathrm{oh}(\mathbf{x})\rangle_{\mathrm{data}}$ are the empirical one-site frequencies and $\mathbf{c}_{a}$ are the eigenvectors of the centered one-hot data covariance. The top coupling eigenvalue integrates the OOE bias over the full training trajectory: when the negative phase is biased ($\nu<\nu_{c}$), the data term in~\eqref{eq:W-langevin} is not canceled, driving the coupling eigenvalues beyond their equilibrium values. The weight \emph{direction}, measured by the subspace overlap $\sum_{a}\|\mathrm{Proj}_{J}\mathbf{c}_{a}\|^{2}$ between the top eigenvectors of $J$ and the data eigenvectors, is predicted to be sampling-rate independent, since eigenvalue detachment occurs at the early-time scale~\eqref{eq:lambda-para} regardless of $\nu$.

\begin{figure}
\centering
\includegraphics[width=\textwidth]{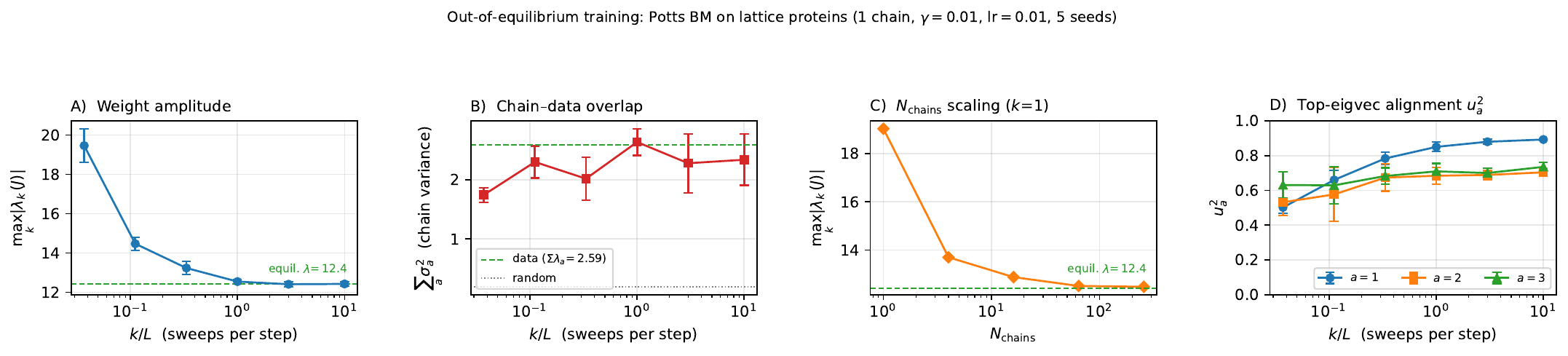}
\caption{\label{fig:ooe-potts}%
\textbf{Out-of-equilibrium training in a Potts BM.}
We train a Potts BM ($L{=}27$, $q{=}20$) on lattice-protein sequences generated at $\beta_{\mathrm{sel}}{=}1000$, using PCD--MAP with $\gamma{=}0.01$ and $\eta_{\mathrm{lr}}{=}0.01$. Training lasts $t_{\mathrm{age}}{=}5000$ steps with $N_{\mathrm{chains}}{=}1$, except in panel~C. Error bars in A, B and D show $\pm 1\sigma$ over five seeds.\newline
\textbf{A)}~Top coupling eigenvalue $\max_{k}|\lambda_{k}(J)|$ versus $k/L$, the number of Gibbs sweeps per gradient step. At small $k$, the single-chain negative phase fails to balance the data gradient, and the weights overshoot equilibrium (dashed line).
\textbf{B)}~Chain variance in the data-spike directions, $\sum_{a}\sigma_{a}^{2}$, averaged over the final $20\%$ of training. It rises from $\approx 1.7$ at $k{=}1$ to $\approx 2.5$ at $k\geq L$, matching the data value $\sum\lambda_{a}=2.59$ (dashed line).
\textbf{C)}~At $k{=}1$, the weight overshoot disappears as $N_{\mathrm{chains}}$ increases and averaging reduces single-chain noise. This agrees with the absence of visible OOE in standard DCA training ($N_{\mathrm{chains}}\sim 500$) on these families.
\textbf{D)}~Alignment $u_{a}^{2}=\max_{k}(\hat v_{k}(J)\!\cdot\!\hat{\mathbf{c}}_{a})^{2}$ for each data mode. The leading-mode alignment rises from $0.50$ at $k{=}1$ to $0.89$ at $k{=}10L$. Other modes lose comparable alignment at small $k$, extending the $u_{1}^{2}<1$ signature to several data directions.
}
\end{figure}

\textbf{Results.}
Figure~\ref{fig:ooe-potts}A shows the final $\max_{k}|\lambda_{k}(J)|$ after $t_{\mathrm{age}}=5000$ gradient steps, as a function of $k/L$ (the number of sweeps per step), with a single persistent chain ($N_{\mathrm{chains}}=1$). At $k=1$ (one site update per step), the top eigenvalue reaches $\lambda\approx 19.5$, overshooting the equilibrium value $\lambda_{\mathrm{eq}}\approx 12.4$ (attained at $k\geq 81$, dashed line) by $\sim 57\%$. The overshoot decreases monotonically with $k$ and plateaus above $k/L\approx 3$. This behavior is robust across five independent training seeds (error bars in Fig.~\ref{fig:ooe-potts}A).

Panel~B shows the chain-side observable: the variance $\sum_{a}\sigma_{a}^{2}$ of the persistent chain in the data-spike directions. At $k=1$ the chain explores only $\sum\sigma^{2}\approx 1.7$, roughly $30\%$ below the data eigenvalue sum $\sum\lambda_{a}=2.59$ (dashed), while at $k\geq L$ the chain variance rises to match the data value. This is the Potts analog of the $s_{\mathrm{st}}(\nu)$ bifurcation of Fig.~\ref{fig:dynamics}C, though noisier because a single chain provides only one sample per measurement.

Panel~C shows the complementary effect: at fixed $k=1$, increasing $N_{\mathrm{chains}}$ from $1$ to $256$ suppresses the weight overshoot, converging to $\lambda_{\mathrm{eq}}\approx 12.5$ by $N_{\mathrm{chains}}\approx 64$. On the protein families considered here, standard DCA runs with $N_{\mathrm{chains}}\sim 500$ and $k\sim 10$ show no visible OOE signature, consistent with this observation.

All three panels confirm that the weight \emph{direction} (eigenvector alignment with data) is $k$-independent ($\sum_{a}\|\mathrm{Proj}_{J}\mathbf{c}_{a}\|^{2}\approx 2.5$ throughout, not shown), consistent with the theory's prediction that representation acquisition precedes, and is independent of, sample condensation: the ``representation without generation'' signature.

\begin{figure}
\centering
\includegraphics[width=\textwidth]{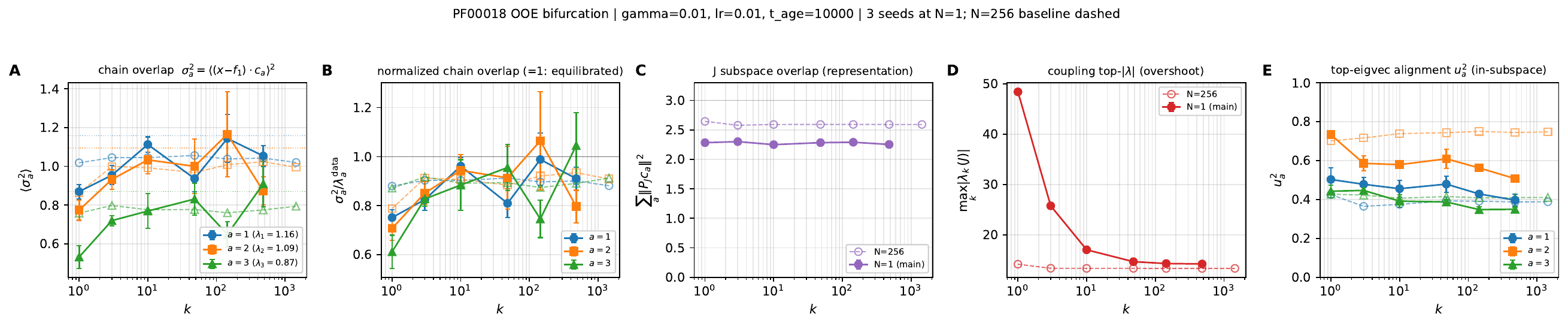}
\caption{\label{fig:ooe-pf00018}%
\textbf{Out-of-equilibrium training dynamics on real protein data (SH3 domain PF00018).}
Following Fig.~\ref{fig:ooe-potts}, we train a pairwise Potts BM on PF00018 ($L{=}48$, $q{=}21$, $M{=}1.76\times 10^{5}$). Training uses PCD--MAP with $\gamma{=}0.01$, $\eta_{\mathrm{lr}}{=}0.01$, and $t_{\mathrm{age}}{=}10^{4}$. Solid markers with error bars show results for $N_{\mathrm{chains}}{=}1$, averaged over 3 seeds. Open markers and dashed lines show the $N_{\mathrm{chains}}{=}256$ equilibrium baseline. The data directions $\mathbf{c}_{1},\mathbf{c}_{2},\mathbf{c}_{3}$ are the leading eigenvectors of the reweighted centered one-hot MSA covariance. Their eigenvalues are $\lambda_{1}^{\mathrm{data}}{=}1.16$, $\lambda_{2}^{\mathrm{data}}{=}1.09$, $\lambda_{3}^{\mathrm{data}}{=}0.87$ (dotted lines in A). Time averages use the second half of training.\newline
\textbf{A)}~Chain variance $\sigma_{a}^{2}$ along each data direction.
\textbf{B)}~Normalized chain variance $\sigma_{a}^{2}/\lambda_{a}^{\mathrm{data}}$. Below $k\approx L/5$, the ratio remains below $1$: the chain has not equilibrated in the data subspace. At larger $k$, it fluctuates about $1$ for all three modes.
\textbf{C)}~Subspace overlap $\sum_{a}\|\mathrm{Proj}_{J}\mathbf{c}_{a}\|^{2}$ (maximum $3$). It is independent of $k$ with both $1$ and $256$ chains. This confirms the ``representation without generation'' prediction: $J$ acquires the data directions regardless of sampler equilibration.
\textbf{D)}~Top coupling eigenvalue $\max_{k}|\lambda_{k}(J)|$. With one chain at $k{=}1$, it overshoots the equilibrium value $\lambda_{\mathrm{eq}}\approx 13.3$ by a factor $\approx 3.6$. This reproduces the overshoot seen in Fig.~\ref{fig:ooe-potts}.
\textbf{E)}~Per-mode alignment $u_{a}^{2}=\max_{k}(\hat v_{k}(J)\!\cdot\!\hat{\mathbf{c}}_{a})^{2}$. Mode $a{=}2$ stays $\sim 0.20$ below the baseline with $256$ chains throughout the $k$ range; modes $a{=}1,3$ agree within the variability across seeds. This alignment loss depends on the training regime and is weaker than for the lattice-protein teacher because the three leading data eigenvalues are nearly degenerate.
}
\end{figure}

\textbf{Extension to real protein data.}
The same protocol on the SH3 domain PF00018 (alignment of App.~\ref{app:temperature-tuning-pf00018}; $t_{\mathrm{age}}=10^{4}$, three seeds) reproduces the three OOE signatures (Fig.~\ref{fig:ooe-pf00018}): at $k{=}1$ the weight amplitude overshoots $\lambda_{\mathrm{eq}}\approx 13.3$ by a factor $3.6$, the normalized chain variance $\sigma_{a}^{2}/\lambda_{a}^{\mathrm{data}}\approx 0.7$ is suppressed in all three top data directions, and the $J$-to-data subspace overlap stays within $1\%$ of its $N_{\mathrm{chains}}{=}256$ value. All three signatures collapse onto equilibrium by $k\approx L/5$--$L/3$, confirming that the OOE mechanism persists when the lattice-protein teacher is replaced by real MSA correlations.

\clearpage
\section{Parameter values used in main-text figures}\label{app:fig-parameters}

This appendix collects the numerical parameter values used to produce each main-text figure.

\paragraph{Fig.~\ref{fig:k1-phase} (Equilibrium phase diagram for $K{=}1$).} Panel~A is a schematic; panel~B is the analytic phase diagram in the $(\gamma,\eta)$ plane. Panel~C uses $\eta=0.5$ (small-$\eta$ regime, $\lambda_{1}$ sticks to the bulk edge); panel~D uses $\eta=5$ (larger-$\eta$ regime, $\lambda_{1}$ detaches before returning to the edge).

\paragraph{Fig.~\ref{fig:dynamics} (Training dynamics).} Panels~A and~B use $\gamma=0.4$, $\eta=10$, $\nu=0.7$, $c_{1}=1.7$, $c_{2}=0.3$, with seed overlaps $s_{1}(0)=s_{2}(0)=0.1$. Panel~C uses $\gamma=0.9$, $\eta=3$. Panel~D shows the MAP limit $\eta\to\infty$.

\paragraph{Fig.~\ref{fig:tt-pmo} (Sampling temperature tuning).} Rank-one teacher with $\omega^{\ast}=2.5$ ($c_{1}=1.6$, $c_{2}=0.4$); student at $\eta=5$. Panel~B shows curves at $\gamma\in\{1.7,2.0,3.0\}$.

\paragraph{Fig.~\ref{fig:double-descent} (Double descent).} Rank-one teacher with $\omega^{\ast}=2.5$ ($c_{1}=1.6$, $c_{2}=0.4$). Solid curves cover $\eta\in\{0.1,0.2,0.4,0.7,1.0,1.5,2.0,3.0,5.0\}$. The two orange dashed curves are the critical analytic anchors: the saddle-node threshold $\eta=\eta_{\mathrm{DD}}(\omega^{\ast}{=}2.5)\approx 1.342$ (upper), below which the first descent disappears, and the MAP limit $\eta\to\infty$ (lower), in which the second descent collapses to a flat PMo plateau.

\paragraph{Fig.~\ref{fig:warm-cold-pp} (Tempered-posterior).} Panel~A sweeps $(\omega^{\ast},\gamma)$. Panel~B fixes $\omega^{\ast}=2.2$ and shows the posterior-predictive reverse KL versus $\eta$ at $\gamma\in\{0.1,0.3,1.0\}$, representative of the warm, cold, and MAP optima.

\paragraph{Fig.~\ref{fig:ooe-wrap} (Out-of-equilibrium training trajectories).} Finite-$N$ Langevin simulations with $N=1500$ and rank-one teacher $\omega^{\ast}=2.5$, at $\gamma=0.4$, $\eta=10$. Color encodes the sampling rate $\nu$, swept over the 15 values $\nu\in\{0.05,0.10,0.15,0.20,0.25,0.30,0.40,0.55,0.70,0.85,1.00,1.30,1.70,2.20,3.00\}$ (roughly log-spaced, straddling $\nu_{c}$).

\newpage

\end{document}